\documentclass{article}
\usepackage{iclr2027_conference,times}

\usepackage{amsmath,amsfonts,bm}

\def\eqref#1{equation~\ref{#1}}

\def\1{\bm{1}}

\DeclareMathAlphabet{\mathsfit}{\encodingdefault}{\sfdefault}{m}{sl}
\SetMathAlphabet{\mathsfit}{bold}{\encodingdefault}{\sfdefault}{bx}{n}

\usepackage{amssymb} 
\usepackage{float}
\usepackage{hyperref}
\hypersetup{
  colorlinks=true,
  linkcolor={black!60!blue},   
  citecolor={black!60!blue},   
  urlcolor={black!60!blue},    
}
\usepackage{url}
\usepackage{xcolor} 
\usepackage[inline]{enumitem} 
\usepackage{booktabs} 
\usepackage{multirow} 
\usepackage{caption} 
\usepackage{subcaption} 
\DeclareCaptionFont{eightpt}{\fontsize{8}{9.5}\selectfont} 
\usepackage{graphicx} 
\usepackage{wrapfig} 
\usepackage{etoc} 
\usepackage[most]{tcolorbox} 
\usepackage{fvextra} 
\newtcolorbox{promptbox}[1]{%
  breakable, enhanced, sharp corners,
  colback=black!2, colframe=black!60!blue, coltitle=white,
  boxrule=0.5pt, left=5pt, right=5pt, top=3pt, bottom=3pt,
  fonttitle=\bfseries\footnotesize, title={#1},
  attach boxed title to top left={xshift=6pt, yshift=-2pt},
  boxed title style={colback=black!60!blue, sharp corners, boxrule=0pt},
}
\DefineVerbatimEnvironment{promptverb}{Verbatim}%
  {fontsize=\footnotesize,breaklines=true,breakanywhere=true,%
   breaksymbolright=\small\textcolor{black!45}{$\hookleftarrow$}}
\graphicspath{{images/}} 

\title{READ-Bench: Benchmarking Historical Instance Retrieval for Time-Series Diagnosis}

\author{%
\textbf{Gerardo Pastrana}$^{1,}$\thanks{Equal contribution.\quad
Corresponding author: \texttt{gerardo.pastrana@c3.ai}.\quad
Haojun Li's work was done during an internship at C3 AI.}$^{*}$,\;
\textbf{Haojun Li}$^{2,*}$,\;
\textbf{Dhruv Mehta}$^{1}$,\;
\textbf{Anoushka Vyas}$^{1}$, \\
\textbf{Sina Khoshfetrat Pakazad}$^{1}$,\;
\textbf{Henrik Ohlsson}$^{1}$,\;
\textbf{John Paparrizos}$^{2}$ \\[2pt]
$^{1}$C3 AI \quad $^{2}$The Ohio State University \\[2pt]
\texttt{\{gerardo.pastrana, dhruv.mehta, anoushka.vyas\}@c3.ai} \\
\texttt{\{sina.pakazad, henrik.ohlsson\}@c3.ai} \\
\texttt{\{li.14118, paparrizos.1\}@osu.edu}
}

\iclrfinalcopy 
\begin{document}

\maketitle

\lhead{Arxiv Preprint}

\begin{abstract}

Time-series diagnostic systems rarely rely on retrieving relevant historical cases, and when they do, retrieval is typically evaluated only indirectly through downstream prediction. We introduce \textbf{READ-Bench}, a benchmark for historical-case retrieval across 12 diagnostic datasets, centered on multivariate time series, that defines relevance by shared fault or event type rather than signal shape, so that visually different traces of the same fault count as relevant while similar-looking traces of different faults do not. Organizing existing datasets into queries, corpora, and explicit relevance judgments, READ-Bench is, to our knowledge, the broadest testbed to date for historical-case retrieval across diagnostic domains. Analogous to retrieval-augmented generation, we treat retrieval as a base retriever followed by a reranker, evaluating classical distances, symbolic retrievers, self-supervised and foundation-model embedders, and their fusion for search, and label-aware and language-model rerankers for reranking, under one protocol that varies supervision, normal-series pollution, and corpus scale with significance testing across datasets. Under a common channel-independent retrieval interface, pretrained representations offer no statistically detectable advantage over strong classical and symbolic baselines for search alone. The decisive factor is instead a small amount of resolved-case supervision at reranking, namely a Gaussian-process reranker that propagates a few neighbor labels in the embedding space and improves rankings far more than swapping among more sophisticated unsupervised representations or language-model reasoning, a gain that holds under corpus pollution and at full corpus scale. Guided by these findings, we introduce normal-residual scoring, which ranks each window by its departure from normal operation, and build a system that fuses a normal-residual-scored foundation-model embedder with a dynamic time warping leg by reciprocal-rank fusion, then reranks with the label-aware Gaussian-process reranker. This system improves NDCG@10 over its own search stage on all 12 datasets, by $+0.11$ from the reranking step alone and by $+0.18$ over the strongest single base retriever applied uniformly across datasets. The benchmark, dataset, and submission protocol are available at \url{https://huggingface.co/spaces/c3aiia3c/read-bench-leaderboard}.

\end{abstract}

\vspace{-0.2cm}
\section{Introduction}
\label{sec:intro}
\vspace{-0.2cm}

Operational systems accumulate a rich history of prior experience, including sensor traces, process histories, supply-chain signals, and other time series, often accompanied by metadata such as annotations, maintenance records, shift notes, incident logs, and transaction histories~\citep{liu2024advancing, varma1999icarus, zhong2018text}. Together, these records capture not only what happened, but also the conditions under which it happened, how it was diagnosed, and what actions followed. Effective diagnostic systems should be able to retrieve relevant historical cases from this repository and use them as evidence for interpreting new observations.

Retrieval-augmented generation (RAG) established this pattern for unstructured corpora that consists in  retrieving relevant external evidence, and then reasoning over that evidence rather than relying only on the parametric knowledge of the language model \citep{lewis2020rag,yasunaga2023racm3}. In knowledge-intensive settings, however, the quality of a RAG system depends on multiple interacting components, including similarity search, reranking, and downstream reasoning. Understanding and improving such systems therefore requires evaluating these components at the appropriate level of granularity. Aggregate end-to-end metrics can mask important differences across retrieval settings, modalities, and query types, whereas targeted evaluation can reveal where performance gains and failures actually arise \citep{wu2024synthetic}. Given the central role of the retrieval component, which commonly encompasses similarity search followed by reranking, benchmarks such as MS MARCO and BEIR \citep{bajaj2016ms,thakur2021beir} helped make retrieval itself a measurable, general capability and enabled systematic comparison of retrieval methods across settings.
\begin{wrapfigure}{r}{0.42\textwidth}
\centering
\vspace{-\baselineskip}
\includegraphics[width=0.42\textwidth]{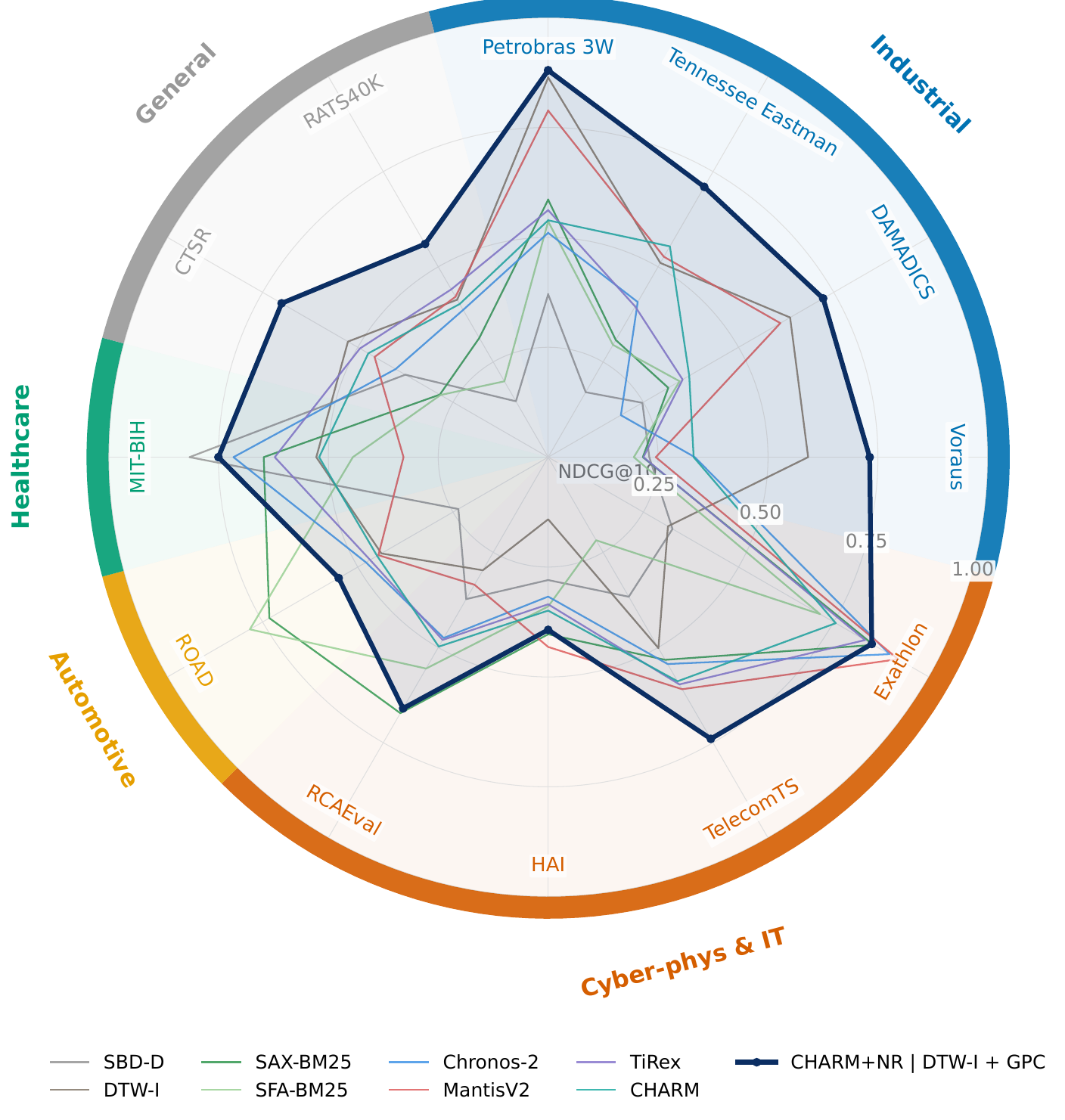}
\caption{Per-dataset NDCG@10 over the 12 READ-Bench datasets (spokes grouped by
diagnostic domain). The base retrievers, coloured by family, form a tight inner
band with no clear winner, while our system (CHARM\,$+$\,NR\,$|$\,DTW-I\,$+$\,GPC)
improves over them on most datasets.}
\label{fig:radar-a}
\vspace{-\baselineskip}
\end{wrapfigure}
Time-series diagnosis presents an analogous problem, but the primary evidence is often structured temporal data. A retrieved historical case can serve as the entry point to its associated metadata and unstructured records, providing richer context for diagnosis. Recent systems approach time-series diagnosis in different ways, including prompting general-purpose LLMs \citep{alnegheimish2024sigllm}, aligning and fine-tuning time-series language models \citep{xie2025chatts,ye2025timera}, and equipping agents with tools and memory \citep{tao2026anomamind}. ARFBench further evaluates multimodal models on diagnostic question answering over real software incidents \citep{xie2026arfbench}. These efforts show rapid progress in reasoning over time-series observations, but retrieving the right historical cases remains a distinct capability. Across these different system designs, retrieval provides a common mechanism for grounding diagnosis in relevant prior experience.

Several lines of work address this retrieval problem. Retrieved examples support forecasting \citep{han2025raft} and anomaly detection \citep{liu2025llmad}. TRACE uses textual context to learn representations for retrieval \citep{chen2025trace}, while comparative studies evaluate representations and unsupervised reranking \citep{rozin2025reranking}. TSRBench~\citep{hu2026tsrbench} provides a common evaluation framework, but its open dataset requires relevant instances to have similar shapes, its diagnostic evaluation is limited to telecom incidents, and it does not test fusion or reranking. Despite this progress, there remains limited work comparing retrieval methods across diagnostic domains, particularly when relevant instances differ in signal shape.

The role of retrieval in this setting is to surface historical cases that are diagnostically relevant to a query. This means cases corresponding to the same underlying fault, anomaly, or event. The central challenge is that diagnostic relevance is not equivalent to signal similarity. The same fault may manifest differently across operating conditions, while similar-looking signals may arise from unrelated causes. In multivariate series, the informative evidence may further be localized to only a few channels or short intervals and obscured by unrelated variation \citep{yeh2017multidimensional}. As with RAG systems, downstream prediction or end-to-end performance alone cannot isolate retrieval quality, since it depends both on which cases are retrieved and on how those cases are subsequently used. Evaluating retrieval therefore requires explicit relevance judgments and controlled analysis of the retrieval pipeline itself, including representation choice, search, reranking, corpus composition, and access to supervision.

Our contributions are:
\begin{itemize}
\item \textbf{READ-Bench (Retrieval Evaluation for Anomaly Diagnosis), a benchmark for diagnostic retrieval.}
We introduce a 12-dataset benchmark spanning five diagnostic domains, with multivariate and complementary univariate evaluations and explicit fault- and event-level relevance judgments.

\item \textbf{A systematic study of retrieval design.}
We compare classical, symbolic, and pretrained retrieval methods across representations, fusion strategies, and reranking methods, and evaluate their robustness to corpus composition and scale.

\item \textbf{A retrieval component for time-series diagnosis.}
Guided by the findings, we combine representation-based search, complementary dynamic-time-warping search, and label-aware Gaussian-process reranking. The resulting component improves NDCG@10 over its underlying search stage on all 12 datasets and outperforms single-method retrievers across the five diagnostic domains (Figure~\ref{fig:radar-a}).
\end{itemize}

\vspace{-0.2cm}
\section{Related Work}
\label{sec:related_work}
\vspace{-0.2cm}

\paragraph{Diagnosis and retrieval augmented systems.}
Retrieving similar historical instances to support time-series fault diagnosis \citep{zhao2017cbr} has been studied before, and recent work extends it with language and multimodal models. ChatTS \citep{xie2025chatts} treats time series as a native input modality for understanding and diagnosis; Time-RA \citep{ye2025timera} frames anomaly analysis as detection, categorization, and diagnosis; ARFBench \citep{xie2026arfbench} evaluates question answering over real software-incident telemetry; and AnomaMind uses tool-augmented agentic reasoning for anomaly detection \citep{tao2026anomamind}. Other systems retrieve historical examples before the downstream task, namely LLMAD for anomaly detection and explanation \citep{liu2025llmad}, retrieval for PHM diagnosis \citep{mizoguchi2023retrieval}, ZARA for language-model reasoning over motion time series \citep{li2026zara}, and retrieval for forecasting \citep{han2025raft,tire2026raf,ning2026ts}. These systems are usually evaluated on the final task, making it difficult to separate retrieval quality from how retrieved instances are later used.

\paragraph{Similarity and representations.}
Time-series retrieval can use several representations and similarity measures. Euclidean distance compares aligned values, dynamic time warping (DTW) \citep{sakoe1978dynamic} allows local temporal alignment, and shape-based distance (SBD) \citep{paparrizos2015k} compares shape using normalized cross-correlation. SAX \citep{lin2007sax} and SFA \citep{schafer2012sfa} convert signals into symbolic sequences searchable with BM25 \citep{robertson2009bm25}. For multivariate series, similarity also depends on how channels are combined \citep{shokoohi2017generalizing,d2025structured}, and irrelevant channels can hide useful patterns \citep{yeh2017multidimensional}. Self-supervised encoders \citep{yue2022ts2vec,franceschi2019tloss,tonekaboni2021tnc} and pretrained time-series models such as CHARM, MantisV2, Chronos-2, and TiRex \citep{dutta2026giving,mantisv2,chronos2,tirex} provide learned representations usable for retrieval, even though search is not their primary training objective.

\paragraph{Retrieval and reranking.}
Several works study time-series retrieval. Deep $r$-RSJBE and DUBCNs
learn supervised and unsupervised representations for multivariate retrieval \citep{song2018deepr,zhu2020dubcns}; CTSR retrieves time series together with metadata \citep{yeh2023ctsr}; and TRACE connects time series with textual context for time-series and cross-modal retrieval \citep{chen2025trace}. Other works study the latter ranking stages by comparing representations and unsupervised reranking \citep{rozin2025reranking}, combining rank information across channels \citep{rozin2026ranked}, or fusing multiple rankings before DTW refinement \citep{barros2026robotretrieval}. 

\paragraph{Benchmarks.}
TS-Haystack evaluates retrieval and reasoning over sparse events within long time-series contexts rather than ranking historical cases from a corpus \citep{zumarraga2026tshaystack}. TSRBench is closer to corpus retrieval, covering similar-series and industrial incident retrieval \citep{hu2026tsrbench}, but its open similar-series setting uses univariate UCR data and ties relevance closely to signal shape, while its multivariate diagnostic setting is limited to a single telecom domain.
\section{READ-Bench}
\label{sec:benchmark}
\label{sec:setup-ref}

\subsection{Problem Formulation}
\label{sec:problem}

\paragraph{Data model.}
The unit of retrieval is a window $x \in \mathbb{R}^{T \times C}$ of $T$
timesteps over $C$ channels, with row $x_t \in \mathbb{R}^{C}$ the reading at
timestep $t$ and column $x^{(c)} \in \mathbb{R}^{T}$ the $c$-th channel. A window
is univariate when $C=1$ and multivariate when $C>1$. Length and channel count
are per-window, so windows live in
$\mathcal{X}=\bigcup_{T,C}\mathbb{R}^{T\times C}$ and a query and a corpus item
need not share either.

\paragraph{Retrieval task.}
Given a corpus of windows $\mathcal{C} = \{x_1, \dots, x_N\}$ and a query window
$q\in\mathcal{X}$, a retrieval method scores each corpus item with
$s(q, \cdot) : \mathcal{C} \to \mathbb{R}$ and ranks $\mathcal{C}$ in descending
order of that score, returning the top-$K$ items. A larger score means greater
estimated relevance, so a distance $d$ enters as $s=-d$. We write $\pi_q$ for
the induced ranking and $\mathrm{rank}(q,x)$ for the position of $x$ in it, and
measure quality by where the relevant items land in $\pi_q$. A second-stage
reranker produces a new ranking of a candidate pool
$\mathcal{P}\subseteq\mathcal{C}$ that a first-stage retriever returns, so it is
scored on $\pi_q$ by the same criterion. The method families that instantiate
$s$ are defined in Section~\ref{sec:methods}.

\paragraph{Relevance.}

Every window carries a single label $y$, either one of the dataset's fault types
or normal $\emptyset$, derived from the window's per-timestep
annotations by the rule of Section~\ref{sec:construction}. An item $x$ is relevant to a
query $q$ when the two share a non-normal fault type,
\begin{wrapfigure}{r}{0.4\textwidth}
\centering
\includegraphics[width=0.4\textwidth]{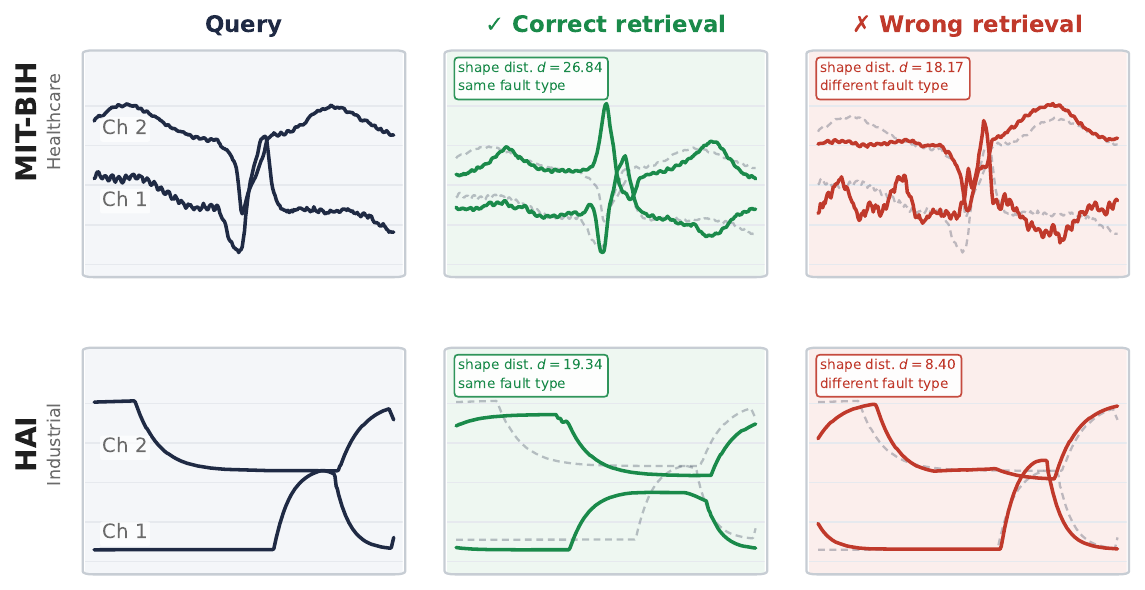}
\caption{The correct retrieval shares the query's fault type but differs in
shape, while a wrong retrieval of a different fault is closer in raw shape, so
relevance is not signal similarity.}
\label{fig:fault-examples}
\end{wrapfigure}
\begin{equation}
  \mathrm{rel}(q, x) \;=\; \mathbb{1}\!\big[\, y_x = y_q \,\wedge\, y_q \neq \emptyset \,\big]
  \;\in\; \{0, 1\},
  \label{eq:relevance}
\end{equation}

so queries are anomalous by construction, corpus anomalies of the query's type
are the retrieval targets, and normal windows are never relevant
(Figure~\ref{fig:fault-examples}). Relevance is
binary and multi-target, so a query typically has several relevant items in the
corpus. These labels define the evaluation target only. The fault label of a
query is never revealed to a retrieval method, and corpus labels are exposed
solely to the methods designated as supervised or label-aware in
Section~\ref{sec:methods}.

\subsection{Benchmark Construction}
\label{sec:construction}
\label{sec:datasets}
\label{sec:querycorpus}

\paragraph{Datasets.}
READ-Bench is built on 12 multivariate datasets drawn from
community and industrial sources that, to our knowledge, have not previously
been curated for retrieval. These are CTSR~\citep{yeh2023ctsr,dau2019ucr},
DAMADICS~\citep{bartys2006damadics}, Exathlon~\citep{jacob2021exathlon},
HAI~\citep{shin2020hai}, MIT-BIH~\citep{moody2001mitbih},
Petrobras~3W~\citep{vargas2019petrobras3w}, RATS40K~\citep{ye2025timera},
RCAEval~\citep{pham2025rcaeval}, ROAD~\citep{verma2024road},
TelecomTS~\citep{feng2025telecomts}, Tennessee Eastman~\citep{downs1993tep}, and
Voraus~\citep{brockmann2023voraus}.
They span condition monitoring, industrial control, microservice and telecom
operations, automotive intrusion, and physiological signals, with fault
signatures that differ markedly across and within domains
(Figure~\ref{fig:fault-strip} and Figure~\ref{fig:anomalies}).
Channel count ranges
from a single channel to 664, window length from 16 to 1024 timesteps, and
the number of fault classes per dataset from four to ninety-four, with per-class
corpus sizes spanning three orders of magnitude and a max-to-min class ratio
that reaches $342{:}1$ on the most skewed dataset
(Appendix~\ref{app:querycorpus}). Prior time-series retrieval evaluations are
built on a single univariate archive with balanced classes and no channel axis,
whereas this variation across 12 independent sources keeps a method from
succeeding by overfitting one domain. The per-dataset composition is given in
Table~\ref{tab:corpus-sizes}.

\paragraph{Windowing and labeling.}
We respect each dataset's existing conventions, keeping the windows and labels
of any dataset that ships pre-windowed and pre-labeled, and otherwise segmenting each series into fixed-length windows. Datasets provide per-timestep labels
$\ell_1, \dots, \ell_T$ over a fault vocabulary in which $0$ is the normal state,
and where a dataset codes a fault's transient onset separately from its steady
phase we merge the two so each fault carries one label. A window takes a fault label only when that fault holds a strict majority of its
timesteps, with ties broken toward the lower-indexed label so the normal state
is preferred; that is, $y=c$ when $|\{t:\ell_t=c\}|>T/2$ for some fault $c$, and
$y=\emptyset$ (normal) otherwise. Because a fault shorter than the window can
never hold this global majority, we relax the rule so that a fault also claims
the window when it fills a majority of its own extent within the window.
Writing $e_c=[\min\{t:\ell_t=c\},\,\max\{t:\ell_t=c\}]$ for the span from a fault
$c$'s first to last occurrence in the window and $|e_c|$ for its length, the
label is
\begin{equation}
  y =
  \begin{cases}
    c, & \text{if } \big|\{\, t : \ell_{t} = c \,\}\big| > T/2 \text{ for some fault } c,\\[2pt]
    c, & \text{else if } \big|\{\, t\in e_c : \ell_{t} = c \,\}\big| > |e_c|/2 \text{ for some fault } c,\\[2pt]
    \emptyset, & \text{otherwise,}
  \end{cases}
  \label{eq:windowlabel}
\end{equation}
with $\emptyset$ denoting a normal window. The relaxation recovers short faults
without changing any window a global majority already decides.
A handful of datasets label at a coarser
granularity that the data dictates, such as a heartbeat window taking the
majority class of the beats it spans, or a robot operation whose fault flag is
constant over the recording. Appendix~\ref{app:labeling} documents how each
dataset's native annotations yield its fault-type labels.

\paragraph{Split protocol.}
For every dataset we draw queries from the test split and build the corpus from
the train split, so no query window appears in the corpus it is
ranked against. We respect each dataset's native split convention where one
exists and otherwise split at a grouping coarser than the window (per fault
instance, recording, entity, run, or patient), so train and test stay disjoint;
per-dataset split units are listed in Table~\ref{tab:labeling}. When a split must
be subsampled to the target size, we sample
anomalies stratified by fault type so the native class imbalance is preserved
rather than flattened, and draw normal windows uniformly at random. All splits
use a fixed seed, and all headline results in
Section~\ref{sec:experiments} are macro-averaged over the 12 datasets.

\paragraph{Corpus composition.}
The corpus a query is ranked against pairs its relevant set, the same-type
anomalies of Equation~\ref{eq:relevance}, with distractors of two kinds.
Anomalies of a different fault type probe whether a method separates fault types
rather than merely detecting abnormality, and normal windows are the easy
negatives that dominate any real repository. We control the second kind
explicitly, drawing normal windows from the training split to report three
pollution levels $\rho$, the fraction of the corpus that is normal, at roughly
$\rho{=}0\%$, $10\%$, and $20\%$, so robustness to easy negatives can be read off
directly. Because fault types are class-imbalanced, the number of relevant items
per query varies widely, and the metrics of Section~\ref{sec:evalprotocol} are
chosen to remain meaningful under that imbalance.

\subsection{Methods}
\label{sec:methods}

READ-Bench organizes the retrieval pipeline into three stages, namely a base retriever that ranks the full corpus, fusion that combines complementary
rankings, and a reranker that refines a short candidate pool. We also include
fully supervised classifiers as a reference ceiling. The methods therefore
differ along two axes, namely the representation used to compare windows and the amount of label information available at ranking time. Unless stated
otherwise, methods are label-free. Exact formulations and implementation details
are given in Appendix~\ref{app:baseline_math}.

\paragraph{Base retrievers.} All base retrievers instantiate the scoring function of Section~\ref{sec:problem}, but encode similarity differently. Raw-window distances compare the series directly, where Euclidean distance is lock-step, SBD uses shape-aligned cross-correlation, and DTW allows elastic alignment, either independently across channels (DTW-I) or under a shared alignment (DTW-D). Symbolic retrievers first discretize each channel and then retrieve with BM25, using either SAX value symbols or SFA frequency symbols. To compare windows of differing length or channel count, the lock-step distance and the symbolic retrievers reconcile a pair onto a common grid (zero-padding channels and interpolating length), the elastic and cross-correlation distances align unequal lengths natively, and the embedders mean-pool over channels and resample length to the encoder's admissible grid (Appendix~\ref{app:baseline_math}). Foundation-model (FM) embedders---CHARM,
Chronos-2, MantisV2, and
TiRex---map each window to a fixed-dimensional embedding and rank corpus items in that space. We use Euclidean distance for the base embedding score; cosine produces nearly identical rankings in our setting, consistent with prior time-series retrieval benchmarks.
.

Absolute embedding distance can, however, be dominated by the operating regime shared by normal and faulty windows, while diagnostic relevance depends on how a window departs from normal operation. The Normal-Residual (NR) scorer changes this reference point without changing the encoder, keeping only the part of a window's embedding that its normal expectation leaves unexplained. Writing $\hat u = u/\|u\|$ for $\ell_2$ normalization, let $\mathbb{E}[\hat\phi(u)\mid \mathrm{normal}]$ be the embedding we would expect for $u$ if it were normal; the residual is
\begin{equation}
r(u)=\hat\phi(u)-\mathbb{E}\left[\hat\phi(u)\mid \mathrm{normal}\right],
\qquad
s_{\mathrm{NR}}(q,x)= \hat r(q)^{\top}\,\hat r(x).
\label{eq:nr}
\end{equation}
We approximate this conditional expectation by a $k$-nearest-neighbour estimate over a train-split pool $\mathcal{N}$ of normal windows: $\mathbb{E}[\hat\phi(u)\mid \mathrm{normal}]\approx\tfrac{1}{k}\sum_{x\in\mathcal{N}_k(u)}\hat\phi(x)$, the mean of the $k$ normal embeddings nearest $\hat\phi(u)$ in cosine similarity (we use $k{=}30$). Averaging a small neighbourhood rather than the single nearest normal is a variance-reduction step, since one nearest normal injects its own embedding noise into every residual, whereas the local mean cancels that noise while staying specific to the window's operating point. NR then compares residual directions by cosine, so $r(u)$ is what remains of a window after removing its expected-if-normal component, and NR measures whether the query and a candidate depart from normal
in a similar direction. It requires normal windows but no fault labels and applies to any
fixed embedder $\phi$; the per-embedder factorial is reported in
Table~\ref{tab:app-nr}.

\paragraph{Supervised reference.}
To estimate how much fault-type structure is recoverable with full label
access, we also train MR-Hydra~\citep{dempster2023hydra} and
RDST~\citep{rdst}, both strong recent time-series classifiers~\citep{middlehurst2024bakeoff}, on the labeled corpus. Their class posteriors are converted
to rankings as detailed in Appendix~\ref{app:baseline_math}. Because these
models are trained as supervised classifiers rather than retrieval systems,
we treat them as a reference ceiling rather than as directly comparable
retrievers.

\paragraph{Fusion.}
Learned embeddings and shape or symbolic retrievers encode different notions of
similarity, so we test whether their rankings contain complementary relevant
items. This mirrors text retrieval, where fusing a lexical retriever such as
BM25 with a dense one recovers relevant documents
that neither leg ranks alone~\citep{karpukhin2020dpr,formal2021splade}; our
symbolic BM25 legs and embedding retrievers play the analogous sparse and dense
roles here. Each fusion pairs one embedder with one shape or symbolic
retriever, and combines them with one of three standard rank- or score-level
schemes, reciprocal-rank fusion (RRF)~\citep{cormack2009rrf}, a $z$-scored
weighted sum, and a two-stage cascade, none of which requires the two legs to
share a score scale. RRF is our default, since it needs no comparable scores and performs best
here (Section~\ref{sec:additions}). The exact combination rules are standard and
are given in Appendix~\ref{app:baseline_math}; the full embedding$\times$shape/symbolic
grid is evaluated in Section~\ref{sec:additions}.

\paragraph{Rerankers.}
Modern retrieval systems are two-stage, where a cheap first stage returns a
candidate pool that a stronger reranker then re-orders~\citep{thakur2021beir,
khattab2020colbert}. A reranker receives the candidate pool $\mathcal{P}$
returned by a base or fused retriever and changes only its ordering. We avoid
training a global relevance model on benchmark labels and instead study three
forms of additional information. Pseudo-relevance feedback (PRF)~\citep{rocchio1971prf} is label-free and
re-scores candidates relative to a pseudo-relevance centroid. Label-aware
rerankers use only a small, query-local set of candidate labels. Purity
rewards candidates whose local neighbourhood is label-consistent,
Majority-Vote infers the query class from nearby labeled items, and
query-conditioned Purity (QPurity) combines the two. This setting models a
diagnostic repository in which some resolved incidents have confirmed labels
without assuming labels for the full corpus
\citep{yu2024amad,del2021active}.

The Gaussian-process classifier (GPC) reranker~\citep{rasmussen2006gpml}
uses the same local supervision but replaces voting with a learned local
decision boundary. It fits a GPC on the labeled candidates in $\mathcal{P}$
and ranks a candidate $x$ by the probability that it shares the query's
inferred class:
\begin{equation}
s_{\mathrm{GPC}}(q,x)
=
\Pr\!\left(y_x = y_q \mid q,x,\mathcal{P}\right),
\label{eq:gpc}
\end{equation}
where the query label $y_q$ is latent and is never provided to the reranker.
GPC is deliberately a cheap, domain-agnostic reranker, since it operates on the base
representation with no per-dataset tuning and adds only a fraction of a second
per query, in contrast to language-model rerankers~\citep{sun2023rankgpt} that
read the raw series through a hosted model at far greater cost.

Finally, we test language-model rerankers that read the retrieved series
directly, namely Toto-1.0-QA~\citep{toto}, ChatTS, Claude,
and Claude\,$+$\,TSAD following LLMAD. Together, these
methods let us separate gains from the base representation, complementary
retrieval signals, and additional information introduced only at reranking
time.

\subsection{Evaluation Protocol}
\label{sec:evalprotocol}

\paragraph{Metrics.}
We report precision at $K$ (P@$K$), hit rate at $K$ (HR@$K$), and normalized
discounted cumulative gain at $K$ (NDCG@$K$). P@$K$ and HR@$K$ measure the
precision of the retrieved set without regard to order, namely the
fraction of the top-$K$ that is relevant and whether any relevant item appears
in the top-$K$, while NDCG@$K$ additionally rewards ranking relevant items
above irrelevant ones. We report all three because a method can satisfy one
axis while failing the other, for instance by retrieving every relevant item
but ranking it last. Under the binary, multi-target relevance of
Equation~\ref{eq:relevance} the ideal ranking places all relevant items first,
so NDCG@$K$ is normalized against a query-specific ideal that accounts for the
varying number of relevant items. Formal definitions of all three metrics are
given in Appendix~\ref{app:metrics}.

\paragraph{Significance testing.}
Because averaged metric differences can arise from noise rather than a genuine
gap, we compare methods with a significance test rather than raw score
differences. We use a Friedman test with a Nemenyi post-hoc test, blocking on
the 12 datasets, so that a difference is reported as significant only when the
methods' mean ranks differ by more than the Nemenyi critical difference. The
test statistic and threshold are given in Appendix~\ref{app:stats}.

\paragraph{Per-dataset breakdown.}
Beyond the pooled tests we report per-dataset average metrics as a descriptive
robustness check across domains. We treat this as a diagnostic rather than a
formal test, since 12 datasets are too few to serve as a reliable blocking
unit for hypothesis testing. The full per-dataset tables appear in
Appendix~\ref{app:alldata}, and the macro-averaged grids in
Appendix~\ref{app:full-results}.

\section{Experiments and Analysis}
\label{sec:experiments}

With the task, methods, and protocol fixed in Section~\ref{sec:setup-ref}, we
ask what retrieves the right cases, whether fusion and reranking improve those
rankings, and how far this reaches toward full supervision. We proceed in the
order the pipeline is built. We start with the
label-free base retrievers, add NR scoring and test its robustness
to corpus pollution, then treat fusion and reranking as separate additions, and
finally combine them. The results tell a consistent story. The base retrievers
land on a tight plateau that NR scoring lifts only modestly, while
a small amount of label information lifts it decisively. Fusion helps only
selectively, and language-model reranking does not help at all. Headline numbers are NDCG@10
macro-averaged over the 12 datasets, reported with the significance test of
Section~\ref{sec:evalprotocol}. Embedders are scored by Euclidean distance
unless stated otherwise.\footnote{We name a composed retriever by its parts, so a
bare embedder name (for example CHARM) denotes that default Euclidean score, an
embedder with normal-residual scoring appends the residual marker (for example
CHARM\,$+$\,NR), a fused leg joins with a bar (CHARM\,$+$\,NR\,$|$\,DTW-I), and a
reranker joins with a plus (CHARM\,$+$\,NR\,$|$\,DTW-I\,$+$\,GPC).}

\subsection{Base retrievers}
\label{sec:base}

We evaluate the label-free retrievers of Section~\ref{sec:setup-ref}, namely the
raw-window distances, the symbolic BM25 retrievers, and the frozen
foundation-model embedders. Table~\ref{tab:backbones} reports their NDCG@10 next
to the supervised reference, on each of the three pollution levels $\rho$ of
Section~\ref{sec:construction}, the fraction of the corpus made up of normal
series ($0\%$, $\approx\!10\%$, $\approx\!20\%$). To keep runtime and memory
comparable across datasets, each corpus is capped to a common target size (the
per-dataset sizes are in Table~\ref{tab:corpus-sizes}); the remaining metrics
(P@$\{1,5\}$, HR@10, NDCG@$\{5,20\}$, and macro-F1)
are in Appendix~\ref{app:backbone-grid}. The cap does not change the
conclusion below, because for the methods that scale to the full uncapped corpus
the scores barely move (Appendix~\ref{app:fullcorpus}).

All of the label-free retrievers score far below the supervised reference, and
they score close to one another (Figure~\ref{fig:box}). The significance test
separates only the weakest of them, SBD-D, from the rest; the top four
(MantisV2, SAX-BM25, CHARM, TiRex) are statistically indistinguishable. The
differences among the leaders are smaller than the variation from one dataset to
the next, so the choice among a pretrained embedding, a classical distance, and
a symbolic retriever does not change retrieval quality. The gap to the
supervised reference, in contrast, is large and statistically clear. The rest of
the section asks which additions can lift a retriever above this level, starting
with NR scoring.

\begin{table}[t]
\centering
\scriptsize
\setlength{\tabcolsep}{3pt}
\renewcommand{\arraystretch}{0.85}
\caption{NDCG@10 of every base retriever, our best composed system, and the
supervised reference, macro-averaged over the 12 datasets at each pollution
level $\rho$, the fraction of the corpus diluted with normal-class series
($0\%/\!\approx\!10\%/\!\approx\!20\%$). The supervised rows are a classification
reference rather than retrieval systems. In each column the best value is bold
and the second best underlined, over the base retrievers and our system (the
supervised reference excluded).}
\label{tab:backbones}
\resizebox{0.6\textwidth}{!}{%
\begin{tabular}{llccc}
\toprule
Method & Family & $\rho{=}0\%$ & $\rho{=}10\%$ & $\rho{=}20\%$ \\
\midrule
Random & Trivial floor & 0.189 & 0.168 & 0.155 \\
Majority &  & 0.246 & 0.242 & 0.166 \\
\addlinespace
SBD-D & Raw-window distance & 0.328 & 0.271 & 0.245 \\
DTW-I &  & 0.480 & 0.414 & 0.395 \\
\addlinespace
SAX-BM25 & Symbolic & 0.489 & 0.457 & 0.431 \\
SFA-BM25 &  & 0.409 & 0.386 & 0.376 \\
\addlinespace
Chronos-2 & FM embedder & 0.471 & 0.455 & 0.432 \\
MantisV2 &  & \underline{0.510} & \underline{0.474} & \underline{0.454} \\
TiRex &  & 0.483 & 0.449 & 0.433 \\
CHARM &  & 0.486 & 0.461 & 0.437 \\
\midrule
CHARM\,$+$\,NR\,$|$\,DTW-I\,$+$\,GPC & Our best system & \textbf{0.687} & \textbf{0.640} & \textbf{0.619} \\
\midrule
MR-Hydra & Supervised reference & 0.776 & 0.672 & 0.658 \\
RDST &  & 0.711 & 0.634 & 0.590 \\
\bottomrule
\end{tabular}%
}
\end{table}

\vspace{-0.1cm}
\subsection{Normal-residual scoring and robustness to pollution}
\label{sec:noise}
\vspace{-0.1cm}

Representation choice starts to matter once the corpus is polluted with normal
series (the pollution levels of Table~\ref{tab:backbones}). We compare every
base retriever against the NR
scorer of Section~\ref{sec:methods}, which subtracts each window's
expected-if-normal embedding so that similarity reflects departure from normal
operation.

\begin{wrapfigure}{r}{0.46\textwidth}
\centering
\vspace{-\baselineskip}
\includegraphics[width=0.46\textwidth]{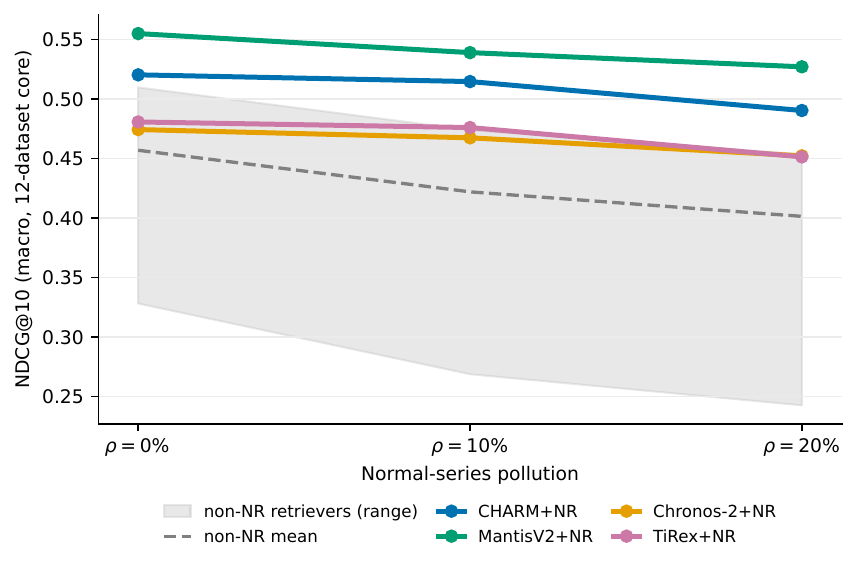}
\caption{The NR embedder variants against the non-NR field as the corpus
is diluted with normal series (NDCG@10, macro-averaged over the 12
datasets). The shaded band spans the non-NR retrievers (raw distances, symbolic
BM25, and ED embedders) with their mean dashed; the four NR embedder
variants (bold) sit above it and stay high at every pollution level while the
field degrades. MantisV2\,$+$\,NR is the strongest base retriever throughout.}
\label{fig:robust}
\vspace{-\baselineskip}
\end{wrapfigure}

Pollution degrades every base retriever, but by very different amounts. The raw
distances fall fastest, while NR is both the most accurate base retriever and the most
robust. It lifts clean accuracy on every embedder it wraps except TiRex, which
is essentially unchanged (the largest gains are on CHARM and
MantisV2), and roughly halves the degradation under pollution, so the NR
retrievers stay well above the raw distances, symbolic retrievers, and ED
embedders as the normal fraction grows (Figure~\ref{fig:robust}). MantisV2\,$+$\,NR
is the winner on both counts, the strongest single base retriever clean and the
one that gives up the least under pollution (the per-embedder sweep is
in Appendix~\ref{app:nr-by-embedder}).

\vspace{-0.1cm}
\subsection{Fusion and reranking}
\label{sec:additions}
\vspace{-0.1cm}
\begin{wraptable}{r}{0.42\textwidth}
\centering
\scriptsize
\setlength{\tabcolsep}{4pt}
\renewcommand{\arraystretch}{0.9}
\caption{Language-model methods on the CHARM pool (NDCG@10). The first block
reranks the retrieved pool and the second classifies the fault type with no
retrieval. Best per block bold, second best underlined.}
\label{tab:llm}
\begin{tabular}{lc}
\toprule
Method & NDCG@10 \\
\midrule
CHARM (base) & 0.486 \\
\midrule
\multicolumn{2}{l}{\emph{Reranking the CHARM pool}} \\
Toto-1.0-QA & \underline{0.493} \\
ChatTS & 0.475 \\
Claude & \textbf{0.496} \\
Claude\,$+$\,TSAD & 0.492 \\
\midrule
\multicolumn{2}{l}{\emph{Classification, no retrieval}} \\
Toto-1.0-QA & 0.159 \\
ChatTS & 0.169 \\
Claude & \underline{0.180} \\
Claude\,$+$\,TSAD & \textbf{0.184} \\
\bottomrule
\end{tabular}
\vspace{-\baselineskip}
\end{wraptable}

Two additions can improve a base retriever. Fusion combines it with a
complementary shape or symbolic leg, and reranking re-orders its candidate pool
using extra information. We report each here, then combine them in
Section~\ref{sec:stacking}.

\paragraph{Fusion.}
Because the base retrievers are close, we test whether pairing an embedder with a
complementary leg helps, fusing by RRF, which needs no
comparable scores and outperforms weighted-sum and two-stage fusion. Fusion
gives a small, mostly on-plateau gain, the best leg for every embedder being
DTW-I, and the gain is consistent across embedders rather than specific to one
(Appendix~\ref{app:fusion-choice}). Fusion alone does not leave the plateau, but a
fused pool is a stronger starting point than either leg alone and reappears as a
building block later.

\paragraph{Reranking.}

Reranking re-orders a fixed pool with the methods of
Section~\ref{sec:setup-ref}, grouped by the label information they use (none,
neighbour labels, or a language model). We apply the rerankers to a
representative set of pools rather than every one, namely the strongest raw distance
(DTW-I), the four foundation-model embedders, and their NR variants; the weaker
distances (SBD-D, DTW-D) and symbolic retrievers add little as pools and are
omitted. We use the language model two ways,
reranking the retrieved pool and, as a no-retrieval baseline, classifying the
anomaly type directly from the series. Appendix~\ref{app:reranking} collects the
statistical rerankers and Table~\ref{tab:llm} the language-model results.
Three tiers emerge (visualised in Figure~\ref{fig:lever}; full grids
across pollution levels in Appendix~\ref{app:reranker-grid}, and the
language-model rerankers in Table~\ref{tab:llm}). Label-aware
reranking is the only addition that reliably helps, since a GPC
reranker improves every pool it is applied to, with a significant gain on every
pool (Appendix~\ref{app:reranking}), and narrows the gap to the supervised reference
without closing it. Label-free reranking hurts most
pools, so we drop it from later use. Language-model reranking of the fixed CHARM
pool barely moves NDCG@10 across all four models (Toto-1.0-QA,
ChatTS, Claude, Claude\,$+$\,TSAD), an order of magnitude less than the statistical
reranker, and the same models asked to name the type without retrieval
score near the random floor. We use CHARM here for budget and because its low
cross-dataset variance (Appendix~\ref{app:backbone-grid}) makes it a stable pool
on which any real effect should be easiest to see, so the pool choice is not to
blame. On this task, then, retrieval is what makes the language model useful, its
reasoning over the raw series adds little, while a small amount of label
information at rerank time adds the most.

\vspace{-0.2cm}
\subsection{Combining fusion and reranking}
\label{sec:stacking}
\vspace{-0.1cm}

\begin{wrapfigure}{r}{0.46\textwidth}
\centering
\vspace{-\baselineskip}
\includegraphics[width=0.46\textwidth]{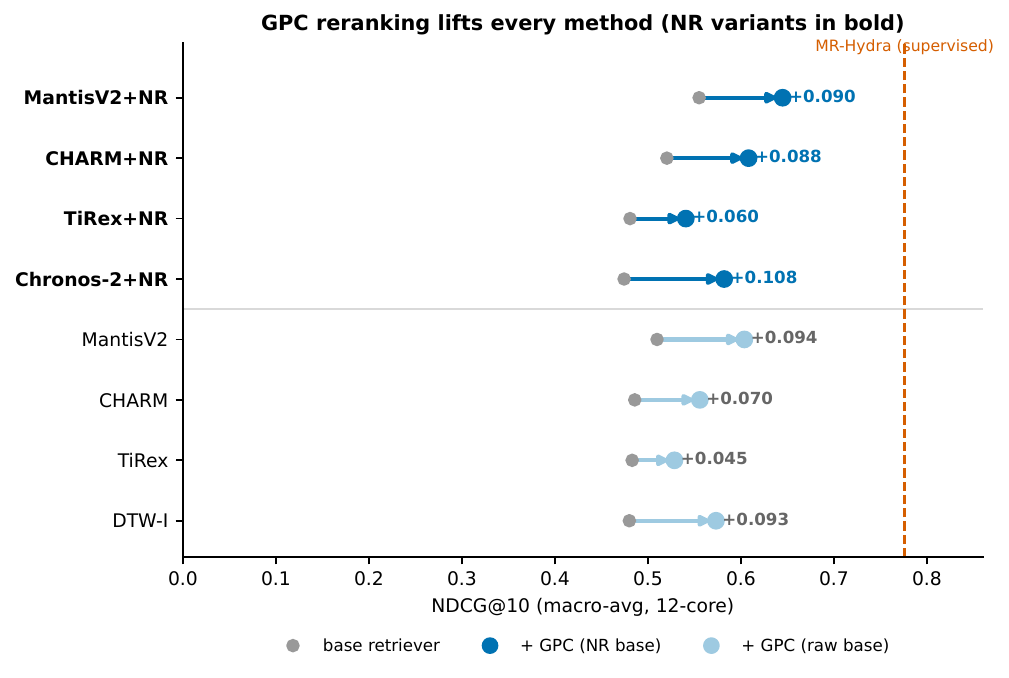}
\caption{Effect of GPC reranking across base retrievers (NDCG@10, macro-averaged over
the 12 datasets), the full grid is in Table~\ref{tab:app-rerankers}.
Each arrow runs from a base retriever to its GPC-reranked version, and the
label-aware reranker gives a consistent gain on every base retriever. The dashed line
marks the supervised reference (MR-Hydra), which no single-pool system reaches.}
\label{fig:lever}
\vspace{-\baselineskip}
\end{wrapfigure}

The preceding subsections added one component at a time, either a fused shape
leg or a label-aware reranker. Because the two draw on different signals, namely
a second view of the series and a few neighbour labels, we now combine them,
fusing a base retriever with its best leg (DTW-I) and reranking the fused pool
with GPC at each pollution level (Appendix~\ref{app:composition}). We build
these stacks on the NR variants and include the plain ED embedders as a non-NR
reference.

The two additions are close to additive, since reranking a fused pool adds about
the same margin it adds to an unfused one. The best combined system,
CHARM\,$+$\,NR\,$|$\,DTW-I\,$+$\,GPC, reaches NDCG@10 of 0.687/0.640/0.619
at $\rho{=}0/10/20\%$ and is the strongest at every level. Its per-dataset
envelope sits above the individual retrievers on most datasets and across all
five diagnostic domains (Figure~\ref{fig:radar-a}). Normal-residual scoring is
what carries it clear of the plain-embedding stacks, as
CHARM\,$|$\,DTW-I\,$+$\,GPC reaches only 0.631, while fusion contributes a
smaller final increment on top of the base representation and the reranker.

Our combined method runs an NR-scored foundation embedder, fuses in a DTW-I leg,
and reranks with GPC, none of which needs task-specific training, and it remains
inexpensive. GPC adds only a fraction of a second per query, so the combined
systems run far below the supervised classifiers while closing much of the gap
to them (Figure~\ref{fig:runtime}).

\vspace{-0.3cm}
\section{Conclusion}
\label{sec:conclusion}
\vspace{-0.2cm}

We introduced READ-Bench, a rigorous framework for evaluating time-series anomaly diagnosis through information retrieval, addressing the gap between raw anomaly detection and actionable root-cause localization across 12 diverse industrial domains. Our experiments yield three main conclusions. First, pretrained embedding models show no advantage over strong classical and symbolic baselines, both falling well below a supervised reference. Second, leveraging additional information, specifically small amounts of label data or normal periods, serves as a decisive lever; non-parametric reranking and Gaussian-process reranking improve whichever retrieval pool is strongest, proving to be the most reliable additions even under normal-series pollution. Third, score fusion helps only in select settings, and language-model reranking of a fixed pool shows almost no detectable effect, indicating that LLMs provide almost no benefit in this setting.

Our evaluation is bounded by the original labeling fidelity and categorization granularity of the source repositories, which may occasionally obscure subtle cascading failures or multi-fault interactions, alongside choices in labeling and window strategies that naturally influence conclusions. Looking beyond static retrieval, our immediate roadmap centers on evolving READ-Bench into a dynamic, multi-turn QA environment that mirrors real-world diagnostic workflows. Ideal diagnostic usage requires interactive troubleshooting where a system must inspect telemetry windows, consult documentation, and iteratively refine its conclusions. To support this, we aim to explore coding agents, the ideation of possible reasons for failures, and robust uncertainty quantification.


\section*{Ethics Statement}
READ-Bench is assembled entirely from publicly available time-series datasets,
released by their originators for research use; we redistribute only curated
queries, corpora, and relevance judgments derived from them, and retain each
source's original license and attribution. No new data were collected from human
subjects, and the physiological source (MIT-BIH) is already de-identified by its
providers. The benchmark is intended to measure how well retrieval methods
recover relevant historical cases for time-series diagnosis, a step that can
support human operators rather than replace their judgment; as with any
diagnostic tool, retrieved evidence should be reviewed by a qualified expert
before it informs a decision. We are not aware of harmful dual-use risks beyond
those already present in the underlying public datasets.

\section*{AI Use Statement}
We used generative AI tools solely to refine and polish the writing of this
paper, namely to improve readability, tighten phrasing, and suggest a title and
keywords. We did not use generative AI tools to generate synthetic data, develop
theoretical models or conceptual frameworks, formulate mathematical claims or
proofs, propose or refine hypotheses, design experiments or methodology,
implement methods, clean or reformat datasets, or interpret results; these were
carried out by the authors. We have reviewed all AI-assisted text, and we take
responsibility for the final content of this work, including all text, claims,
and artifacts.

\section*{Reproducibility Statement}
We designed READ-Bench for reproducibility. The task, relevance definition, and
windowing rules are specified in Section~\ref{sec:benchmark}, with per-dataset
curation in Appendix~\ref{app:curation} and the labeling and split protocols in
Appendix~\ref{app:labeling}. Formal
definitions of every method are given in
Appendix~\ref{app:baseline_math}, the evaluation metrics in
Appendix~\ref{app:metrics}, and the significance-testing procedure in
Appendix~\ref{app:stats}; hyperparameters and the compute budget are listed in
Appendix~\ref{app:hyperparams}. All splits and subsampling use a fixed seed
(Section~\ref{sec:construction}). The benchmark, including the curated corpora,
queries, and relevance judgments, is publicly available as a versioned dataset
on the Hugging Face Hub, with a public leaderboard and submission protocol at
\url{https://huggingface.co/spaces/c3aiia3c/read-bench-leaderboard}; the code for
building corpora, running the retrieval methods, and reproducing every table and
figure will be released.

\bibliography{iclr2027_conference}
\bibliographystyle{iclr2027_conference}
\newpage
\appendix
\section{Appendix}

\localtableofcontents
\vspace{1em}

\subsection{Dataset Details}
\label{app:datasets}

This appendix details the twelve datasets of READ-Bench
(Section~\ref{sec:construction}), covering how they were curated, how queries
and corpora are constructed from them, and how each dataset's native
annotations are turned into fault-type labels.

\subsubsection{Curation protocol}
\label{app:curation}
Each dataset enters the benchmark through the same pipeline. We start from a
public source, retain the channels and label vocabulary of the originators, and
keep the windows of any dataset that ships pre-windowed, otherwise segmenting
each series into fixed-length windows sized to its sampling rate. Each window
takes one fault-type label by Equation~\ref{eq:windowlabel}. Missing values are
cleaned within each series before windowing, never across a fault boundary, and
by one of two source conventions. Most datasets forward-fill a gap with the last
valid reading, back-fill a leading gap, and set a wholly-absent channel to zero,
since a missing sample means no fresh reading rather than a true zero.
Petrobras~3W instead follows its upstream drop
convention, coercing out-of-range sensor sentinels to missing and dropping the
affected rows without interpolating them.

CTSR is the one univariate dataset we retain, derived from
the UCR archive~\citep{dau2019ucr}. Its classes are not anomalies, but
diagnostic relevance in the multivariate datasets often concentrates in a subset
of channels, so the single-channel case is a close analog and provides a
well-established, shape-driven reference point alongside them. Because CTSR has
far more relevance classes than the query budget, its corpus is drawn to keep at
least ten same-class entries per query, so scores stay stable rather than
resting on a single relevant item.

\subsubsection{Query and corpus construction}
\label{app:querycorpus}
For every dataset we draw queries from the test split and build the retrieval
corpus from the train split. Queries are anomalous windows; the
corpus is either anomalies alone (the clean setting) or anomalies mixed with a
controlled fraction of normal windows drawn from the training split. We report
three pollution levels, roughly $0\%$, $10\%$, and $20\%$ normal windows, so
that robustness to easy negatives can be read off directly. All splits use a
fixed seed for reproducibility, and the split unit of Table~\ref{tab:labeling}
is always coarser than a window, so no query window, or a near-duplicate of
one, enters the corpus it is ranked against. Table~\ref{tab:corpus-sizes} lists
the resulting corpus and query sizes.

\begin{table}[H]
\centering
\scriptsize
\setlength{\tabcolsep}{2.5pt}
\renewcommand{\arraystretch}{0.9}
\caption{Per-dataset composition of READ-Bench. $C$ is the channel count, $T$
the window length, and ``classes'' the number of non-normal fault types.
$|\mathcal{C}|$ is the corpus size at the three pollution levels
($0/10/20\%$ normal windows), $|\mathcal{Q}|$ the query count, and
``rel$/q$'' the mean number of same-class corpus entries per query at
$\rho{=}0\%$. ``Imb.'' is the class-imbalance ratio (most frequent over least frequent
fault class, at $\rho{=}0\%$).}
\label{tab:corpus-sizes}
\begin{tabular}{@{}lrrrrrrr@{}}
\toprule
Dataset & $C$ & $T$ & Classes & $|\mathcal{C}|$ ($0/10/20\%$) & $|\mathcal{Q}|$ & rel$/q$ & Imb. \\
\midrule
CTSR~\citep{yeh2023ctsr}              &   1 &  512 & 94 & 1000/1111/1250 & 100 & 10.6 & 1.2 \\
DAMADICS~\citep{bartys2006damadics}          &  32 &  256 &  4 & 1000/1111/1250 &  18 & 308.8 & 83.9 \\
Exathlon~\citep{jacob2021exathlon}          &  51 &  256 &  6 & 1000/1111/1250 & 100 & 215.9 & 13.2 \\
HAI~\citep{shin2020hai}               &  79 &  256 &  6 & 305/339/381    &  28 & 60.8 & 10.0 \\
MIT-BIH~\citep{moody2001mitbih}           &   2 &  256 &  4 & 1000/1111/1250 & 100 & 336.9 & 10.7 \\
Petrobras 3W~\citep{vargas2019petrobras3w}      & 2--7 &  256 &  9 & 1000/1111/1250 & 100 & 146.8 & 23.7 \\
RATS40K~\citep{ye2025timera}           & 1--9 & 16--128 & 20 & 1000/1111/1250 & 100 & 151.2 & 342.0 \\
RCAEval~\citep{pham2025rcaeval}           & 429 &  256 &  5 & 700/778/875    &  50 & 140.0 & 1.0 \\
ROAD~\citep{verma2024road}              & 664 &  256 &  6 & 71/79/89       &  25 & 22.4 & 37.0 \\
TelecomTS~\citep{feng2025telecomts}         &  18 &  128 & 11 & 977/1086/1221  & 100 & 117.8 & 5.3 \\
Tennessee Eastman~\citep{downs1993tep} &  52 &  256 & 20 & 1000/1111/1250 & 100 & 50.0 & 1.0 \\
Voraus~\citep{brockmann2023voraus}            & 130 & 1024 & 12 & 528/587/660    & 100 & 68.9 & 15.6 \\
\bottomrule
\end{tabular}
\end{table}

\begin{figure}[t]
\centering
\includegraphics[width=0.72\columnwidth]{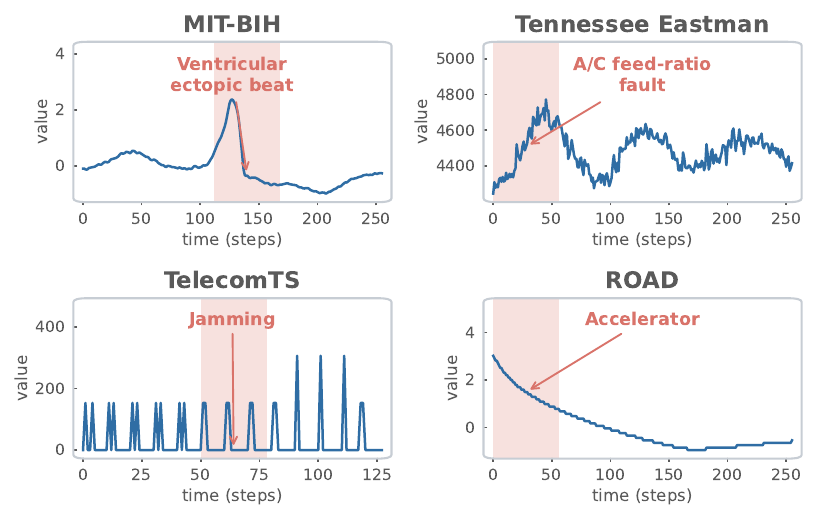}
\caption{Example fault signatures across four READ-Bench domains, one per
panel, with the anomalous span shaded and the most informative channel shown.}
\label{fig:fault-strip}
\end{figure}

\begin{figure}[p]
\centering
\rotatebox{90}{%
\includegraphics[width=0.86\textheight,keepaspectratio]{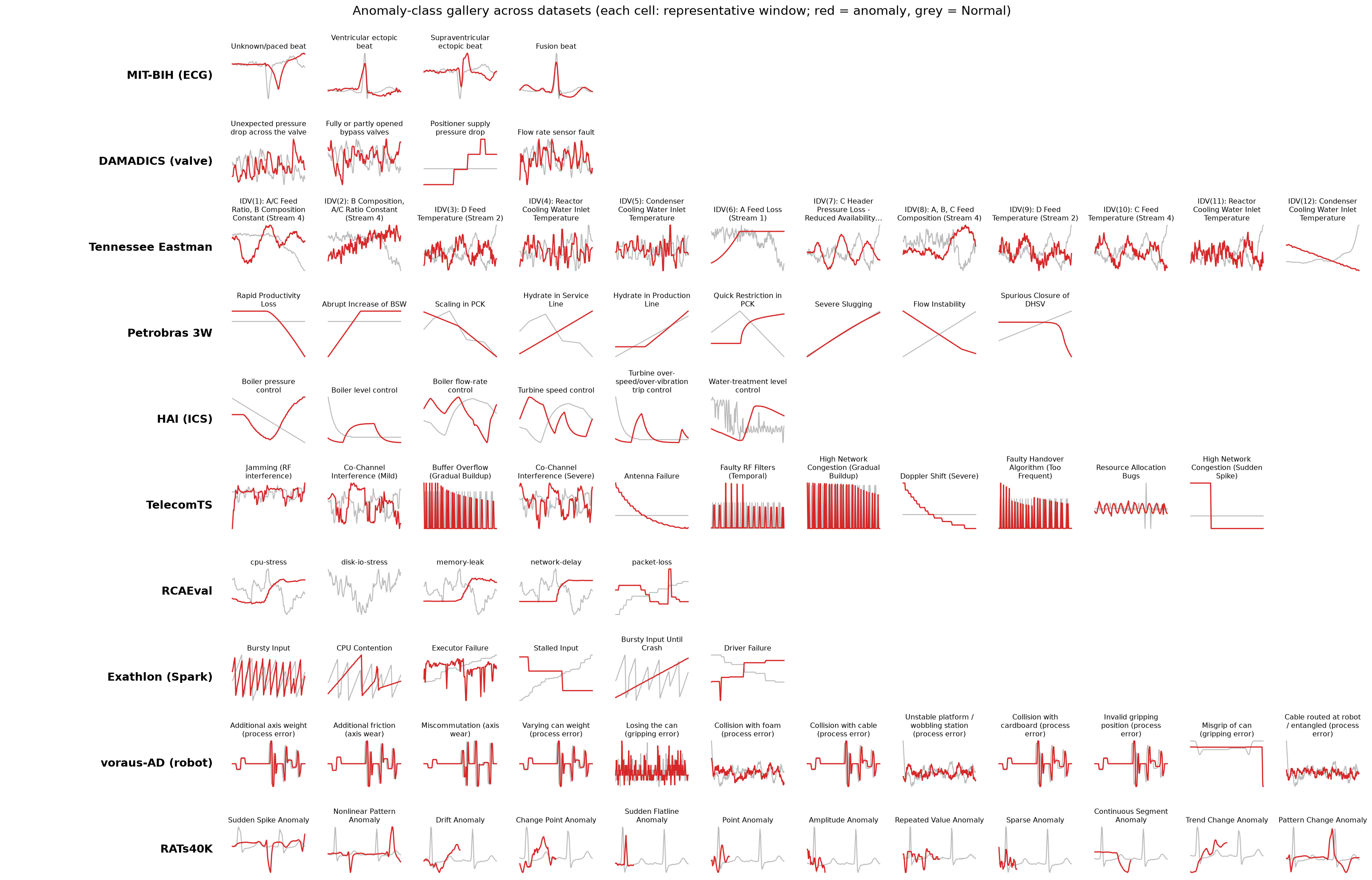}}
\caption{Representative anomaly classes across the READ-Bench datasets. Each row
is one dataset and each cell shows a representative window of one fault type
(red) against a normal window (grey).}
\label{fig:anomalies}
\end{figure}

The datasets span three axes of difficulty that Section~\ref{sec:problem}
identifies, and their fault signatures vary widely across domains
(Figure~\ref{fig:fault-strip}). Channel count ranges from univariate ($C=1$ for CTSR) to
several hundred channels (ROAD at 664, RCAEval at 429), window length from
16 to 1024 timesteps, and class balance from perfectly uniform
(RCAEval and Tennessee Eastman, both balanced by construction) to severely
skewed (RATS40K and DAMADICS, whose rarest fault type has one to a few corpus
windows against hundreds for the most common). Two datasets are themselves
channel-heterogeneous. RATS40K mixes a univariate variant with a multivariate
one, so within it $C$ ranges over 1, 3, and 9 channels and $T$ over 16,
32, 64, and 128 timesteps, and Petrobras~3W drops any of its 7 nominal
sensors that a well never recorded, so its windows carry 2 to 7 channels.
The rarest classes in RATS40K and
ROAD contain a single corpus window, and the split protocol keeps such a class in
the corpus rather than the query set when it cannot supply both, so every
retained query still has a same-class neighbour to retrieve. The fault classes
themselves are also visually diverse, both across datasets and among the classes
within one dataset (Figure~\ref{fig:anomalies}), which is what makes fault-type
retrieval harder than detecting mere abnormality.

\subsubsection{Label construction}
\label{app:labeling}
The datasets differ in how their native annotations define a fault type, and
therefore in how the fault-type label $y$ of Equation~\ref{eq:windowlabel} is
derived. Table~\ref{tab:labeling} records, for each dataset, the source of the
fault-type label, the rule that resolves a window's label, and the split unit
that keeps queries disjoint from their corpus. The label rule takes one of
three forms. Most datasets take the strict majority of the per-timestep labels
within a window (``window majority''), and MIT-BIH is the same rule over a
coarser unit, cut around an annotated heartbeat and labeled by the majority
beat class among the beats it contains. Datasets whose faults are often
shorter than a window, namely the intrusion, control, and disturbance datasets whose
anomalies are brief injected events, additionally apply the short-fault
relaxation of Section~\ref{sec:construction}, labeling a window by a fault that
fills a majority of its own extent even without a window majority (``majority or
coverage''). The remaining datasets carry a ``native label'' that their
annotations dictate, where Voraus, RATS40K, and TelecomTS inherit a single
as-shipped label per recording or window.

A few datasets annotate a fault's brief transient onset with a code separate
from the one for the same fault in its steady phase, as Tennessee Eastman, for
instance, marks the initial ramp of a disturbance apart from its post-onset
regime. Because both codes denote the same underlying fault, we merge each
onset code into its steady fault before applying Equation~\ref{eq:windowlabel},
so a window is labeled by the fault itself regardless of whether it captures the
onset or the settled phase.

\begin{table}[H]
\centering
\scriptsize
\setlength{\tabcolsep}{2.5pt}
\renewcommand{\arraystretch}{0.9}
\caption{How each dataset's fault-type label is derived. ``Label source''
is the native annotation the label is read from; ``label rule'' is how a
window's fault type is resolved from it, one of ``window majority'' (a fault
holds over half the window), ``majority or coverage'' (window majority or the
short-fault relaxation of Section~\ref{sec:construction}), or ``native label''
(the window inherits its as-shipped annotation); ``split unit'' is the
granularity at which train and test are kept disjoint so that queries are held
out from their corpus.}
\label{tab:labeling}
\begin{tabular}{@{}llll@{}}
\toprule
Dataset & Label source & Label rule & Split unit \\
\midrule
CTSR              & UCR dataset and class label & native label & random hold-out \\
DAMADICS          & actuator-fault schedule & majority or coverage & fault instance \\
Exathlon          & disturbance interval & majority or coverage & trace \\
HAI               & attack schedule & majority or coverage & attack instance \\
MIT-BIH           & per-beat class & window majority & patient recording \\
Petrobras 3W      & per-timestep event class & window majority & instance \\
RATS40K           & per-window anomaly type & native label & native split \\
RCAEval           & per-timestep fault type & window majority & service $\times$ fault \\
ROAD              & attack family and variant & majority or coverage & capture \\
TelecomTS         & per-window anomaly type & native label & shuffled split \\
Tennessee Eastman & per-timestep fault mode & window majority & simulation run \\
Voraus            & per-recording anomaly category & native label & recording \\
\bottomrule
\end{tabular}
\end{table}

\subsection{Method Suite Formal Definitions}
\label{app:baseline_math}

We give the exact formulation of every method family of
Section~\ref{sec:methods}. Let
$x,y$ denote two windows, where $x_t\in\mathbb{R}^{C}$ is the value at timestep $t$ of a
$C$-channel window of length $T$, and $x^{(c)}$ is channel $c$. A query is $q$ and
a corpus item is $x\in\mathcal{C}$, and every method induces a score $s(q,x)$ that is
ranked in descending order to return the top-$K$.

\subsubsection{Trivial baselines}
Two label-blind floors bound the achievable range. Random assigns each corpus
item an independent uniform score $s(q,x)\sim\mathrm{Unif}[0,1]$, giving a
chance-level ranking. Majority scores every query by corpus-class prevalence,
$s(q,x)=\mathbb{1}[\,y_x=y^\star\,]$ with $y^\star$ the most frequent corpus
label, so the ranking places the majority class first and is identical across
queries.

\subsubsection{Raw-window distances}
Distances $d$ are turned into scores by $s(q,x)=-d(q,x)$. The lock-step (ED) and
symbolic (SAX/SFA-BM25) measures require the query and corpus window to share a
length and channel count; when they differ we right-pad the narrower window's
channels with zeros up to the shared maximum $C$ and linearly interpolate both
windows' lengths onto a shared grid (the maximum observed length, so no measured
point is discarded) before scoring. The elastic (DTW) and cross-correlation
(SBD) measures align series of unequal length natively and need only the
zero-channel-padding step.

\paragraph{Euclidean (ED).}
A lock-step measure whose multivariate extension is inherently
channel-independent~\citep{paparrizos2024survey},
$d_{\mathrm{ED}}(x,y)=\bigl(\sum_{c=1}^{C}\sum_{t=1}^{T}(x^{(c)}_t-y^{(c)}_t)^2\bigr)^{1/2}$.

\paragraph{Shape-Based Distance (SBD)~\citep{paparrizos2015k}.}
A sliding measure that finds the best alignment between two time series via
normalized cross-correlation (NCC). Following the same channel-dependency taxonomy,
the channel-independent variant
(SBD-I) applies the univariate NCC to each channel separately; for channel
$c$, $\mathrm{NCC}(x^{(c)},y^{(c)})[s^{(c)}]=\dfrac{(x^{(c)}\star
y^{(c)})[s^{(c)}]}{\|x^{(c)}\|\,\|y^{(c)}\|}$, accelerated in the frequency
domain through the Fast Fourier Transform (FFT) and its inverse (IFFT), and
SBD-I sums the resulting channel scores,
$d_{\mathrm{SBD\text{-}I}}(x,y)=\sum_{c=1}^{C}\bigl(1-\max_{s^{(c)}}
\mathrm{NCC}(x^{(c)},y^{(c)})[s^{(c)}]\bigr)$.
The channel-dependent variant (SBD-D) instead correlates $x,y$
jointly, as a single $T\times C$ matrix, under one shared lag $s$; the
multivariate cross-correlation is
$\mathrm{MCC}(x,y)=\mathrm{IFFT2}\bigl(\mathrm{FFT2}(x)\odot\overline{\mathrm{FFT2}(y)}\bigr)$,
where $\odot$ denotes the elementwise product and the overline the complex
conjugate. Normalized by the norms of
the two windows, $\|x\|_F=\bigl(\sum_{c=1}^{C}\sum_{t=1}^{T}(x^{(c)}_t)^2\bigr)^{1/2}$,
this gives $\mathrm{MNCC}(x,y)[s]=\dfrac{\mathrm{MCC}(x,y)[s]}{\|x\|_F\,\|y\|_F}$
and $d_{\mathrm{SBD\text{-}D}}(x,y)=1-\max_{s}\,\mathrm{MNCC}(x,y)[s]$.
We use SBD-D throughout due to its better performance compared to SBD-I when evaluated under both classification and clustering settings~\citep{li2026mufasa}.

\paragraph{Dynamic Time Warping (DTW)~\citep{sakoe1978dynamic}.}
An elastic measure that permits one-to-many point matching, rather than the
one-to-one matching of lock-step measures, to achieve local alignment. DTW
finds the optimal alignment path and the minimum distance between two
series via the accumulated cost
$\gamma(i,j)=\delta(i,j)+\min\{\gamma(i-1,j),\gamma(i,j-1),\gamma(i-1,j-1)\}$
and total distance $\gamma(T,T)$, optionally under a Sakoe--Chiba band of
radius $r$. Following the same channel-dependency taxonomy, the independent variant (DTW-I)
sets $\delta^{(c)}(i,j)=(x^{(c)}_i-y^{(c)}_j)^2$ separately for each channel $c$,
giving each channel its own optimal alignment and per-channel distance
$d_{\mathrm{DTW}}(x^{(c)},y^{(c)})=\gamma^{(c)}(T,T)$, and sums the
channel distances,
$d_{\mathrm{DTW\text{-}I}}(x,y)=\sum_{c=1}^{C} d_{\mathrm{DTW}}(x^{(c)},y^{(c)})$
\citep{shokoohi2017generalizing}. The dependent variant (DTW-D)
instead finds a single alignment shared by every channel; treating $x,y$ as
one $T\times C$ matrix, the joint cost
$\delta(i,j)=\sum_{c=1}^{C}(x^{(c)}_i-y^{(c)}_j)^2$ feeds the same
recursion, giving $d_{\mathrm{DTW\text{-}D}}(x,y)=\gamma(T,T)$.

\subsubsection{Embedding-space scorers}
Let $\phi(x)\in\mathbb{R}^{D}$ be a fixed (precomputed) embedding.

\paragraph{Cosine.}
$s_{\cos}(q,x)=\dfrac{\phi(q)^\top\phi(x)}{\|\phi(q)\|\,\|\phi(x)\|}$.

\paragraph{Normal-Residual (NR).}
NR treats a fault as a deviation from
the embedding expected under normal operation rather than as an absolute
location in embedding space. It first $\ell_2$-normalizes every embedding,
$\hat\phi(x)=\phi(x)/\|\phi(x)\|$, and subtracts from each window the embedding
it would be expected to have if it were normal,
$\mathbb{E}[\hat\phi(u)\mid\mathrm{normal}]$. We approximate this conditional
expectation by a $k$-nearest-neighbour estimate over a train-split pool
$\mathcal{N}$ of normal windows, letting $\mathcal{N}_k(u)\subseteq\mathcal{N}$
be the $k$ normal windows whose normalized embeddings have the highest cosine
similarity to $\hat\phi(u)$, and take their mean
$\mu_k(u)=\tfrac{1}{k}\sum_{x\in\mathcal{N}_k(u)}\hat\phi(x)$ (we use $k{=}30$).
Averaging the $k$ nearest normals rather than the single closest is a
variance-reduction step, since one nearest normal injects its own embedding noise
into every residual, whereas the local mean cancels that noise while remaining
specific to the window's operating point. A $k$-sweep across the 12 datasets
finds $k{=}1$ consistently weakest and a mid-range $k$ ($\approx30$) best.
Subtracting this normal reference gives the residual
$r(u)=\hat\phi(u)-\mu_k(u)$ (the part of a window its expected-if-normal
embedding does not explain), which is renormalized and scored by cosine,
\[
  \hat r(u)=\frac{r(u)}{\|r(u)\|},\qquad
  s_{\mathrm{NR}}(q,x)=\hat r(q)^{\top}\hat r(x).
\]
Removing the expected-if-normal component leaves only what departs from
normal, so similarity is measured relative to the normal operating envelope
rather than absolute position. NR is defined for any fixed embedder $\phi$; the
per-embedder factorial is given in Table~\ref{tab:app-nr}.

\subsubsection{Symbolic bag-of-words retrievers}
Each channel is discretized into a word sequence and scored with Okapi
BM25~\citep{robertson2009bm25}.
For a query word set with term frequencies and a corpus ``document'' $x$ of
length $|x|$ (number of words) with average length $\overline{|x|}$,
\[
  \mathrm{BM25}(q,x)=\sum_{w\in q}\mathrm{IDF}(w)\,
  \frac{f(w,x)\,(k_1+1)}{f(w,x)+k_1\bigl(1-b+b\,|x|/\overline{|x|}\bigr)},
\]
with $\mathrm{IDF}(w)=\log\frac{N-n(w)+0.5}{n(w)+0.5}$, $n(w)$ the number of
corpus items containing $w$, and defaults $k_1=1.5,\,b=0.75$. Scores are averaged
over channels.

\paragraph{SAX-BM25.}
Words are SAX symbols~\citep{lin2007sax}, a piecewise-aggregate
approximation of the $z$-normalized window followed by Gaussian breakpoints.

\paragraph{SFA-BM25.}
Words are SFA symbols~\citep{schafer2012sfa}, low-frequency DFT
coefficients discretized by multiple-coefficient binning (MCB).

\subsubsection{Foundation-model embedders}
The frozen encoders CHARM, Chronos-2, MantisV2,
and TiRex each map a window to $\phi(x)$ and are retrieved by Euclidean
distance on the embeddings, $s(q,x)=-d_{\mathrm{ED}}(\phi(q),\phi(x))$, with
$s_{\cos}$ giving near-identical rankings~\citep{hu2026tsrbench}. Because windows vary in both length and channel count across
(and within) datasets while each encoder expects a fixed interface, we apply a
uniform adaptation along both axes. \textbf{Channels.} Each channel $x^{(c)}$ is
embedded independently to $\phi(x^{(c)})\in\mathbb{R}^{D}$, and the per-channel
embeddings are averaged over the channel axis,
$\phi(x)=\tfrac{1}{C}\sum_{c=1}^{C}\phi(x^{(c)})$, so the output dimension is $D$
regardless of $C$ (mean-pooling, rather than concatenation, is what makes the
representation channel-count invariant). \textbf{Length.} When an encoder
constrains the admissible input length, the window is resampled to the nearest
admissible length by linearly interpolating each channel onto a shared time
grid before encoding, adding interpolated points for shorter windows without
discarding any measured point, and never zero-padding the time axis (MantisV2
requires the length to be a multiple of its patch grid; TiRex and Chronos-2
additionally mean-pool over their internal patch axis). This keeps
$\phi(x)\in\mathbb{R}^{D}$ fixed-dimensional regardless of the
window's native $T$ and $C$, so a query and a corpus item are always comparable
even when their shapes differ.

\paragraph{MantisV2~\citep{mantisv2}.}
A Transformer encoder pretrained on synthetic data. Each channel is tokenized into a fixed grid of 32 patches, with each patch token formed by concatenating three descriptors, a convolutional feature of the raw signal, the same feature computed on its first-order difference, and patch-level mean and standard-deviation statistics, linearly projected and layer-normalized. These tokens, with a prepended class token, are processed by 6 Transformer layers, and $\phi(x^{(c)})$ is read from the class token's final hidden state.

\paragraph{Chronos-2~\citep{chronos2}.}
An encoder-only Transformer pretrained for zero-shot forecasting. The series is split into fixed-length patches and embedded by a residual network, then processed by alternating time attention (self-attention along the temporal axis) and group attention layers, the latter aggregating information across related series or channels, $h_i=\sum_{j:\,g_j=g_i}\alpha_{ij}\,v_j$, for in-context learning; $\phi(x)$ is read from the resulting patch embeddings.

\paragraph{TiRex~\citep{tirex}.}
An xLSTM-based recurrent sequence model pretrained for zero-shot forecasting. Patches are embedded by a residual block and processed sequentially by stacked sLSTM blocks in place of self-attention, $h_p=\mathrm{xLSTM}(h_{p-1},u_p)$, with $\phi(x)$ read from the resulting hidden states.

\paragraph{CHARM~\citep{dutta2026giving}.}
Trained with a Joint-Embedding Predictive Architecture (JEPA). A contextual temporal convolutional network first extracts per-timestep, per-channel features $\mathbf{X}=\mathrm{TCN}_\theta(x)$ conditioned on textual channel descriptions, and a stack of contextual attention layers fuses the channel and temporal dimensions, $\mathbf{Y}[i,j,:]=\sum_{j'}\alpha_{j,j'}(E_d)\,\mathbf{X}[i,j',:]$, to produce $\phi(x)$.

The per-encoder embedding dimension and context handling are reported in
Appendix~\ref{app:hyperparams}.

\subsubsection{Supervised classifiers}
MR-Hydra~\citep{dempster2023hydra} and RDST~\citep{rdst} are trained on the
labeled corpus to produce a class posterior $g_\theta(\cdot)$. Used as a retrieval
scorer, an item $x$ receives the query's posterior mass on its class,
$s(q,x)=[g_\theta(q)]_{\,y_x}$. Because they ultimately emit a label
$\hat y=\arg\max_y [g_\theta(q)]_y$ rather than a ranking, we also report
macro-F1 $=\tfrac{1}{|\mathcal{Y}|}\sum_{y\in\mathcal{Y}}\mathrm{F1}_y$ as an
upper-bound reference. The kernel and shapelet counts and the training regime
are given in Appendix~\ref{app:hyperparams}.

\subsubsection{Score fusion}
Given two base scorers $A,B$ over $\mathcal{C}$, we form a fused score in three
ways.

\paragraph{Reciprocal-rank fusion (RRF)~\citep{cormack2009rrf}.}
With $\operatorname{rank}_A(q,x)$ the rank
of $x$ under $A$,
$s_{\mathrm{RRF}}(q,x)=\dfrac{1}{k_0+\operatorname{rank}_A(q,x)}+\dfrac{1}{k_0+\operatorname{rank}_B(q,x)}$,
default $k_0=60$.

\paragraph{Weighted sum (WS).}
$s_{\mathrm{WS}}(q,x)=\alpha\,z\!\left(s_A\right)+(1-\alpha)\,z\!\left(s_B\right)$,
where $z(\cdot)$ is per-query $z$-scoring and $\alpha\in[0,1]$.

\paragraph{Two-step cascade.}
Retrieve the top-$K'$ by $A$, then re-rank that
shortlist by $B$.

\noindent The fusion grid pairs each of the four FM embedders with each of five
shape/symbolic legs under these three modes.

\subsubsection{Rerankers}
A reranker adjusts the base score on the retrieved pool $\mathcal{P}$ (the
top-$k$ neighbours of $q$). Let $y_x$ be the label of $x$.

\paragraph{PRF (pseudo-relevance feedback)~\citep{rocchio1971prf}.}
Form the centroid
$\bar c=\tfrac{1}{k}\sum_{x\in\mathcal{P}}\phi(x)$ of the top-$k$ and re-score
$s'(q,x)=\beta\,s_{\mathrm{base}}(q,x)-(1-\beta)\,\|\phi(x)-\bar c\|$.

\paragraph{Purity.}
Let $\pi(x)$ be the fraction of $x$'s $k$ nearest
neighbours sharing its label; $s'(q,x)=\alpha\,z(s_{\mathrm{base}})+(1-\alpha)\,z(\pi)$.

\paragraph{QPurity (query-conditioned purity).}
Predict the query label
$\hat y_q$ by a vote over $q$'s top-$k$, then credit purity only on candidates
with $y_x=\hat y_q$.

\paragraph{Majority-Vote.}
Predict $\hat y_q$ from the labels in
$\mathcal{P}$ and promote candidates with $y_x=\hat y_q$ above the rest.

\paragraph{GPC (Gaussian-process classifier)~\citep{rasmussen2006gpml}.}
Fit a GP classifier on the
labeled neighbours in $\mathcal{P}$ and set $s'(q,x)=\Pr(y_x=y_q\mid
q,x,\mathcal{P})$, its predicted probability that $x$ shares the query's
class. The kernel and PCA settings are given in Appendix~\ref{app:hyperparams}.

\subsubsection{Language-model rerankers}
Toto-1.0-QA~\citep{toto}, ChatTS~\citep{xie2025chatts}, Claude, and Claude\,$+$\,TSAD (after
LLMAD~\citep{liu2025llmad}) reorder the candidate pool $\mathcal{P}$ by reading the
retrieved series directly rather than by evaluating a closed-form score. For all
four methods $\mathcal{P}$ is the top-$20$ pool returned by the CHARM base
retriever, so $|\mathcal{P}|{=}20$, and each candidate is presented with its
class label, so these rerankers are label-aware in the sense of
Section~\ref{sec:methods} (the query's own label is never shown). Each window is first reduced to its
$k_{\mathrm{ch}}{=}20$ highest-variance channels, and this reduction is applied
identically to every representation the model receives (numeric tensor,
placeholder token, or rendered image) so that a reduced window is never shown as
mismatched modalities of the same series. The methods differ in how a ranking is
elicited from $\mathcal{P}$.

The pointwise rerankers, Toto-1.0-QA and ChatTS, issue one request per pair
$(q,x)$ with $x\in\mathcal{P}$. Each request returns a scalar relevance
judgment, a binary match with an associated confidence for Toto-1.0-QA and a
single similarity score on the $0$ to $100$ scale for ChatTS, and the resulting
$|\mathcal{P}|$ scores are sorted in
descending order to induce the reranking. A pointwise reranker therefore issues
$|\mathcal{P}|$ model calls per query.

The listwise rerankers, Claude and Claude\,$+$\,TSAD, issue a single request per
query that presents $q$ together with all of $\mathcal{P}$ and returns the ranked
candidate identifiers directly, so one query induces one model call.
Claude\,$+$\,TSAD differs from Claude only in its input representation, comparing the
per-channel deseasonalized residual of each series rather than the raw series, so
that ranking is driven by anomalous deviation rather than shared periodic shape.

Each model's output is turned into the same $(q,x)$ score matrix that every
other method produces, so the LLM rerankers are scored by the identical metrics
of Appendix~\ref{app:metrics}. A reranker promotes the candidates the model
returned, in the model's order, above the base retriever's remaining ordering of
the pool, leaving corpus items outside $\mathcal{P}$ at their base scores. The
no-retrieval classification baseline instead reads a per-class confidence and
assigns it to every corpus item of that class, so an unrecognized class
contributes zero, matching how the supervised classifiers are scored.

The serving, decoding, and series-rendering settings for all four methods are
given in Appendix~\ref{app:hyperparams}, and the exact prompts in
Appendix~\ref{app:llm-prompts}.

\subsection{Evaluation Metric Definitions}
\label{app:metrics}

Let $q$ be a query, $\mathrm{rel}(q,x) \in \{0, 1\}$ indicate whether corpus item
$x$ is relevant to $q$ (Section~\ref{sec:problem}), and let $x_1, \dots, x_K$ denote the
top-$K$ items returned for $q$, ranked by descending score.

\paragraph{Precision at $K$ (P@$K$).}
\[
  \mathrm{P@}K(q) = \frac{1}{K}\sum_{i=1}^{K} \mathrm{rel}(q,x_i).
\]

\paragraph{Hit Rate at $K$ (HR@$K$).}
\[
  \mathrm{HR@}K(q) = \mathbb{1}\!\left[\textstyle\sum_{i=1}^{K} \mathrm{rel}(q,x_i) > 0\right],
\]
i.e.\ whether at least one relevant item appears in the top $K$.

\paragraph{Normalized Discounted Cumulative Gain (NDCG@$K$)~\citep{jarvelin2002ndcg}.}
\[
  \mathrm{DCG@}K(q) = \sum_{i=1}^{K} \frac{\mathrm{rel}(q,x_i)}{\log_2(i+1)}, \qquad
  \mathrm{NDCG@}K(q) = \frac{\mathrm{DCG@}K(q)}{\mathrm{IDCG@}K(q)},
\]
where $\mathrm{IDCG@}K(q)$ is $\mathrm{DCG@}K$ under the ideal ranking (all
relevant items first). Unlike P@$K$/HR@$K$, NDCG@$K$ is sensitive to the
order of retrieved items, not just set membership.

All three metrics are averaged over queries (and, where noted, over datasets)
to produce the per-method scores compared in Section~\ref{sec:experiments}.

\subsection{Statistical Testing Definitions}
\label{app:stats}

To compare methods beyond raw score differences we use a Friedman test with a
Nemenyi post-hoc test, blocking on the 12 datasets~\citep{demsar2006}. Let $m$ be the number of
methods compared and $n=12$ the number of datasets.

\paragraph{Friedman test.}
For each dataset $j$, rank the $m$ methods by their metric value, giving rank
$r_i^j$ to method $i$. The Friedman statistic is
\[
  \chi^2_F = \frac{12n}{m(m+1)} \left[ \sum_{i=1}^{m} \bar{r}_i^2 - \frac{m(m+1)^2}{4} \right],
  \qquad \bar{r}_i = \frac{1}{n}\sum_{j=1}^{n} r_i^j,
\]
tested against a $\chi^2$ distribution with $m-1$ degrees of freedom; rejection
indicates that at least one method differs significantly from the others.

\paragraph{Nemenyi post-hoc test.}
Following a significant Friedman result, two methods $i, i'$ are declared
significantly different when their mean ranks differ by more than the
critical difference,
\[
  |\bar{r}_i - \bar{r}_{i'}| > q_\alpha \sqrt{\frac{m(m+1)}{6n}},
\]
where $q_\alpha$ is the critical value of the studentized range distribution
at family-wise significance level $\alpha=0.05$.

\subsection{Hyperparameters and Compute Budget}
\label{app:hyperparams}

This appendix records the exact search space or fixed setting and the final
chosen value for every method family of Section~\ref{sec:methods}, together with
the compute budget. Values fixed by a pretrained checkpoint or inherited from a
library default are marked as such in the ``source'' column, so that our own
choices are distinguishable from upstream defaults. All retrieval metrics are
computed at cutoffs $K\in\{1,3,5,10,20\}$.

\paragraph{Distance and symbolic baselines.}
Euclidean distance and SBD carry no tunable settings. SBD is run in its
dependent (SBD-D) form, and DTW is banded (Sakoe--Chiba) and defaults to its
independent (DTW-I) channel mode. The symbolic
retrievers discretize each channel with a sliding window and score the
resulting words with Okapi BM25, SAX with piecewise-aggregate approximation
(PAA) over Gaussian breakpoints and SFA with discrete-Fourier-transform (DFT)
coefficients under equi-depth multiple-coefficient binning (MCB). Table~\ref{tab:hp-distance} lists the tunable settings.

\begin{table}[H]
\centering
\scriptsize
\setlength{\tabcolsep}{4pt}
\renewcommand{\arraystretch}{0.9}
\caption{Distance and symbolic baseline settings. $T$ is the window length, so
the DTW band scales with the window. The symbolic vocabulary size is
alphabet$^{\text{word length}}$.}
\label{tab:hp-distance}
\begin{tabular}{@{}lll@{}}
\toprule
Method & Hyperparameter & Value \\
\midrule
DTW      & Sakoe--Chiba band radius   & $\max(1,\operatorname{round}(0.1\,T))$ \\
NR       & nearest-normal count $k$   & 30 \\
\midrule
\multirow{3}{*}{SAX-BM25} & window / stride         & 96/8 \\
                         & PAA word length         & 6 \\
                         & alphabet size           & 4 \\
\midrule
\multirow{3}{*}{SFA-BM25} & window / stride         & 128/8 \\
                         & word length (DFT coefficients) & 8 \\
                         & alphabet size           & 3 \\
\midrule
BM25     & $k_1$ / $b$             & 1.5/0.75 \\
\bottomrule
\end{tabular}
\end{table}

\paragraph{Foundation-model embedders.}
Each embedder is a frozen pretrained encoder, so its embedding dimension and
context handling are fixed by the released checkpoint rather than tuned here.
We embed each channel independently and mean-pool over channels, so the
representation is channel-count invariant, and we set only the inference batch
size at the call site. Table~\ref{tab:hp-fm} records the per-encoder settings.

\begin{table}[H]
\centering
\scriptsize
\setlength{\tabcolsep}{4pt}
\renewcommand{\arraystretch}{0.9}
\caption{Foundation-model embedder settings. The embedding dimension and any
patch-grid constraint are fixed by the released checkpoint; pooling and
inference batch size are set by us.}
\label{tab:hp-fm}
\begin{tabular}{@{}lll@{}}
\toprule
Encoder & Hyperparameter & Value \\
\midrule
\multirow{2}{*}{CHARM} & embedding dimension & 384 \\
                       & pooling / batch     & mean / 32 \\
\midrule
\multirow{2}{*}{Chronos-2} & embedding dimension & 768 \\
                           & pooling / batch     & mean / 32 \\
\midrule
\multirow{3}{*}{MantisV2}  & per-channel width   & 256 \\
                           & patch grid          & 32 patches ($T$ a multiple of 32, $T\ge64$) \\
                           & pooling / batch     & mean / 32 \\
\midrule
\multirow{2}{*}{TiRex}     & per-channel width   & $12\times512$ \\
                           & pooling / batch     & mean / 512 \\
\bottomrule
\end{tabular}
\end{table}

\paragraph{Supervised classifiers.}
MR-Hydra and RDST are the aeon implementations, run at their library defaults
for the kernel, shapelet, and ensemble counts. MR-Hydra averages the class
posteriors of its MultiRocket and Hydra members in equal weight.

\paragraph{Fusion and reranking.}
Fusion pairs one embedder with one shape or symbolic leg, and each reranker
adjusts a base score on a retrieved pool. Table~\ref{tab:hp-fusion} records the
constants.

\begin{table}[H]
\centering
\scriptsize
\setlength{\tabcolsep}{4pt}
\renewcommand{\arraystretch}{0.9}
\caption{Fusion and reranking settings. The weighted sum and the purity-based
rerankers combine per-query $z$-scored terms with weight $\alpha$.}
\label{tab:hp-fusion}
\begin{tabular}{@{}lll@{}}
\toprule
Method & Hyperparameter & Value \\
\midrule
RRF            & fusion constant $k_0$        & 60 \\
Weighted sum   & weight $\alpha$              & 0.5 \\
Two-step cascade & shortlist size $K'$        & 100 \\
\midrule
PRF            & neighbourhood $k$ / weight $\beta$ & 10/0.5 \\
Purity         & neighbourhood $k$ / weight $\alpha$ & 10/0.5 \\
Majority-Vote  & neighbourhood $k$            & 10 \\
QPurity        & neighbourhood $k$ / vote $k$ / weight $\alpha$ & 10/10/0.5 \\
\midrule
\multirow{3}{*}{GPC} & fit-window $K$               & 20 \\
                     & PCA variance retained        & 0.95 \\
                     & kernel                       & L1-norm, length scale 1.0 (fixed) \\
\bottomrule
\end{tabular}
\end{table}

\paragraph{Language-model rerankers.}
The method mechanics of the four language-model rerankers, namely the shared
top-$20$ CHARM pool, the pointwise versus listwise scoring, and the
highest-variance channel cap, are given in Appendix~\ref{app:baseline_math}, and
their exact prompts, including the no-retrieval classification baseline of
Table~\ref{tab:llm}, in Appendix~\ref{app:llm-prompts}. Toto-1.0-QA%
\footnote{\url{https://huggingface.co/Datadog/Toto-1.0-QA-Experimental}} and
ChatTS%
\footnote{\url{https://huggingface.co/bytedance-research/ChatTS-14B}} run
locally from their released checkpoints and decode greedily (no sampling), up to
2000 and 512 new tokens respectively. Claude and Claude\,$+$\,TSAD are Claude
Sonnet 5 (1M-context) called through the AWS Bedrock Messages API at the default
temperature with extended thinking disabled and up to 2048 output tokens,
retried up to four times under exponential backoff on a throttling or empty
response. Each retrieved window is rendered for the model as one plot per
channel and a numeric series, and Claude\,$+$\,TSAD substitutes the per-channel
deseasonalized residual (following LLM-TSAD)%
\footnote{\url{https://github.com/junwoopark92/LLM-TSAD}} for the raw series.

\begin{figure}[t]
\centering
\begin{subfigure}[b]{0.49\columnwidth}
\centering
\includegraphics[width=\linewidth]{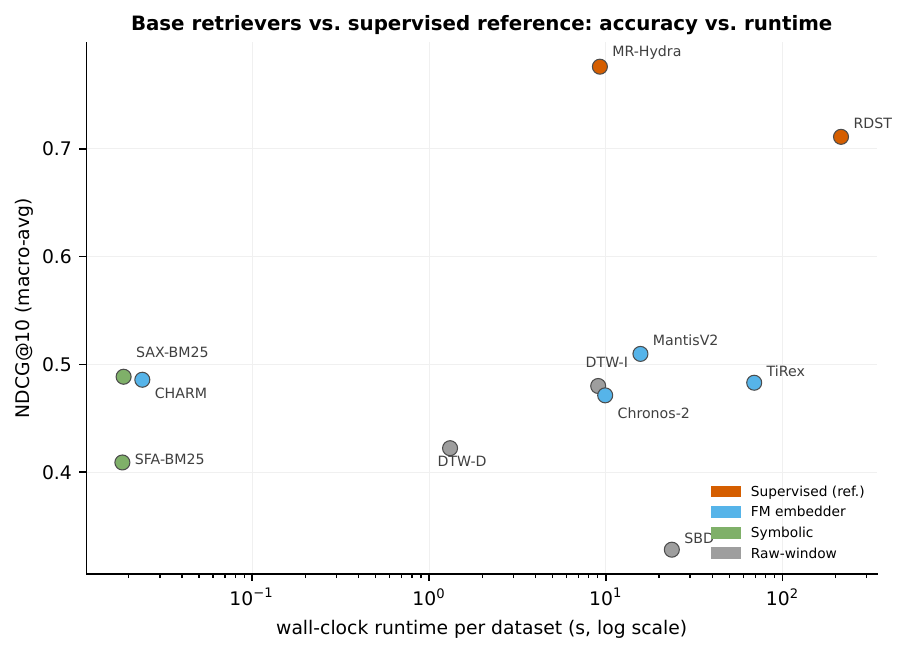}
\caption{}
\label{fig:runtime}
\end{subfigure}\hfill
\begin{subfigure}[b]{0.49\columnwidth}
\centering
\includegraphics[width=\linewidth]{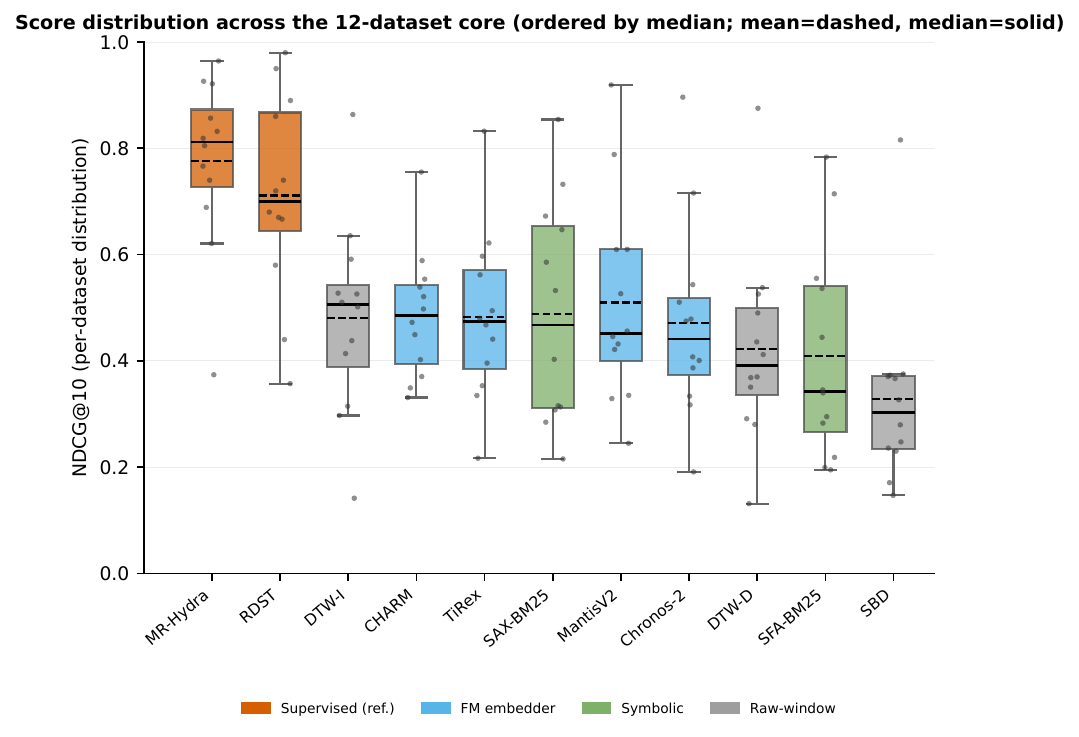}
\caption{}
\label{fig:box}
\end{subfigure}
\caption{Base retrievers and the supervised reference over the 12 datasets,
both at $\rho{=}0\%$ and family-coloured.
(\subref{fig:runtime}) Retrieval accuracy against wall-clock scoring time (log
scale, median per-dataset runtime). The base retrievers span two orders of
magnitude in scoring time yet cluster in a narrow accuracy band well below the
supervised reference, so among label-free methods the choice of a classical
distance, a symbolic retriever, or a pretrained embedder trades more compute
than retrieval quality.
(\subref{fig:box}) The matching per-dataset NDCG@10 distribution, one box per
method ordered by median (dashed line is the mean, solid the median, dots the 12
per-dataset scores), where the label-free retrievers overlap heavily and remain
below the supervised reference.}
\label{fig:runtime-box}
\end{figure}

\paragraph{Compute and hardware.}
All experiments ran on a single node with two NVIDIA A100-SXM4-80GB GPUs, an
Intel Xeon Platinum 8275CL (96 cores) and 1.1\,TiB of RAM. The foundation-model
embedders, supervised classifiers, and the two locally-hosted language-model
rerankers (Toto-1.0-QA, ChatTS) use the GPU, while the distance, symbolic, and
fusion or reranking stages are CPU-only; Claude and Claude\,$+$\,TSAD instead
call the hosted AWS Bedrock API. The scoring cost of the retrieval stage itself
is modest and varies widely across method families, spanning about two orders of
magnitude, as Figure~\ref{fig:runtime} shows for the base retrievers.

\subsection{Language-Model Reranker Prompts}
\label{app:llm-prompts}

This appendix gives the exact system and question prompts sent to each
language-model method, verbatim from the run scripts, for both the
reranking task and the no-retrieval classification baseline, whose results
are reported in Table~\ref{tab:llm}. Every prompt
below is instantiated per query; \texttt{\{...\}} placeholders are filled in
at request time (a topk count, the option list, etc.). Every reranking prompt
appends one candidate label line per candidate, shown inline in each prompt
below, disclaiming that class membership implies similarity so the model treats
the label as neutral metadata; the query's own label is never shown.

\paragraph{Toto-1.0-QA reranking.}
One request per (query, candidate) pair; the model is shown two entities
(the QUERY, then this one CANDIDATE) as two separate images plus a packed
numeric tensor, and returns a match/confidence judgment that is converted
to a score (confidence if the match is ``yes'', else $100-{}$confidence) and
sorted across the pool's 20 candidates to obtain the ranking.
\begin{promptbox}{Toto-1.0-QA reranking prompt}
\begin{promptverb}
SYSTEM:
You are an expert observability anomaly analyst comparing time-series
patterns for retrieval reranking.

Input:
- Images: one plot per time series shown, in order -- the QUERY first,
  then each CANDIDATE.
- Time Series: the same series, embedded per channel.
- Question: the ranking instruction below.

QUESTION (per candidate):
In the following time-series, does the anomaly in the QUERY series
correlate with the anomaly in the CANDIDATE series, if anomalies
exist? Respond with ONLY a JSON object of the form
{"match": "yes"|"no", "confidence": <integer 0-100>} -- confidence
is how sure you are of your match/no-match judgment. No explanation.

(Appended to the CANDIDATE:)
For reference, this CANDIDATE is drawn from the class: {label}. This
class membership does NOT indicate similarity to the QUERY -- base
your similarity judgment ONLY on the actual time-series pattern.
\end{promptverb}
\end{promptbox}
When a dataset provides one, a short prose description of its sensors and
labels is appended to the system prompt under a ``Dataset context'' heading,
and a per-request time-series metadata string (channel names and timestamps)
is appended to the question when non-empty.

\paragraph{ChatTS reranking.}
Same shape as Toto-1.0-QA (one request per (query, candidate) pair, scores
sorted over the pool), but ChatTS has no vision input; each
channel of each entity becomes one \texttt{<ts></ts>} placeholder token in
the text, paired with its raw numeric series fed to the model's own
time-series encoder. It rates an anchored 0--100 similarity score.
\begin{promptbox}{ChatTS reranking prompt}
\begin{promptverb}
SYSTEM:
You are an expert observability anomaly analyst comparing time-series
patterns for retrieval reranking.

QUESTION:
Rate how closely the CANDIDATE time series resembles the QUERY time
series, considering shape, timing, magnitude, and duration of any
anomalous pattern. Use the FULL 0-100 scale -- do not default to a
high score just because the candidate was retrieved as a plausible
match; most candidates in a retrieval pool are only partial matches
and should score accordingly. Scale: 0-20 = no resemblance,
21-40 = weak/superficial resemblance only, 41-60 = partial
resemblance (some but not all features match), 61-80 = strong
resemblance (most features match), 81-100 = near-identical pattern.
Respond with ONLY a JSON object of the form {"similarity": <integer
0-100>}. No explanation.

(Appended to the CANDIDATE:)
For reference, this CANDIDATE is drawn from the class: {label}. This
class membership does NOT indicate similarity to the QUERY -- base
your similarity judgment ONLY on the actual time-series pattern.
\end{promptverb}
\end{promptbox}

\paragraph{Claude reranking.}
One request per query, with all 20 candidates shown together (each as
its own image plus a serialized numeric text block) and ranked directly.
\begin{promptbox}{Claude reranking prompt}
\begin{promptverb}
SYSTEM:
You are an expert observability anomaly analyst comparing time-series
patterns for retrieval reranking. You will be shown one plotted image
and one text block of numeric values per series -- the QUERY first,
then each CANDIDATE, in order.

QUESTION:
The first image and text block is the QUERY time series. Every
other image/text block is a CANDIDATE time series, labeled with its
candidate id (e.g. "sample_00007"). Rank the up to {topk} candidates
whose pattern most closely resembles/correlates with the query, best
match first. Respond with ONLY a JSON object of the form
{"ranked_ids": ["<candidate_id>", ...]} using the exact candidate id
strings given, most relevant first, no explanation.

(Appended to each CANDIDATE:)
For reference, this CANDIDATE is drawn from the class: {label}. This
class membership does NOT indicate similarity to the QUERY -- base
your similarity judgment ONLY on the actual time-series pattern.
\end{promptverb}
\end{promptbox}

\paragraph{Claude\,$+$\,TSAD reranking.}
Identical request shape to Claude above, except every series (query and
every candidate) is first deseasonalized per channel (autocorrelation
peak-period detection, clamped to a minimum period of 8, then an additive
seasonal decomposition; channels where this is degenerate fall back to
the raw values) -- both the image and the numeric text show this
residual, not the raw series.
\begin{promptbox}{Claude\,$+$\,TSAD reranking prompt}
\begin{promptverb}
SYSTEM:
You are an expert observability anomaly analyst comparing
deseasonalized time-series residual patterns for retrieval
reranking. You will be shown one plotted image and one text block
of numeric values per series -- the QUERY first, then each
CANDIDATE, in order.

QUESTION:
The first image and text block is the QUERY time series. Every
other image/text block is a CANDIDATE time series, labeled with its
candidate id (e.g. "sample_00007"). Each series has had its
seasonal component removed where detectable (deseasonalized
residual, noted per-channel) so you are comparing the anomalous
deviation pattern, not the raw periodic shape. Rank the up to
{topk} candidates whose pattern most closely resembles/correlates
with the query, best match first. Respond with ONLY a JSON object
of the form {"ranked_ids": ["<candidate_id>", ...]} using the exact
candidate id strings given, most relevant first, no explanation.

(Appended to each CANDIDATE:)
For reference, this CANDIDATE is drawn from the class: {label}. This
class membership does NOT indicate similarity to the QUERY -- base
your similarity judgment ONLY on the actual time-series pattern.
\end{promptverb}
\end{promptbox}

\paragraph{No-retrieval classification.}
Rather than reranking a retrieved pool, the query is compared directly
against one representative labeled EXAMPLE per class in the dataset's fault
taxonomy (no retrieval step), and the model rates its confidence that the
query matches each. The wording is shared near-verbatim across all four
models, differing only in whether it refers to ``image/series'' or ``entry''
for the respective input modality.
\begin{promptbox}{No-retrieval classification prompt}
\begin{promptverb}
SYSTEM:
You are an expert observability anomaly analyst rating how well a
query time series matches each of several labeled example patterns.
You will be shown one plotted image and one text block of numeric
values per series -- the QUERY first, then one labeled EXAMPLE per
category.

QUESTION:
Rate your confidence (0-100) that the QUERY belongs to the SAME
category as each numbered EXAMPLE below, based on how closely its
pattern matches the QUERY's:
{option_lines}

Use the full 0-100 scale -- most examples will only partially match,
so most scores should NOT be high. Respond with ONLY a JSON object
of the form {"scores": {"1": <0-100>, "2": <0-100>, ...}} covering
every number above. No explanation.
\end{promptverb}
\end{promptbox}

\subsection{Additional Results}
\label{app:full-results}

This appendix collects the complete result tables behind
Section~\ref{sec:experiments}. They cover the full set of metrics for the base
retrievers, how each method behaves as the corpus is polluted, normal-residual
scoring on every embedder, the complete fusion and reranking results, their
combination, and a check that the findings hold on the uncapped corpus. All numbers
are NDCG@10 unless stated, macro-averaged over the 12 datasets, at pollution
levels $\rho{=}0\%/\!\approx\!10\%/\!\approx\!20\%$ where shown. In every table
the best value in each column is bold and the second best is underlined.

\subsubsection{Full base-retriever metric grid}
\label{app:backbone-grid}
Table~\ref{tab:app-backbones} extends the headline NDCG@10 of
Table~\ref{tab:backbones} to the full metric set at $\rho{=}0\%$, adding P@1,
P@5, HR@10, NDCG@5, NDCG@20, and macro-F1, as well as DTW-D. Across every metric the
label-free retrievers stay close and below the supervised reference, so the
choice among them does not change retrieval quality on any single metric.
$\sigma$, the standard deviation of NDCG@10 across the 12 datasets, is smallest
for CHARM, so it is also the most consistent across domains.

\begin{table}[H]\centering
\scriptsize
\setlength{\tabcolsep}{2.5pt}
\renewcommand{\arraystretch}{0.9}
\caption{Full base-retriever metric grid at $\rho{=}0\%$, macro-averaged over the 12
datasets. $\sigma$ is the standard deviation of NDCG@10 across datasets. The
supervised rows are a classification reference rather than retrieval systems.}
\label{tab:app-backbones}
\begin{tabular}{llcccccccc}
\toprule
Method & Family & P@1 & P@5 & HR@10 & NDCG@5 & NDCG@10 & NDCG@20 & macro-F1 & $\sigma$ \\
\midrule
Random & Trivial floor & 0.172 & 0.183 & 0.699 & 0.182 & 0.189 & 0.192 & 0.125 & 0.10 \\
Majority &  & 0.246 & 0.246 & 0.246 & 0.246 & 0.246 & 0.247 & 0.060 & 0.13 \\
\addlinespace
SBD-D & Raw-window distance & 0.353 & 0.340 & 0.739 & 0.345 & 0.328 & 0.317 & 0.330 & 0.17 \\
DTW-I &  & 0.568 & 0.492 & 0.857 & 0.510 & 0.480 & 0.470 & 0.464 & 0.18 \\
DTW-D &  & 0.507 & 0.434 & 0.787 & 0.451 & 0.422 & 0.412 & 0.435 & 0.18 \\
\addlinespace
SAX-BM25 & Symbolic & 0.553 & 0.498 & \underline{0.897} & 0.512 & \underline{0.489} & 0.466 & \underline{0.503} & 0.20 \\
SFA-BM25 &  & 0.490 & 0.422 & 0.862 & 0.440 & 0.409 & 0.390 & 0.428 & 0.19 \\
\addlinespace
Chronos-2 & FM embedder & 0.530 & 0.486 & 0.885 & 0.497 & 0.471 & 0.454 & 0.445 & 0.18 \\
MantisV2 &  & \textbf{0.583} & \textbf{0.525} & 0.878 & \textbf{0.538} & \textbf{0.510} & \textbf{0.489} & \textbf{0.518} & 0.19 \\
TiRex &  & \underline{0.579} & 0.491 & \textbf{0.905} & 0.510 & 0.483 & 0.466 & 0.473 & \underline{0.15} \\
CHARM &  & 0.550 & \underline{0.505} & 0.897 & \underline{0.515} & 0.486 & \underline{0.471} & 0.481 & \textbf{0.11} \\
\midrule
MR-Hydra & Supervised reference & 0.759 & 0.771 & 0.850 & 0.772 & 0.776 & 0.785 & 0.666 & 0.15 \\
RDST &  & 0.711 & 0.711 & 0.711 & 0.711 & 0.711 & 0.712 & 0.617 & 0.18 \\
\bottomrule
\end{tabular}
\end{table}

\subsubsection{Scaling to the full corpus}
\label{app:fullcorpus}
The main-text results cap each corpus to a common size
(Table~\ref{tab:corpus-sizes}) for uniformity and compute. To check that the
closeness of the label-free retrievers (Section~\ref{sec:base}) is not an
artifact of that cap, we rerun them on the uncapped corpus of each dataset, up
to over a million windows. Five datasets are already at or below the cap, so for
them the uncapped corpus is the one already reported; the other seven scale up
substantially. The elastic distances (DTW-I, DTW-D), the shape-based distance
(SBD-D), the symbolic BM25 retrievers, and the supervised classifiers do not scale
to the largest corpora and are omitted; the table compares the
foundation-model embedders (which do scale) against their capped scores across
all 12 datasets. Absolute scores drop as the
corpus grows, as expected with far more candidates to rank against, and the
structure holds, with the label-free embedders still close and no single
retriever uniformly best.

\begin{table}[H]
\centering
\scriptsize
\setlength{\tabcolsep}{2.5pt}
\renewcommand{\arraystretch}{0.9}
\caption{Capped versus full-corpus NDCG@10, macro-averaged over the 12 datasets
(five are already at or below the cap, so their full corpus is the one already
reported). The elastic distances, the symbolic BM25 retrievers, and the
supervised reference do not scale to the largest corpora and are omitted. In each
column the best value is bold and the second best underlined.}
\label{tab:fullcorpus}
\begin{tabular}{lccc}
\toprule
Method & Capped & Full & $\Delta$ \\
\midrule
CHARM & 0.486 & 0.437 & $-0.049$ \\
CHARM\,$+$\,NR & \textbf{0.523} & \underline{0.442} & $-0.081$ \\
Chronos-2 & 0.471 & 0.437 & $-0.034$ \\
MantisV2 & \underline{0.510} & \textbf{0.447} & $-0.063$ \\
TiRex & 0.483 & 0.434 & $-0.049$ \\
\bottomrule
\end{tabular}
\end{table}

\subsubsection{Normal-residual scoring by embedder}
\label{app:nr-by-embedder}
Table~\ref{tab:app-nr} applies NR scoring to each embedder and
compares it against ED at every pollution level. NR improves
pollution robustness on every embedder it wraps, roughly
halving the degradation in each case, and improves clean accuracy on all of them
except TiRex, whose clean score is essentially unchanged ($0.483 \to 0.481$);
MantisV2\,$+$\,NR is the strongest single
base retriever, clean and under pollution.

\begin{table}[H]\centering
\scriptsize
\setlength{\tabcolsep}{2.5pt}
\renewcommand{\arraystretch}{0.9}
\caption{NR scoring versus ED on each
embedder, NDCG@10 at each pollution level $\rho$, where $\Delta$ is
the drop from $\rho{=}0\%$ to $\rho{=}20\%$ and a smaller-magnitude $\Delta$
means more robust.}
\label{tab:app-nr}
\begin{tabular}{llcccc}
\toprule
Embedder & Scoring & $\rho{=}0\%$ & $\rho{=}10\%$ & $\rho{=}20\%$ & $\Delta$ \\
\midrule
\multirow{2}{*}{CHARM} & ED & 0.486 & 0.461 & 0.437 & $-0.049$ \\
                       & NR & \underline{0.523} & 0.511 & 0.502 & $-0.021$ \\
\addlinespace
\multirow{2}{*}{MantisV2} & ED & 0.510 & 0.474 & 0.454 & $-0.056$ \\
                          & NR & \textbf{0.555} & 0.539 & 0.527 & $-0.028$ \\
\addlinespace
\multirow{2}{*}{Chronos-2} & ED & 0.471 & 0.455 & 0.432 & $-0.039$ \\
                           & NR & 0.474 & 0.467 & 0.452 & $-0.022$ \\
\addlinespace
\multirow{2}{*}{TiRex} & ED & 0.483 & 0.449 & 0.433 & $-0.050$ \\
                       & NR & 0.481 & 0.476 & 0.451 & $-0.029$ \\
\bottomrule
\end{tabular}
\end{table}

\subsubsection{Choosing the fusion leg and mode}
\label{app:fusion-choice}
Fusion involves two choices, namely which complementary leg to pair with the
embedder and which rule combines the two rankings.
Table~\ref{tab:fusion-joint} sweeps both at once, crossing every leg with every
mode for each embedder. Across all seven embedder blocks the best cell is the
DTW-I leg fused by RRF, so the joint optimum is the
same regardless of embedder. RRF is also the best mode on every leg except SBD-D,
which is strongest under weighted sum in each block but never competes for the
top cell, and DTW-I is the best leg under RRF throughout. RRF over a DTW-I leg
is therefore the fusion recipe used in the rest of the paper. The same grid
across pollution levels is in Appendix~\ref{app:fusion-grid}.

\begin{table}[H]\centering
\scriptsize
\setlength{\tabcolsep}{4pt}
\renewcommand{\arraystretch}{0.9}
\caption{Choosing the fusion leg and mode. NDCG@10 at $\rho{=}0\%$,
macro-averaged over the 12 datasets, for every complementary leg (rows) crossed
with every fusion mode (columns), per embedder. Modes are RRF,
WS ($\alpha{=}0.5$), and 2-step cascade
(shortlist 100). Marks are per embedder block, and the DTW-I leg under RRF
is the joint optimum in every block.}
\label{tab:fusion-joint}
\begin{tabular}{llccc}
\toprule
Embedder & Leg & RRF & WS & 2-step \\
\midrule
\multirow{5}{*}{CHARM}
 & DTW-I    & \textbf{0.536} & \underline{0.521} & 0.498 \\
 & DTW-D    & 0.492 & 0.477 & 0.444 \\
 & SBD-D    & 0.445 & 0.454 & 0.397 \\
 & SAX-BM25 & 0.519 & 0.500 & 0.497 \\
 & SFA-BM25 & 0.485 & 0.448 & 0.456 \\
\addlinespace
\multirow{5}{*}{CHARM\,$+$\,NR}
 & DTW-I    & \textbf{0.575} & 0.543 & 0.538 \\
 & DTW-D    & 0.543 & 0.528 & 0.500 \\
 & SBD-D    & 0.476 & 0.520 & 0.438 \\
 & SAX-BM25 & \underline{0.551} & 0.543 & 0.537 \\
 & SFA-BM25 & 0.510 & 0.493 & 0.491 \\
\addlinespace
\multirow{5}{*}{MantisV2}
 & DTW-I    & \textbf{0.532} & 0.509 & 0.509 \\
 & DTW-D    & 0.507 & 0.488 & 0.471 \\
 & SBD-D    & 0.462 & 0.490 & 0.410 \\
 & SAX-BM25 & \underline{0.531} & 0.495 & 0.502 \\
 & SFA-BM25 & 0.496 & 0.448 & 0.465 \\
\addlinespace
\multirow{5}{*}{MantisV2\,$+$\,NR}
 & DTW-I    & \textbf{0.590} & 0.572 & 0.561 \\
 & DTW-D    & 0.562 & 0.556 & 0.519 \\
 & SBD-D    & 0.493 & 0.541 & 0.440 \\
 & SAX-BM25 & \underline{0.582} & 0.541 & 0.540 \\
 & SFA-BM25 & 0.543 & 0.493 & 0.498 \\
\addlinespace
\multirow{5}{*}{Chronos-2\,$+$\,NR}
 & DTW-I    & \textbf{0.551} & 0.514 & 0.531 \\
 & DTW-D    & 0.528 & 0.500 & 0.497 \\
 & SBD-D    & 0.442 & 0.486 & 0.413 \\
 & SAX-BM25 & \underline{0.538} & 0.533 & 0.529 \\
 & SFA-BM25 & 0.499 & 0.485 & 0.482 \\
\addlinespace
\multirow{5}{*}{TiRex}
 & DTW-I    & \textbf{0.520} & 0.501 & 0.484 \\
 & DTW-D    & 0.486 & 0.470 & 0.444 \\
 & SBD-D    & 0.445 & 0.463 & 0.399 \\
 & SAX-BM25 & \underline{0.511} & 0.507 & 0.506 \\
 & SFA-BM25 & 0.479 & 0.463 & 0.471 \\
\addlinespace
\multirow{5}{*}{TiRex\,$+$\,NR}
 & DTW-I    & \textbf{0.554} & 0.520 & \underline{0.540} \\
 & DTW-D    & 0.521 & 0.504 & 0.497 \\
 & SBD-D    & 0.442 & 0.492 & 0.413 \\
 & SAX-BM25 & 0.530 & 0.528 & 0.520 \\
 & SFA-BM25 & 0.491 & 0.480 & 0.472 \\
\bottomrule
\end{tabular}
\end{table}

\subsubsection{Complete fusion grid}
\label{app:fusion-grid}
Table~\ref{tab:app-hybrids} gives the full fusion grid, every embedder (ED and
NR) crossed with every shape or symbolic leg, at all three pollution levels. The
best leg is DTW-I, MantisV2\,$+$\,NR is the strongest fused pool, and the gains are
consistent across embedders rather than specific to one.

\begin{table}[H]\centering
\scriptsize
\setlength{\tabcolsep}{2.5pt}
\renewcommand{\arraystretch}{0.9}
\caption{Complete fusion grid. RRF of each embedder (rows) with
each shape or symbolic leg (column groups), NDCG@10 at pollution levels
$\rho{=}0/\!\approx\!10/\!\approx\!20\%$,
macro-averaged over the 12 datasets. Both the ED and
NR variant of each embedder are shown, so the fusion gains are
not specific to a single embedder. Marks are on the clean ($\rho{=}0\%$)
sub-column of each leg; DTW-I is the best complementary
leg and MantisV2\,$+$\,NR the strongest fused pool.}
\label{tab:app-hybrids}
\resizebox{\textwidth}{!}{%
\begin{tabular}{l *{5}{ccc}}
\toprule
& \multicolumn{3}{c}{$|$\,DTW-I} & \multicolumn{3}{c}{$|$\,DTW-D}
& \multicolumn{3}{c}{$|$\,SBD-D} & \multicolumn{3}{c}{$|$\,SAX-BM25}
& \multicolumn{3}{c}{$|$\,SFA-BM25} \\
\cmidrule(lr){2-4}\cmidrule(lr){5-7}\cmidrule(lr){8-10}\cmidrule(lr){11-13}\cmidrule(lr){14-16}
Embedder
& $\rho{=}0\%$ & $10\%$ & $20\%$ & $0\%$ & $10\%$ & $20\%$ & $0\%$ & $10\%$ & $20\%$
& $0\%$ & $10\%$ & $20\%$ & $0\%$ & $10\%$ & $20\%$ \\
\midrule
CHARM & \underline{0.536} & 0.493 & 0.453 & 0.492 & 0.461 & 0.425 & 0.445 & 0.426 & 0.397 & 0.519 & 0.492 & 0.459 & 0.485 & 0.459 & 0.430 \\
CHARM\,$+$\,NR & \underline{0.575} & 0.549 & 0.514 & \underline{0.543} & 0.524 & 0.491 & \underline{0.476} & 0.456 & 0.432 & \underline{0.551} & 0.540 & 0.521 & \underline{0.510} & 0.503 & 0.480 \\
MantisV2 & 0.532 & 0.490 & 0.452 & 0.507 & 0.468 & 0.432 & 0.462 & 0.440 & 0.408 & 0.531 & 0.500 & 0.472 & 0.496 & 0.466 & 0.447 \\
MantisV2\,$+$\,NR & \textbf{0.590} & 0.556 & 0.534 & \textbf{0.562} & 0.533 & 0.509 & \textbf{0.493} & 0.475 & 0.454 & \textbf{0.582} & 0.560 & 0.540 & \textbf{0.543} & 0.522 & 0.507 \\
Chronos-2\,$+$\,NR & 0.551 & 0.521 & 0.498 & 0.528 & 0.508 & 0.485 & 0.442 & 0.422 & 0.404 & 0.538 & 0.527 & 0.510 & 0.499 & 0.490 & 0.470 \\
TiRex & 0.520 & 0.487 & 0.447 & 0.486 & 0.455 & 0.418 & 0.445 & 0.426 & 0.392 & 0.511 & 0.490 & 0.456 & 0.479 & 0.456 & 0.430 \\
TiRex\,$+$\,NR & 0.554 & 0.524 & 0.495 & 0.521 & 0.497 & 0.470 & 0.442 & 0.425 & 0.399 & 0.530 & 0.522 & 0.496 & 0.491 & 0.478 & 0.456 \\
\bottomrule
\end{tabular}%
}
\end{table}

\subsubsection{Reranking a fixed pool}
\label{app:reranking}
Table~\ref{tab:rerankers} reports each reranker on each pool at $\rho{=}0\%$.
The label-aware GPC reranker gives the largest gain on most pools, with
Majority-Vote (MajVote) ahead on the CHARM\,$+$\,NR and TiRex\,$+$\,NR pools; the
label-free rerankers give little or no gain, since PRF moves the score only
marginally and Purity hurts.

\begin{table}[H]
\centering
\scriptsize
\setlength{\tabcolsep}{4pt}
\renewcommand{\arraystretch}{0.9}
\caption{Reranking a fixed pool, NDCG@10. Label-free rerankers are PRF and
Purity; label-aware ones are GPC, MajVote, and QPurity.
Language-model rerankers in
Table~\ref{tab:llm}.}
\label{tab:rerankers}
\begin{tabular}{lcccccc}
\toprule
Pool & Base & $+$GPC & $+$MajVote & $+$QPurity & $+$PRF & $+$Purity \\
\midrule
CHARM & 0.486 & \textbf{0.556} & \underline{0.538} & 0.511 & 0.455 & 0.386 \\
CHARM\,$+$\,NR & 0.523 & \underline{0.608} & \textbf{0.612} & 0.558 & 0.528 & 0.461 \\
DTW-I & 0.480 & \textbf{0.573} & \underline{0.525} & 0.500 & 0.483 & 0.342 \\
MantisV2 & 0.510 & \textbf{0.604} & \underline{0.571} & 0.528 & 0.501 & 0.445 \\
MantisV2\,$+$\,NR & 0.555 & \textbf{0.645} & \underline{0.616} & 0.562 & 0.511 & 0.489 \\
Chronos-2\,$+$\,NR & 0.474 & \textbf{0.582} & \underline{0.545} & 0.512 & 0.463 & 0.432 \\
TiRex & 0.483 & \textbf{0.528} & \underline{0.517} & 0.485 & 0.463 & 0.394 \\
TiRex\,$+$\,NR & 0.481 & \underline{0.541} & \textbf{0.568} & 0.518 & 0.445 & 0.426 \\
\bottomrule
\end{tabular}
\end{table}

\subsubsection{Complete reranker grid}
\label{app:reranker-grid}
Table~\ref{tab:app-rerankers} extends the reranker comparison to every pool
(ED and NR) at all three pollution levels, confirming that the GPC gain holds
across embedders and pollution and is not specific to the CHARM-based NR pool.
NR's advantage also survives reranking, since MantisV2\,$+$\,NR\,$+$\,GPC (0.645) exceeds the
MantisV2\,$+$\,ED\,$+$\,GPC reference (0.604), so the strongest reranked pool is
itself an NR variant.

\begin{table}[H]\centering
\scriptsize
\setlength{\tabcolsep}{2.5pt}
\renewcommand{\arraystretch}{0.9}
\caption{Complete reranker grid. Each reranker on each pool, NDCG@10 at pollution
levels $\rho{=}0/\!\approx\!10/\!\approx\!20\%$, macro-averaged over the 12 datasets.
Pools include both the ED and NR variant of each
embedder, so the reranking gains are shown to hold across embedders and are not
specific to the CHARM-based NR. The label-free rerankers are PRF and Purity; the
label-aware ones are GPC, MajVote,
and QPurity. Marks are on the clean ($\rho{=}0\%$) sub-column of each reranker.}
\label{tab:app-rerankers}
\resizebox{\textwidth}{!}{%
\begin{tabular}{l *{5}{ccc}}
\toprule
& \multicolumn{3}{c}{$+$GPC} & \multicolumn{3}{c}{$+$MajVote}
& \multicolumn{3}{c}{$+$QPurity} & \multicolumn{3}{c}{$+$PRF}
& \multicolumn{3}{c}{$+$Purity} \\
\cmidrule(lr){2-4}\cmidrule(lr){5-7}\cmidrule(lr){8-10}\cmidrule(lr){11-13}\cmidrule(lr){14-16}
Pool
& $\rho{=}0\%$ & $10\%$ & $20\%$ & $0\%$ & $10\%$ & $20\%$ & $0\%$ & $10\%$ & $20\%$
& $0\%$ & $10\%$ & $20\%$ & $0\%$ & $10\%$ & $20\%$ \\
\midrule
CHARM & 0.556 & 0.562 & 0.517 & 0.538 & 0.536 & 0.500 & 0.511 & 0.513 & 0.481 & 0.455 & 0.443 & 0.407 & 0.386 & 0.383 & 0.332 \\
CHARM\,$+$\,NR & \underline{0.608} & 0.608 & 0.578 & \underline{0.612} & 0.614 & 0.579 & \underline{0.558} & 0.566 & 0.535 & \textbf{0.528} & 0.524 & 0.499 & \underline{0.461} & 0.464 & 0.427 \\
MantisV2 & 0.604 & 0.556 & 0.532 & 0.571 & 0.548 & 0.522 & 0.528 & 0.507 & 0.486 & 0.501 & 0.470 & 0.450 & 0.445 & 0.431 & 0.413 \\
MantisV2\,$+$\,NR & \textbf{0.645} & 0.613 & 0.582 & \textbf{0.616} & 0.610 & 0.589 & \textbf{0.562} & 0.549 & 0.538 & \underline{0.511} & 0.498 & 0.490 & \textbf{0.489} & 0.478 & 0.458 \\
Chronos-2\,$+$\,NR & 0.582 & 0.573 & 0.563 & 0.545 & 0.546 & 0.534 & 0.512 & 0.508 & 0.492 & 0.463 & 0.464 & 0.451 & 0.432 & 0.440 & 0.417 \\
TiRex & 0.528 & 0.514 & 0.486 & 0.517 & 0.513 & 0.456 & 0.485 & 0.488 & 0.440 & 0.463 & 0.450 & 0.411 & 0.394 & 0.400 & 0.359 \\
TiRex\,$+$\,NR & 0.541 & 0.525 & 0.509 & 0.568 & 0.566 & 0.551 & 0.518 & 0.523 & 0.498 & 0.445 & 0.447 & 0.424 & 0.426 & 0.427 & 0.392 \\
DTW-I & 0.573 & 0.517 & 0.464 & 0.525 & 0.468 & 0.423 & 0.500 & 0.503 & 0.416 & 0.483 & 0.427 & 0.392 & 0.342 & 0.324 & 0.286 \\
\bottomrule
\end{tabular}%
}
\end{table}

\subsubsection{Composing fusion and reranking}
\label{app:composition}
Table~\ref{tab:stack} composes fusion and reranking, reporting the fused pool and
the same pool reranked with GPC across pollution levels. The two additions stack
close to additively, GPC adds about the same margin to a fused pool as it does to
an unfused one, and CHARM\,$+$\,NR\,$|$\,DTW-I\,$+$\,GPC is the strongest system at every
pollution level. Normal-residual scoring is what lifts these stacks above their
plain-embedding counterparts (e.g.\ CHARM\,$|$\,DTW-I\,$+$\,GPC).

\begin{table}[H]
\centering
\scriptsize
\setlength{\tabcolsep}{2.5pt}
\renewcommand{\arraystretch}{0.9}
\caption{Composing fusion and reranking. NDCG@10 of the fused pool (fusion
only, at $\rho{=}0\%$) and of the same pool reranked with GPC across pollution
levels $\rho$, macro-averaged over the 12 datasets. In each column the best
value is bold and the second best underlined.}
\label{tab:stack}
\begin{tabular}{lcccc}
\toprule
& Fusion & \multicolumn{3}{c}{Fusion $+$ GPC} \\
\cmidrule(lr){2-2}\cmidrule(lr){3-5}
Fusion & $\rho{=}0\%$ & $\rho{=}0\%$ & $\rho{=}10\%$ & $\rho{=}20\%$ \\
\midrule
CHARM\,$+$\,NR\,$|$\,DTW-I & \underline{0.575} & \textbf{0.687} & \textbf{0.640} & \textbf{0.619} \\
MantisV2\,$+$\,NR\,$|$\,DTW-I & \textbf{0.590} & \underline{0.665} & \underline{0.641} & \underline{0.604} \\
CHARM\,$|$\,DTW-I & 0.536 & 0.631 & 0.582 & 0.535 \\
MantisV2\,$|$\,DTW-I & 0.532 & 0.609 & 0.564 & 0.552 \\
\bottomrule
\end{tabular}
\end{table}

\subsection{Complete Per-Dataset Results}
\label{app:alldata}

This appendix reports the full metric grid for every non-fusion method in
READ-Bench, with one table per method listing all 12 datasets (and their
macro-average) across the three pollution levels. Metrics are precision (P),
hit rate (HR), and NDCG at cutoffs $\{1,5,10,20\}$, plus macro-F1. In every
table the column headers abbreviate N$=$NDCG and mF1$=$macro-F1, and the
\emph{macro} row is the macro-average over the 12 datasets. Fusion combinations
are reported in Appendix~\ref{app:full-results}.

\begin{table}[H]\centering
\tiny
\setlength{\tabcolsep}{2.2pt}
\renewcommand{\arraystretch}{0.85}
\caption{Full metric grid for \textbf{Random}, per dataset and macro-averaged, at each pollution level.}
\label{tab:a8-random}
\begin{tabular}{lccccccccccc}
\toprule
 & P@1 & P@5 & P@10 & HR@1 & HR@5 & HR@10 & N@1 & N@5 & N@10 & N@20 & mF1 \\
\midrule
\multicolumn{12}{l}{\emph{$\rho{=}0\%$}} \\
CTSR & 0.000 & 0.008 & 0.012 & 0.000 & 0.040 & 0.120 & 0.000 & 0.005 & 0.009 & 0.014 & 0.000 \\
DAMADICS & 0.333 & 0.300 & 0.317 & 0.333 & 0.611 & 0.667 & 0.333 & 0.299 & 0.309 & 0.325 & 0.170 \\
Exathlon & 0.180 & 0.224 & 0.219 & 0.180 & 0.740 & 0.890 & 0.180 & 0.221 & 0.219 & 0.211 & 0.134 \\
HAI & 0.179 & 0.214 & 0.225 & 0.179 & 0.679 & 0.893 & 0.179 & 0.203 & 0.215 & 0.199 & 0.163 \\
MIT-BIH & 0.300 & 0.340 & 0.364 & 0.300 & 0.840 & 0.960 & 0.300 & 0.334 & 0.353 & 0.351 & 0.214 \\
Petrobras 3W & 0.120 & 0.140 & 0.151 & 0.120 & 0.470 & 0.730 & 0.120 & 0.138 & 0.146 & 0.149 & 0.130 \\
\bottomrule
\end{tabular}
\end{table}

\begin{table}[H]\centering
\tiny
\setlength{\tabcolsep}{2.2pt}
\renewcommand{\arraystretch}{0.85}
\caption{Full metric grid for \textbf{Majority}, per dataset and macro-averaged, at each pollution level.}
\label{tab:a8-majority}
\begin{tabular}{lccccccccccc}
\toprule
 & P@1 & P@5 & P@10 & HR@1 & HR@5 & HR@10 & N@1 & N@5 & N@10 & N@20 & mF1 \\
\midrule
\multicolumn{12}{l}{\emph{$\rho{=}0\%$}} \\
CTSR & 0.010 & 0.010 & 0.010 & 0.010 & 0.010 & 0.010 & 0.010 & 0.010 & 0.010 & 0.014 & 0.000 \\
DAMADICS & 0.222 & 0.222 & 0.222 & 0.222 & 0.222 & 0.222 & 0.222 & 0.222 & 0.222 & 0.222 & 0.091 \\
Exathlon & 0.420 & 0.420 & 0.420 & 0.420 & 0.420 & 0.420 & 0.420 & 0.420 & 0.420 & 0.420 & 0.099 \\
HAI & 0.214 & 0.214 & 0.214 & 0.214 & 0.214 & 0.214 & 0.214 & 0.214 & 0.214 & 0.214 & 0.059 \\
MIT-BIH & 0.370 & 0.370 & 0.370 & 0.370 & 0.370 & 0.370 & 0.370 & 0.370 & 0.370 & 0.370 & 0.135 \\
Petrobras 3W & 0.240 & 0.240 & 0.240 & 0.240 & 0.240 & 0.240 & 0.240 & 0.240 & 0.240 & 0.240 & 0.043 \\
\bottomrule
\end{tabular}
\end{table}

\begin{table}[H]\centering
\tiny
\setlength{\tabcolsep}{2.2pt}
\renewcommand{\arraystretch}{0.85}
\caption{Full metric grid for \textbf{ED}, per dataset and macro-averaged, at each pollution level.}
\label{tab:a8-ed}
\begin{tabular}{lccccccccccc}
\toprule
 & P@1 & P@5 & P@10 & HR@1 & HR@5 & HR@10 & N@1 & N@5 & N@10 & N@20 & mF1 \\
\midrule
\multicolumn{12}{l}{\emph{$\rho{=}0\%$}} \\
CTSR & 0.650 & 0.516 & 0.430 & 0.650 & 0.850 & 0.890 & 0.650 & 0.540 & 0.472 & 0.552 & 0.637 \\
DAMADICS & 0.389 & 0.411 & 0.333 & 0.389 & 0.722 & 0.889 & 0.389 & 0.411 & 0.370 & 0.406 & 0.302 \\
Exathlon & 0.860 & 0.760 & 0.732 & 0.860 & 0.920 & 0.950 & 0.860 & 0.782 & 0.755 & 0.720 & 0.744 \\
HAI & 0.286 & 0.350 & 0.354 & 0.286 & 0.679 & 0.786 & 0.286 & 0.337 & 0.349 & 0.337 & 0.514 \\
MIT-BIH & 0.430 & 0.496 & 0.543 & 0.430 & 0.710 & 0.790 & 0.430 & 0.484 & 0.521 & 0.548 & 0.224 \\
Petrobras 3W & 0.680 & 0.564 & 0.507 & 0.680 & 0.880 & 0.940 & 0.680 & 0.588 & 0.539 & 0.484 & 0.530 \\
\bottomrule
\end{tabular}
\end{table}

\begin{table}[H]\centering
\tiny
\setlength{\tabcolsep}{2.2pt}
\renewcommand{\arraystretch}{0.85}
\caption{Full metric grid for \textbf{SBD-D}, per dataset and macro-averaged, at each pollution level.}
\label{tab:a8-sbd-d}
\begin{tabular}{lccccccccccc}
\toprule
 & P@1 & P@5 & P@10 & HR@1 & HR@5 & HR@10 & N@1 & N@5 & N@10 & N@20 & mF1 \\
\midrule
\multicolumn{12}{l}{\emph{$\rho{=}0\%$}} \\
CTSR & 0.010 & 0.454 & 0.403 & 0.010 & 0.830 & 0.840 & 0.010 & 0.386 & 0.375 & 0.454 & 0.572 \\
DAMADICS & 0.389 & 0.233 & 0.217 & 0.389 & 0.444 & 0.556 & 0.389 & 0.263 & 0.247 & 0.255 & 0.095 \\
Exathlon & 0.370 & 0.362 & 0.306 & 0.370 & 0.450 & 0.500 & 0.370 & 0.367 & 0.327 & 0.275 & 0.427 \\
HAI & 0.286 & 0.286 & 0.279 & 0.286 & 0.643 & 0.750 & 0.286 & 0.279 & 0.280 & 0.272 & 0.269 \\
MIT-BIH & 0.830 & 0.832 & 0.808 & 0.830 & 0.950 & 0.950 & 0.830 & 0.832 & 0.816 & 0.777 & 0.673 \\
Petrobras 3W & 0.430 & 0.388 & 0.352 & 0.430 & 0.590 & 0.720 & 0.430 & 0.399 & 0.370 & 0.333 & 0.350 \\
\bottomrule
\end{tabular}
\end{table}

\begin{table}[H]\centering
\tiny
\setlength{\tabcolsep}{2.2pt}
\renewcommand{\arraystretch}{0.85}
\caption{Full metric grid for \textbf{DTW-I}, per dataset and macro-averaged, at each pollution level.}
\label{tab:a8-dtw-i}
\begin{tabular}{lccccccccccc}
\toprule
 & P@1 & P@5 & P@10 & HR@1 & HR@5 & HR@10 & N@1 & N@5 & N@10 & N@20 & mF1 \\
\midrule
\multicolumn{12}{l}{\emph{$\rho{=}0\%$}} \\
CTSR & 0.730 & 0.572 & 0.473 & 0.730 & 0.840 & 0.890 & 0.730 & 0.605 & 0.526 & 0.583 & 0.610 \\
DAMADICS & 0.667 & 0.622 & 0.583 & 0.667 & 0.722 & 0.833 & 0.667 & 0.641 & 0.635 & 0.651 & 0.506 \\
Exathlon & 0.350 & 0.318 & 0.307 & 0.350 & 0.400 & 0.500 & 0.350 & 0.325 & 0.315 & 0.329 & 0.372 \\
HAI & 0.107 & 0.121 & 0.154 & 0.107 & 0.464 & 0.786 & 0.107 & 0.115 & 0.141 & 0.171 & 0.032 \\
MIT-BIH & 0.560 & 0.510 & 0.522 & 0.560 & 0.800 & 0.900 & 0.560 & 0.523 & 0.527 & 0.529 & 0.445 \\
Petrobras 3W & 0.930 & 0.876 & 0.847 & 0.930 & 0.940 & 0.960 & 0.930 & 0.888 & 0.864 & 0.823 & 0.879 \\
\bottomrule
\end{tabular}
\end{table}

\begin{table}[H]\centering
\tiny
\setlength{\tabcolsep}{2.2pt}
\renewcommand{\arraystretch}{0.85}
\caption{Full metric grid for \textbf{DTW-D}, per dataset and macro-averaged, at each pollution level.}
\label{tab:a8-dtw-d}
\begin{tabular}{lccccccccccc}
\toprule
 & P@1 & P@5 & P@10 & HR@1 & HR@5 & HR@10 & N@1 & N@5 & N@10 & N@20 & mF1 \\
\midrule
\multicolumn{12}{l}{\emph{$\rho{=}0\%$}} \\
CTSR & 0.730 & 0.572 & 0.473 & 0.730 & 0.840 & 0.890 & 0.730 & 0.605 & 0.526 & 0.583 & 0.610 \\
DAMADICS & 0.500 & 0.356 & 0.300 & 0.500 & 0.667 & 0.667 & 0.500 & 0.384 & 0.350 & 0.318 & 0.295 \\
Exathlon & 0.400 & 0.388 & 0.357 & 0.400 & 0.460 & 0.500 & 0.400 & 0.394 & 0.370 & 0.361 & 0.408 \\
HAI & 0.107 & 0.114 & 0.136 & 0.107 & 0.393 & 0.571 & 0.107 & 0.114 & 0.131 & 0.156 & 0.123 \\
MIT-BIH & 0.470 & 0.442 & 0.426 & 0.470 & 0.630 & 0.720 & 0.470 & 0.450 & 0.436 & 0.430 & 0.389 \\
Petrobras 3W & 0.930 & 0.888 & 0.859 & 0.930 & 0.950 & 0.950 & 0.930 & 0.900 & 0.875 & 0.827 & 0.894 \\
\bottomrule
\end{tabular}
\end{table}

\begin{table}[H]\centering
\tiny
\setlength{\tabcolsep}{2.2pt}
\renewcommand{\arraystretch}{0.85}
\caption{Full metric grid for \textbf{SAX-BM25}, per dataset and macro-averaged, at each pollution level.}
\label{tab:a8-sax-bm25}
\begin{tabular}{lccccccccccc}
\toprule
 & P@1 & P@5 & P@10 & HR@1 & HR@5 & HR@10 & N@1 & N@5 & N@10 & N@20 & mF1 \\
\midrule
\multicolumn{12}{l}{\emph{$\rho{=}0\%$}} \\
CTSR & 0.410 & 0.302 & 0.257 & 0.410 & 0.730 & 0.810 & 0.410 & 0.324 & 0.285 & 0.333 & 0.268 \\
DAMADICS & 0.333 & 0.311 & 0.289 & 0.333 & 0.722 & 0.889 & 0.333 & 0.322 & 0.315 & 0.333 & 0.231 \\
Exathlon & 0.870 & 0.856 & 0.849 & 0.870 & 0.970 & 1.000 & 0.870 & 0.860 & 0.854 & 0.822 & 0.825 \\
HAI & 0.429 & 0.421 & 0.393 & 0.429 & 0.750 & 0.857 & 0.429 & 0.425 & 0.403 & 0.371 & 0.360 \\
MIT-BIH & 0.680 & 0.644 & 0.639 & 0.680 & 0.900 & 0.950 & 0.680 & 0.654 & 0.647 & 0.621 & 0.697 \\
Petrobras 3W & 0.670 & 0.614 & 0.559 & 0.670 & 0.940 & 0.970 & 0.670 & 0.628 & 0.586 & 0.557 & 0.647 \\
\bottomrule
\end{tabular}
\end{table}

\begin{table}[H]\centering
\tiny
\setlength{\tabcolsep}{2.2pt}
\renewcommand{\arraystretch}{0.85}
\caption{Full metric grid for \textbf{SFA-BM25}, per dataset and macro-averaged, at each pollution level.}
\label{tab:a8-sfa-bm25}
\begin{tabular}{lccccccccccc}
\toprule
 & P@1 & P@5 & P@10 & HR@1 & HR@5 & HR@10 & N@1 & N@5 & N@10 & N@20 & mF1 \\
\midrule
\multicolumn{12}{l}{\emph{$\rho{=}0\%$}} \\
CTSR & 0.360 & 0.308 & 0.262 & 0.360 & 0.680 & 0.750 & 0.360 & 0.318 & 0.283 & 0.335 & 0.313 \\
DAMADICS & 0.444 & 0.333 & 0.306 & 0.444 & 0.667 & 0.833 & 0.444 & 0.362 & 0.345 & 0.346 & 0.300 \\
Exathlon & 0.790 & 0.732 & 0.695 & 0.790 & 0.960 & 0.980 & 0.790 & 0.744 & 0.714 & 0.658 & 0.800 \\
HAI & 0.357 & 0.364 & 0.325 & 0.357 & 0.750 & 0.786 & 0.357 & 0.368 & 0.340 & 0.320 & 0.282 \\
MIT-BIH & 0.480 & 0.468 & 0.433 & 0.480 & 0.800 & 0.940 & 0.480 & 0.468 & 0.444 & 0.429 & 0.379 \\
Petrobras 3W & 0.660 & 0.538 & 0.510 & 0.660 & 0.870 & 0.940 & 0.660 & 0.564 & 0.536 & 0.478 & 0.576 \\
\bottomrule
\end{tabular}
\end{table}

\begin{table}[H]\centering
\tiny
\setlength{\tabcolsep}{2.2pt}
\renewcommand{\arraystretch}{0.85}
\caption{Full metric grid for \textbf{CHARM}, per dataset and macro-averaged, at each pollution level.}
\label{tab:a8-cosine}
\begin{tabular}{lccccccccccc}
\toprule
 & P@1 & P@5 & P@10 & HR@1 & HR@5 & HR@10 & N@1 & N@5 & N@10 & N@20 & mF1 \\
\midrule
\multicolumn{12}{l}{\emph{$\rho{=}0\%$}} \\
CTSR & 0.650 & 0.516 & 0.430 & 0.650 & 0.850 & 0.890 & 0.650 & 0.540 & 0.472 & 0.552 & 0.637 \\
DAMADICS & 0.389 & 0.411 & 0.333 & 0.389 & 0.722 & 0.889 & 0.389 & 0.411 & 0.370 & 0.406 & 0.302 \\
Exathlon & 0.860 & 0.760 & 0.732 & 0.860 & 0.920 & 0.950 & 0.860 & 0.782 & 0.755 & 0.720 & 0.744 \\
HAI & 0.286 & 0.350 & 0.354 & 0.286 & 0.679 & 0.786 & 0.286 & 0.337 & 0.349 & 0.337 & 0.514 \\
MIT-BIH & 0.430 & 0.496 & 0.543 & 0.430 & 0.710 & 0.790 & 0.430 & 0.484 & 0.521 & 0.548 & 0.224 \\
Petrobras 3W & 0.680 & 0.564 & 0.507 & 0.680 & 0.880 & 0.940 & 0.680 & 0.588 & 0.539 & 0.484 & 0.530 \\
\bottomrule
\end{tabular}
\end{table}

\begin{table}[H]\centering
\tiny
\setlength{\tabcolsep}{2.2pt}
\renewcommand{\arraystretch}{0.85}
\caption{Full metric grid for \textbf{CHARM\,$+$\,NR}, per dataset and macro-averaged, at each pollution level.}
\label{tab:a8-delta}
\begin{tabular}{lccccccccccc}
\toprule
 & P@1 & P@5 & P@10 & HR@1 & HR@5 & HR@10 & N@1 & N@5 & N@10 & N@20 & mF1 \\
\midrule
\multicolumn{12}{l}{\emph{$\rho{=}0\%$}} \\
CTSR & 0.630 & 0.512 & 0.424 & 0.630 & 0.800 & 0.870 & 0.630 & 0.541 & 0.469 & 0.544 & 0.629 \\
DAMADICS & 0.556 & 0.489 & 0.478 & 0.556 & 0.778 & 1.000 & 0.556 & 0.499 & 0.518 & 0.521 & 0.354 \\
Exathlon & 0.920 & 0.850 & 0.818 & 0.920 & 0.990 & 1.000 & 0.920 & 0.863 & 0.837 & 0.782 & 0.886 \\
HAI & 0.393 & 0.321 & 0.332 & 0.393 & 0.643 & 0.786 & 0.393 & 0.331 & 0.338 & 0.326 & 0.462 \\
MIT-BIH & 0.460 & 0.588 & 0.638 & 0.460 & 0.790 & 0.880 & 0.460 & 0.565 & 0.608 & 0.650 & 0.328 \\
Petrobras 3W & 0.680 & 0.526 & 0.481 & 0.680 & 0.800 & 0.890 & 0.680 & 0.553 & 0.511 & 0.462 & 0.530 \\
\bottomrule
\end{tabular}
\end{table}

\begin{table}[H]\centering
\tiny
\setlength{\tabcolsep}{2.2pt}
\renewcommand{\arraystretch}{0.85}
\caption{Full metric grid for \textbf{Chronos-2}, per dataset and macro-averaged, at each pollution level.}
\label{tab:a8-chronos2-ed}
\begin{tabular}{lccccccccccc}
\toprule
 & P@1 & P@5 & P@10 & HR@1 & HR@5 & HR@10 & N@1 & N@5 & N@10 & N@20 & mF1 \\
\midrule
\multicolumn{12}{l}{\emph{$\rho{=}0\%$}} \\
CTSR & 0.510 & 0.452 & 0.365 & 0.510 & 0.810 & 0.830 & 0.510 & 0.465 & 0.401 & 0.453 & 0.496 \\
DAMADICS & 0.111 & 0.189 & 0.200 & 0.111 & 0.611 & 0.611 & 0.111 & 0.176 & 0.191 & 0.210 & 0.029 \\
Exathlon & 0.960 & 0.912 & 0.881 & 0.960 & 0.990 & 0.990 & 0.960 & 0.921 & 0.896 & 0.858 & 0.897 \\
HAI & 0.321 & 0.321 & 0.311 & 0.321 & 0.750 & 0.964 & 0.321 & 0.325 & 0.317 & 0.322 & 0.346 \\
MIT-BIH & 0.720 & 0.708 & 0.719 & 0.720 & 0.860 & 0.890 & 0.720 & 0.708 & 0.716 & 0.705 & 0.393 \\
Petrobras 3W & 0.680 & 0.534 & 0.476 & 0.680 & 0.840 & 0.880 & 0.680 & 0.560 & 0.510 & 0.458 & 0.594 \\
\bottomrule
\end{tabular}
\end{table}

\begin{table}[H]\centering
\tiny
\setlength{\tabcolsep}{2.2pt}
\renewcommand{\arraystretch}{0.85}
\caption{Full metric grid for \textbf{Chronos-2\,$+$\,NR}, per dataset and macro-averaged, at each pollution level.}
\label{tab:a8-chronos2-delta}
\begin{tabular}{lccccccccccc}
\toprule
 & P@1 & P@5 & P@10 & HR@1 & HR@5 & HR@10 & N@1 & N@5 & N@10 & N@20 & mF1 \\
\midrule
\multicolumn{12}{l}{\emph{$\rho{=}0\%$}} \\
CTSR & 0.600 & 0.444 & 0.357 & 0.600 & 0.770 & 0.820 & 0.600 & 0.474 & 0.403 & 0.463 & 0.535 \\
DAMADICS & 0.167 & 0.233 & 0.267 & 0.167 & 0.611 & 0.889 & 0.167 & 0.230 & 0.261 & 0.292 & 0.163 \\
Exathlon & 0.910 & 0.852 & 0.810 & 0.910 & 0.950 & 0.980 & 0.910 & 0.860 & 0.828 & 0.781 & 0.748 \\
HAI & 0.357 & 0.364 & 0.354 & 0.357 & 0.714 & 0.786 & 0.357 & 0.363 & 0.357 & 0.359 & 0.369 \\
MIT-BIH & 0.680 & 0.752 & 0.759 & 0.680 & 0.870 & 0.920 & 0.680 & 0.739 & 0.748 & 0.745 & 0.416 \\
Petrobras 3W & 0.580 & 0.484 & 0.439 & 0.580 & 0.850 & 0.910 & 0.580 & 0.510 & 0.470 & 0.421 & 0.549 \\
\bottomrule
\end{tabular}
\end{table}

\begin{table}[H]\centering
\tiny
\setlength{\tabcolsep}{2.2pt}
\renewcommand{\arraystretch}{0.85}
\caption{Full metric grid for \textbf{MantisV2}, per dataset and macro-averaged, at each pollution level.}
\label{tab:a8-mantisv2-ed}
\begin{tabular}{lccccccccccc}
\toprule
 & P@1 & P@5 & P@10 & HR@1 & HR@5 & HR@10 & N@1 & N@5 & N@10 & N@20 & mF1 \\
\midrule
\multicolumn{12}{l}{\emph{$\rho{=}0\%$}} \\
CTSR & 0.590 & 0.522 & 0.411 & 0.590 & 0.780 & 0.850 & 0.590 & 0.539 & 0.456 & 0.528 & 0.596 \\
DAMADICS & 0.500 & 0.578 & 0.611 & 0.500 & 0.944 & 1.000 & 0.500 & 0.555 & 0.610 & 0.622 & 0.359 \\
Exathlon & 0.950 & 0.934 & 0.908 & 0.950 & 1.000 & 1.000 & 0.950 & 0.939 & 0.919 & 0.897 & 0.964 \\
HAI & 0.536 & 0.429 & 0.396 & 0.536 & 0.714 & 0.964 & 0.536 & 0.460 & 0.432 & 0.397 & 0.539 \\
MIT-BIH & 0.310 & 0.328 & 0.333 & 0.310 & 0.400 & 0.450 & 0.310 & 0.325 & 0.329 & 0.324 & 0.355 \\
Petrobras 3W & 0.860 & 0.816 & 0.766 & 0.860 & 0.980 & 0.980 & 0.860 & 0.827 & 0.788 & 0.709 & 0.776 \\
\bottomrule
\end{tabular}
\end{table}

\begin{table}[H]\centering
\tiny
\setlength{\tabcolsep}{2.2pt}
\renewcommand{\arraystretch}{0.85}
\caption{Full metric grid for \textbf{MantisV2\,$+$\,NR}, per dataset and macro-averaged, at each pollution level.}
\label{tab:a8-mantisv2-delta}
\begin{tabular}{lccccccccccc}
\toprule
 & P@1 & P@5 & P@10 & HR@1 & HR@5 & HR@10 & N@1 & N@5 & N@10 & N@20 & mF1 \\
\midrule
\multicolumn{12}{l}{\emph{$\rho{=}0\%$}} \\
CTSR & 0.610 & 0.508 & 0.410 & 0.610 & 0.770 & 0.850 & 0.610 & 0.534 & 0.458 & 0.527 & 0.580 \\
DAMADICS & 0.611 & 0.578 & 0.472 & 0.611 & 0.778 & 0.889 & 0.611 & 0.596 & 0.544 & 0.540 & 0.389 \\
Exathlon & 0.940 & 0.938 & 0.924 & 0.940 & 0.970 & 0.980 & 0.940 & 0.939 & 0.929 & 0.914 & 0.958 \\
HAI & 0.357 & 0.307 & 0.357 & 0.357 & 0.464 & 0.786 & 0.357 & 0.312 & 0.350 & 0.354 & 0.351 \\
MIT-BIH & 0.510 & 0.538 & 0.546 & 0.510 & 0.700 & 0.830 & 0.510 & 0.528 & 0.537 & 0.550 & 0.356 \\
Petrobras 3W & 0.780 & 0.728 & 0.676 & 0.780 & 0.930 & 0.990 & 0.780 & 0.744 & 0.703 & 0.639 & 0.639 \\
\bottomrule
\end{tabular}
\end{table}

\begin{table}[H]\centering
\tiny
\setlength{\tabcolsep}{2.2pt}
\renewcommand{\arraystretch}{0.85}
\caption{Full metric grid for \textbf{TiRex}, per dataset and macro-averaged, at each pollution level.}
\label{tab:a8-tirex-ed}
\begin{tabular}{lccccccccccc}
\toprule
 & P@1 & P@5 & P@10 & HR@1 & HR@5 & HR@10 & N@1 & N@5 & N@10 & N@20 & mF1 \\
\midrule
\multicolumn{12}{l}{\emph{$\rho{=}0\%$}} \\
CTSR & 0.670 & 0.518 & 0.456 & 0.670 & 0.840 & 0.880 & 0.670 & 0.548 & 0.494 & 0.565 & 0.543 \\
DAMADICS & 0.444 & 0.322 & 0.333 & 0.444 & 0.611 & 0.833 & 0.444 & 0.345 & 0.353 & 0.361 & 0.208 \\
Exathlon & 0.860 & 0.840 & 0.821 & 0.860 & 0.990 & 1.000 & 0.860 & 0.849 & 0.832 & 0.786 & 0.860 \\
HAI & 0.429 & 0.321 & 0.314 & 0.429 & 0.786 & 0.964 & 0.429 & 0.344 & 0.335 & 0.333 & 0.350 \\
MIT-BIH & 0.580 & 0.614 & 0.633 & 0.580 & 0.820 & 0.880 & 0.580 & 0.607 & 0.622 & 0.641 & 0.458 \\
Petrobras 3W & 0.740 & 0.582 & 0.521 & 0.740 & 0.890 & 0.920 & 0.740 & 0.616 & 0.562 & 0.508 & 0.586 \\
\bottomrule
\end{tabular}
\end{table}

\begin{table}[H]\centering
\tiny
\setlength{\tabcolsep}{2.2pt}
\renewcommand{\arraystretch}{0.85}
\caption{Full metric grid for \textbf{TiRex\,$+$\,NR}, per dataset and macro-averaged, at each pollution level.}
\label{tab:a8-tirex-delta}
\begin{tabular}{lccccccccccc}
\toprule
 & P@1 & P@5 & P@10 & HR@1 & HR@5 & HR@10 & N@1 & N@5 & N@10 & N@20 & mF1 \\
\midrule
\multicolumn{12}{l}{\emph{$\rho{=}0\%$}} \\
CTSR & 0.670 & 0.516 & 0.435 & 0.670 & 0.850 & 0.870 & 0.670 & 0.552 & 0.483 & 0.550 & 0.596 \\
DAMADICS & 0.444 & 0.322 & 0.344 & 0.444 & 0.778 & 0.889 & 0.444 & 0.349 & 0.373 & 0.407 & 0.215 \\
Exathlon & 0.880 & 0.828 & 0.809 & 0.880 & 0.990 & 0.990 & 0.880 & 0.838 & 0.822 & 0.758 & 0.820 \\
HAI & 0.321 & 0.329 & 0.339 & 0.321 & 0.679 & 0.786 & 0.321 & 0.328 & 0.335 & 0.333 & 0.374 \\
MIT-BIH & 0.650 & 0.662 & 0.656 & 0.650 & 0.800 & 0.850 & 0.650 & 0.661 & 0.657 & 0.663 & 0.367 \\
Petrobras 3W & 0.610 & 0.484 & 0.441 & 0.610 & 0.800 & 0.910 & 0.610 & 0.513 & 0.474 & 0.420 & 0.537 \\
\bottomrule
\end{tabular}
\end{table}

\begin{table}[H]\centering
\tiny
\setlength{\tabcolsep}{2.2pt}
\renewcommand{\arraystretch}{0.85}
\caption{Full metric grid for \textbf{CHARM\,$+$\,GPC}, per dataset and macro-averaged, at each pollution level.}
\label{tab:a8-cosine-gpc}
\begin{tabular}{lccccccccccc}
\toprule
 & P@1 & P@5 & P@10 & HR@1 & HR@5 & HR@10 & N@1 & N@5 & N@10 & N@20 & mF1 \\
\midrule
\multicolumn{12}{l}{\emph{$\rho{=}0\%$}} \\
CTSR & 0.630 & 0.630 & 0.630 & 0.630 & 0.630 & 0.630 & 0.630 & 0.630 & 0.630 & 0.711 & 0.572 \\
DAMADICS & 0.278 & 0.278 & 0.278 & 0.278 & 0.278 & 0.333 & 0.278 & 0.278 & 0.289 & 0.305 & 0.217 \\
Exathlon & 0.820 & 0.820 & 0.820 & 0.820 & 0.820 & 0.820 & 0.820 & 0.820 & 0.820 & 0.820 & 0.677 \\
HAI & 0.429 & 0.429 & 0.421 & 0.429 & 0.429 & 0.429 & 0.429 & 0.429 & 0.429 & 0.429 & 0.388 \\
MIT-BIH & 0.400 & 0.400 & 0.400 & 0.400 & 0.400 & 0.400 & 0.400 & 0.400 & 0.400 & 0.400 & 0.243 \\
Petrobras 3W & 0.670 & 0.670 & 0.670 & 0.670 & 0.670 & 0.670 & 0.670 & 0.670 & 0.670 & 0.670 & 0.562 \\
\bottomrule
\end{tabular}
\end{table}

\begin{table}[H]\centering
\tiny
\setlength{\tabcolsep}{2.2pt}
\renewcommand{\arraystretch}{0.85}
\caption{Full metric grid for \textbf{CHARM\,$+$\,MajVote}, per dataset and macro-averaged, at each pollution level.}
\label{tab:a8-cosine-majvote}
\begin{tabular}{lccccccccccc}
\toprule
 & P@1 & P@5 & P@10 & HR@1 & HR@5 & HR@10 & N@1 & N@5 & N@10 & N@20 & mF1 \\
\midrule
\multicolumn{12}{l}{\emph{$\rho{=}0\%$}} \\
CTSR & 0.570 & 0.570 & 0.570 & 0.570 & 0.570 & 0.570 & 0.570 & 0.570 & 0.570 & 0.631 & 0.516 \\
DAMADICS & 0.278 & 0.278 & 0.272 & 0.278 & 0.278 & 0.333 & 0.278 & 0.278 & 0.285 & 0.289 & 0.217 \\
Exathlon & 0.780 & 0.780 & 0.780 & 0.780 & 0.780 & 0.780 & 0.780 & 0.780 & 0.780 & 0.780 & 0.541 \\
HAI & 0.464 & 0.464 & 0.457 & 0.464 & 0.464 & 0.464 & 0.464 & 0.464 & 0.464 & 0.464 & 0.417 \\
MIT-BIH & 0.440 & 0.440 & 0.440 & 0.440 & 0.440 & 0.440 & 0.440 & 0.440 & 0.440 & 0.440 & 0.275 \\
Petrobras 3W & 0.660 & 0.660 & 0.660 & 0.660 & 0.660 & 0.660 & 0.660 & 0.660 & 0.660 & 0.660 & 0.556 \\
\bottomrule
\end{tabular}
\end{table}

\begin{table}[H]\centering
\tiny
\setlength{\tabcolsep}{2.2pt}
\renewcommand{\arraystretch}{0.85}
\caption{Full metric grid for \textbf{CHARM\,$+$\,QPurity}, per dataset and macro-averaged, at each pollution level.}
\label{tab:a8-cosine-qpurity}
\begin{tabular}{lccccccccccc}
\toprule
 & P@1 & P@5 & P@10 & HR@1 & HR@5 & HR@10 & N@1 & N@5 & N@10 & N@20 & mF1 \\
\midrule
\multicolumn{12}{l}{\emph{$\rho{=}0\%$}} \\
CTSR & 0.570 & 0.510 & 0.417 & 0.570 & 0.730 & 0.840 & 0.570 & 0.523 & 0.454 & 0.502 & 0.525 \\
DAMADICS & 0.222 & 0.289 & 0.278 & 0.222 & 0.444 & 0.444 & 0.222 & 0.273 & 0.274 & 0.282 & 0.167 \\
Exathlon & 0.810 & 0.786 & 0.781 & 0.810 & 0.810 & 0.810 & 0.810 & 0.792 & 0.786 & 0.781 & 0.608 \\
HAI & 0.429 & 0.407 & 0.414 & 0.429 & 0.429 & 0.536 & 0.429 & 0.414 & 0.417 & 0.404 & 0.299 \\
MIT-BIH & 0.440 & 0.434 & 0.434 & 0.440 & 0.440 & 0.490 & 0.440 & 0.435 & 0.435 & 0.435 & 0.275 \\
Petrobras 3W & 0.690 & 0.672 & 0.648 & 0.690 & 0.730 & 0.730 & 0.690 & 0.675 & 0.658 & 0.609 & 0.571 \\
\bottomrule
\end{tabular}
\end{table}

\begin{table}[H]\centering
\tiny
\setlength{\tabcolsep}{2.2pt}
\renewcommand{\arraystretch}{0.85}
\caption{Full metric grid for \textbf{CHARM\,$+$\,PRF}, per dataset and macro-averaged, at each pollution level.}
\label{tab:a8-cosine-prf}
\begin{tabular}{lccccccccccc}
\toprule
 & P@1 & P@5 & P@10 & HR@1 & HR@5 & HR@10 & N@1 & N@5 & N@10 & N@20 & mF1 \\
\midrule
\multicolumn{12}{l}{\emph{$\rho{=}0\%$}} \\
CTSR & 0.480 & 0.444 & 0.412 & 0.480 & 0.790 & 0.880 & 0.480 & 0.452 & 0.428 & 0.508 & 0.428 \\
DAMADICS & 0.389 & 0.311 & 0.289 & 0.389 & 0.611 & 0.778 & 0.389 & 0.327 & 0.315 & 0.360 & 0.274 \\
Exathlon & 0.740 & 0.728 & 0.703 & 0.740 & 0.860 & 0.920 & 0.740 & 0.731 & 0.713 & 0.681 & 0.513 \\
HAI & 0.429 & 0.379 & 0.364 & 0.429 & 0.571 & 0.786 & 0.429 & 0.395 & 0.383 & 0.367 & 0.331 \\
MIT-BIH & 0.440 & 0.450 & 0.454 & 0.440 & 0.480 & 0.520 & 0.440 & 0.447 & 0.451 & 0.464 & 0.275 \\
Petrobras 3W & 0.660 & 0.560 & 0.495 & 0.660 & 0.790 & 0.870 & 0.660 & 0.575 & 0.524 & 0.473 & 0.561 \\
\bottomrule
\end{tabular}
\end{table}

\begin{table}[H]\centering
\tiny
\setlength{\tabcolsep}{2.2pt}
\renewcommand{\arraystretch}{0.85}
\caption{Full metric grid for \textbf{CHARM\,$+$\,Purity}, per dataset and macro-averaged, at each pollution level.}
\label{tab:a8-cosine-purity}
\begin{tabular}{lccccccccccc}
\toprule
 & P@1 & P@5 & P@10 & HR@1 & HR@5 & HR@10 & N@1 & N@5 & N@10 & N@20 & mF1 \\
\midrule
\multicolumn{12}{l}{\emph{$\rho{=}0\%$}} \\
CTSR & 0.130 & 0.106 & 0.106 & 0.130 & 0.170 & 0.220 & 0.130 & 0.110 & 0.109 & 0.138 & 0.061 \\
DAMADICS & 0.056 & 0.156 & 0.189 & 0.056 & 0.278 & 0.333 & 0.056 & 0.133 & 0.164 & 0.191 & 0.075 \\
Exathlon & 0.750 & 0.728 & 0.719 & 0.750 & 0.850 & 0.910 & 0.750 & 0.735 & 0.727 & 0.688 & 0.512 \\
HAI & 0.250 & 0.271 & 0.246 & 0.250 & 0.429 & 0.571 & 0.250 & 0.267 & 0.252 & 0.254 & 0.143 \\
MIT-BIH & 0.380 & 0.446 & 0.469 & 0.380 & 0.580 & 0.660 & 0.380 & 0.431 & 0.453 & 0.476 & 0.275 \\
Petrobras 3W & 0.580 & 0.532 & 0.504 & 0.580 & 0.630 & 0.680 & 0.580 & 0.544 & 0.520 & 0.488 & 0.448 \\
\bottomrule
\end{tabular}
\end{table}

\begin{table}[H]\centering
\tiny
\setlength{\tabcolsep}{2.2pt}
\renewcommand{\arraystretch}{0.85}
\caption{Full metric grid for \textbf{CHARM\,$+$\,NR\,$+$\,GPC}, per dataset and macro-averaged, at each pollution level.}
\label{tab:a8-delta-gpc}
\begin{tabular}{lccccccccccc}
\toprule
 & P@1 & P@5 & P@10 & HR@1 & HR@5 & HR@10 & N@1 & N@5 & N@10 & N@20 & mF1 \\
\midrule
\multicolumn{12}{l}{\emph{$\rho{=}0\%$}} \\
CTSR & 0.580 & 0.580 & 0.580 & 0.580 & 0.580 & 0.580 & 0.580 & 0.580 & 0.580 & 0.661 & 0.499 \\
DAMADICS & 0.444 & 0.444 & 0.444 & 0.444 & 0.444 & 0.444 & 0.444 & 0.444 & 0.444 & 0.444 & 0.224 \\
Exathlon & 0.900 & 0.900 & 0.900 & 0.900 & 0.900 & 0.900 & 0.900 & 0.900 & 0.900 & 0.900 & 0.776 \\
HAI & 0.393 & 0.393 & 0.393 & 0.393 & 0.393 & 0.393 & 0.393 & 0.393 & 0.393 & 0.393 & 0.339 \\
MIT-BIH & 0.710 & 0.710 & 0.710 & 0.710 & 0.710 & 0.710 & 0.710 & 0.710 & 0.710 & 0.710 & 0.444 \\
Petrobras 3W & 0.650 & 0.650 & 0.650 & 0.650 & 0.650 & 0.650 & 0.650 & 0.650 & 0.650 & 0.650 & 0.544 \\
\bottomrule
\end{tabular}
\end{table}

\begin{table}[H]\centering
\tiny
\setlength{\tabcolsep}{2.2pt}
\renewcommand{\arraystretch}{0.85}
\caption{Full metric grid for \textbf{CHARM\,$+$\,NR\,$+$\,MajVote}, per dataset and macro-averaged, at each pollution level.}
\label{tab:a8-delta-majvote}
\begin{tabular}{lccccccccccc}
\toprule
 & P@1 & P@5 & P@10 & HR@1 & HR@5 & HR@10 & N@1 & N@5 & N@10 & N@20 & mF1 \\
\midrule
\multicolumn{12}{l}{\emph{$\rho{=}0\%$}} \\
CTSR & 0.580 & 0.580 & 0.580 & 0.580 & 0.580 & 0.580 & 0.580 & 0.580 & 0.580 & 0.632 & 0.520 \\
DAMADICS & 0.556 & 0.556 & 0.522 & 0.556 & 0.556 & 0.722 & 0.556 & 0.556 & 0.566 & 0.596 & 0.342 \\
Exathlon & 0.920 & 0.920 & 0.920 & 0.920 & 0.920 & 0.920 & 0.920 & 0.920 & 0.920 & 0.920 & 0.858 \\
HAI & 0.429 & 0.429 & 0.429 & 0.429 & 0.429 & 0.429 & 0.429 & 0.429 & 0.429 & 0.429 & 0.374 \\
MIT-BIH & 0.670 & 0.670 & 0.670 & 0.670 & 0.670 & 0.670 & 0.670 & 0.670 & 0.670 & 0.670 & 0.355 \\
Petrobras 3W & 0.610 & 0.610 & 0.610 & 0.610 & 0.610 & 0.610 & 0.610 & 0.610 & 0.610 & 0.610 & 0.505 \\
\bottomrule
\end{tabular}
\end{table}

\begin{table}[H]\centering
\tiny
\setlength{\tabcolsep}{2.2pt}
\renewcommand{\arraystretch}{0.85}
\caption{Full metric grid for \textbf{CHARM\,$+$\,NR\,$+$\,QPurity}, per dataset and macro-averaged, at each pollution level.}
\label{tab:a8-delta-qpurity}
\begin{tabular}{lccccccccccc}
\toprule
 & P@1 & P@5 & P@10 & HR@1 & HR@5 & HR@10 & N@1 & N@5 & N@10 & N@20 & mF1 \\
\midrule
\multicolumn{12}{l}{\emph{$\rho{=}0\%$}} \\
CTSR & 0.600 & 0.522 & 0.430 & 0.600 & 0.730 & 0.860 & 0.600 & 0.540 & 0.470 & 0.519 & 0.529 \\
DAMADICS & 0.389 & 0.444 & 0.472 & 0.389 & 0.778 & 0.944 & 0.389 & 0.429 & 0.461 & 0.481 & 0.205 \\
Exathlon & 0.890 & 0.854 & 0.838 & 0.890 & 0.930 & 0.960 & 0.890 & 0.861 & 0.847 & 0.829 & 0.624 \\
HAI & 0.464 & 0.436 & 0.414 & 0.464 & 0.607 & 0.750 & 0.464 & 0.443 & 0.427 & 0.391 & 0.394 \\
MIT-BIH & 0.570 & 0.616 & 0.653 & 0.570 & 0.720 & 0.830 & 0.570 & 0.609 & 0.637 & 0.674 & 0.348 \\
Petrobras 3W & 0.550 & 0.538 & 0.515 & 0.550 & 0.670 & 0.740 & 0.550 & 0.541 & 0.524 & 0.498 & 0.458 \\
\bottomrule
\end{tabular}
\end{table}

\begin{table}[H]\centering
\tiny
\setlength{\tabcolsep}{2.2pt}
\renewcommand{\arraystretch}{0.85}
\caption{Full metric grid for \textbf{CHARM\,$+$\,NR\,$+$\,PRF}, per dataset and macro-averaged, at each pollution level.}
\label{tab:a8-delta-prf}
\begin{tabular}{lccccccccccc}
\toprule
 & P@1 & P@5 & P@10 & HR@1 & HR@5 & HR@10 & N@1 & N@5 & N@10 & N@20 & mF1 \\
\midrule
\multicolumn{12}{l}{\emph{$\rho{=}0\%$}} \\
CTSR & 0.530 & 0.480 & 0.418 & 0.530 & 0.800 & 0.850 & 0.530 & 0.495 & 0.447 & 0.523 & 0.433 \\
DAMADICS & 0.556 & 0.511 & 0.461 & 0.556 & 0.833 & 1.000 & 0.556 & 0.518 & 0.510 & 0.530 & 0.354 \\
Exathlon & 0.870 & 0.846 & 0.838 & 0.870 & 0.970 & 1.000 & 0.870 & 0.854 & 0.845 & 0.806 & 0.749 \\
HAI & 0.393 & 0.371 & 0.329 & 0.393 & 0.643 & 0.786 & 0.393 & 0.372 & 0.343 & 0.343 & 0.446 \\
MIT-BIH & 0.700 & 0.674 & 0.691 & 0.700 & 0.760 & 0.830 & 0.700 & 0.679 & 0.689 & 0.704 & 0.360 \\
Petrobras 3W & 0.640 & 0.526 & 0.487 & 0.640 & 0.760 & 0.830 & 0.640 & 0.551 & 0.516 & 0.473 & 0.502 \\
\bottomrule
\end{tabular}
\end{table}

\begin{table}[H]\centering
\tiny
\setlength{\tabcolsep}{2.2pt}
\renewcommand{\arraystretch}{0.85}
\caption{Full metric grid for \textbf{CHARM\,$+$\,NR\,$+$\,Purity}, per dataset and macro-averaged, at each pollution level.}
\label{tab:a8-delta-purity}
\begin{tabular}{lccccccccccc}
\toprule
 & P@1 & P@5 & P@10 & HR@1 & HR@5 & HR@10 & N@1 & N@5 & N@10 & N@20 & mF1 \\
\midrule
\multicolumn{12}{l}{\emph{$\rho{=}0\%$}} \\
CTSR & 0.280 & 0.230 & 0.206 & 0.280 & 0.460 & 0.490 & 0.280 & 0.236 & 0.218 & 0.259 & 0.148 \\
DAMADICS & 0.167 & 0.344 & 0.372 & 0.167 & 0.611 & 0.611 & 0.167 & 0.314 & 0.346 & 0.344 & 0.222 \\
Exathlon & 0.890 & 0.796 & 0.755 & 0.890 & 0.910 & 0.940 & 0.890 & 0.814 & 0.779 & 0.733 & 0.576 \\
HAI & 0.286 & 0.329 & 0.318 & 0.286 & 0.571 & 0.750 & 0.286 & 0.320 & 0.315 & 0.313 & 0.211 \\
MIT-BIH & 0.410 & 0.578 & 0.638 & 0.410 & 0.790 & 0.870 & 0.410 & 0.547 & 0.600 & 0.649 & 0.291 \\
Petrobras 3W & 0.560 & 0.528 & 0.512 & 0.560 & 0.660 & 0.710 & 0.560 & 0.535 & 0.522 & 0.478 & 0.431 \\
\bottomrule
\end{tabular}
\end{table}

\begin{table}[H]\centering
\tiny
\setlength{\tabcolsep}{2.2pt}
\renewcommand{\arraystretch}{0.85}
\caption{Full metric grid for \textbf{DTW-I\,$+$\,GPC}, per dataset and macro-averaged, at each pollution level.}
\label{tab:a8-dtw-i-gpc}
\begin{tabular}{lccccccccccc}
\toprule
 & P@1 & P@5 & P@10 & HR@1 & HR@5 & HR@10 & N@1 & N@5 & N@10 & N@20 & mF1 \\
\midrule
\multicolumn{12}{l}{\emph{$\rho{=}0\%$}} \\
CTSR & 0.690 & 0.690 & 0.690 & 0.690 & 0.690 & 0.690 & 0.690 & 0.690 & 0.690 & 0.765 & 0.624 \\
DAMADICS & 0.611 & 0.611 & 0.594 & 0.611 & 0.611 & 0.667 & 0.611 & 0.611 & 0.622 & 0.638 & 0.416 \\
Exathlon & 0.330 & 0.330 & 0.330 & 0.330 & 0.330 & 0.330 & 0.330 & 0.330 & 0.330 & 0.330 & 0.322 \\
HAI & 0.214 & 0.214 & 0.200 & 0.214 & 0.214 & 0.214 & 0.214 & 0.214 & 0.214 & 0.214 & 0.230 \\
MIT-BIH & 0.580 & 0.580 & 0.580 & 0.580 & 0.580 & 0.580 & 0.580 & 0.580 & 0.580 & 0.580 & 0.544 \\
Petrobras 3W & 0.900 & 0.900 & 0.900 & 0.900 & 0.900 & 0.900 & 0.900 & 0.900 & 0.900 & 0.904 & 0.879 \\
\bottomrule
\end{tabular}
\end{table}

\begin{table}[H]\centering
\tiny
\setlength{\tabcolsep}{2.2pt}
\renewcommand{\arraystretch}{0.85}
\caption{Full metric grid for \textbf{DTW-I\,$+$\,MajVote}, per dataset and macro-averaged, at each pollution level.}
\label{tab:a8-dtw-i-majvote}
\begin{tabular}{lccccccccccc}
\toprule
 & P@1 & P@5 & P@10 & HR@1 & HR@5 & HR@10 & N@1 & N@5 & N@10 & N@20 & mF1 \\
\midrule
\multicolumn{12}{l}{\emph{$\rho{=}0\%$}} \\
CTSR & 0.610 & 0.610 & 0.610 & 0.610 & 0.610 & 0.610 & 0.610 & 0.610 & 0.610 & 0.662 & 0.539 \\
DAMADICS & 0.556 & 0.556 & 0.544 & 0.556 & 0.556 & 0.611 & 0.556 & 0.556 & 0.559 & 0.572 & 0.346 \\
Exathlon & 0.350 & 0.318 & 0.307 & 0.350 & 0.400 & 0.500 & 0.350 & 0.325 & 0.315 & 0.329 & 0.372 \\
HAI & 0.071 & 0.071 & 0.071 & 0.071 & 0.071 & 0.071 & 0.071 & 0.071 & 0.071 & 0.071 & 0.033 \\
MIT-BIH & 0.630 & 0.630 & 0.630 & 0.630 & 0.630 & 0.630 & 0.630 & 0.630 & 0.630 & 0.630 & 0.535 \\
Petrobras 3W & 0.930 & 0.874 & 0.850 & 0.930 & 0.940 & 0.960 & 0.930 & 0.886 & 0.865 & 0.823 & 0.873 \\
\bottomrule
\end{tabular}
\end{table}

\begin{table}[H]\centering
\tiny
\setlength{\tabcolsep}{2.2pt}
\renewcommand{\arraystretch}{0.85}
\caption{Full metric grid for \textbf{DTW-I\,$+$\,QPurity}, per dataset and macro-averaged, at each pollution level.}
\label{tab:a8-dtw-i-qpurity}
\begin{tabular}{lccccccccccc}
\toprule
 & P@1 & P@5 & P@10 & HR@1 & HR@5 & HR@10 & N@1 & N@5 & N@10 & N@20 & mF1 \\
\midrule
\multicolumn{12}{l}{\emph{$\rho{=}0\%$}} \\
CTSR & 0.580 & 0.534 & 0.435 & 0.580 & 0.710 & 0.810 & 0.580 & 0.547 & 0.475 & 0.507 & 0.532 \\
DAMADICS & 0.500 & 0.533 & 0.533 & 0.500 & 0.556 & 0.556 & 0.500 & 0.525 & 0.528 & 0.530 & 0.300 \\
Exathlon & 0.340 & 0.348 & 0.413 & 0.340 & 0.350 & 0.550 & 0.340 & 0.347 & 0.392 & 0.431 & 0.212 \\
HAI & 0.071 & 0.100 & 0.136 & 0.071 & 0.179 & 0.464 & 0.071 & 0.093 & 0.121 & 0.149 & 0.032 \\
MIT-BIH & 0.610 & 0.616 & 0.633 & 0.610 & 0.630 & 0.700 & 0.610 & 0.614 & 0.626 & 0.633 & 0.535 \\
Petrobras 3W & 0.840 & 0.826 & 0.819 & 0.840 & 0.850 & 0.890 & 0.840 & 0.827 & 0.822 & 0.782 & 0.731 \\
\bottomrule
\end{tabular}
\end{table}

\begin{table}[H]\centering
\tiny
\setlength{\tabcolsep}{2.2pt}
\renewcommand{\arraystretch}{0.85}
\caption{Full metric grid for \textbf{DTW-I\,$+$\,PRF}, per dataset and macro-averaged, at each pollution level.}
\label{tab:a8-dtw-i-prf}
\begin{tabular}{lccccccccccc}
\toprule
 & P@1 & P@5 & P@10 & HR@1 & HR@5 & HR@10 & N@1 & N@5 & N@10 & N@20 & mF1 \\
\midrule
\multicolumn{12}{l}{\emph{$\rho{=}0\%$}} \\
CTSR & 0.730 & 0.576 & 0.474 & 0.730 & 0.840 & 0.890 & 0.730 & 0.607 & 0.526 & 0.584 & 0.610 \\
DAMADICS & 0.667 & 0.622 & 0.583 & 0.667 & 0.722 & 0.833 & 0.667 & 0.641 & 0.635 & 0.649 & 0.506 \\
Exathlon & 0.350 & 0.318 & 0.307 & 0.350 & 0.400 & 0.500 & 0.350 & 0.325 & 0.315 & 0.329 & 0.372 \\
HAI & 0.107 & 0.121 & 0.154 & 0.107 & 0.464 & 0.786 & 0.107 & 0.115 & 0.141 & 0.171 & 0.032 \\
MIT-BIH & 0.500 & 0.576 & 0.574 & 0.500 & 0.850 & 0.910 & 0.500 & 0.562 & 0.566 & 0.564 & 0.458 \\
Petrobras 3W & 0.930 & 0.876 & 0.847 & 0.930 & 0.940 & 0.960 & 0.930 & 0.888 & 0.864 & 0.823 & 0.879 \\
\bottomrule
\end{tabular}
\end{table}

\begin{table}[H]\centering
\tiny
\setlength{\tabcolsep}{2.2pt}
\renewcommand{\arraystretch}{0.85}
\caption{Full metric grid for \textbf{DTW-I\,$+$\,Purity}, per dataset and macro-averaged, at each pollution level.}
\label{tab:a8-dtw-i-purity}
\begin{tabular}{lccccccccccc}
\toprule
 & P@1 & P@5 & P@10 & HR@1 & HR@5 & HR@10 & N@1 & N@5 & N@10 & N@20 & mF1 \\
\midrule
\multicolumn{12}{l}{\emph{$\rho{=}0\%$}} \\
CTSR & 0.140 & 0.100 & 0.105 & 0.140 & 0.200 & 0.310 & 0.140 & 0.110 & 0.110 & 0.139 & 0.058 \\
DAMADICS & 0.500 & 0.500 & 0.494 & 0.500 & 0.556 & 0.611 & 0.500 & 0.499 & 0.496 & 0.476 & 0.295 \\
Exathlon & 0.330 & 0.386 & 0.376 & 0.330 & 0.600 & 0.710 & 0.330 & 0.383 & 0.378 & 0.377 & 0.288 \\
HAI & 0.107 & 0.150 & 0.182 & 0.107 & 0.286 & 0.571 & 0.107 & 0.139 & 0.165 & 0.168 & 0.046 \\
MIT-BIH & 0.500 & 0.574 & 0.570 & 0.500 & 0.830 & 0.930 & 0.500 & 0.560 & 0.563 & 0.563 & 0.403 \\
Petrobras 3W & 0.740 & 0.538 & 0.508 & 0.740 & 0.780 & 0.810 & 0.740 & 0.581 & 0.546 & 0.481 & 0.462 \\
\bottomrule
\end{tabular}
\end{table}

\begin{table}[H]\centering
\tiny
\setlength{\tabcolsep}{2.2pt}
\renewcommand{\arraystretch}{0.85}
\caption{Full metric grid for \textbf{MantisV2\,$+$\,GPC}, per dataset and macro-averaged, at each pollution level.}
\label{tab:a8-mantisv2-ed-gpc}
\begin{tabular}{lccccccccccc}
\toprule
 & P@1 & P@5 & P@10 & HR@1 & HR@5 & HR@10 & N@1 & N@5 & N@10 & N@20 & mF1 \\
\midrule
\multicolumn{12}{l}{\emph{$\rho{=}0\%$}} \\
CTSR & 0.630 & 0.630 & 0.630 & 0.630 & 0.630 & 0.630 & 0.630 & 0.630 & 0.630 & 0.667 & 0.573 \\
DAMADICS & 0.611 & 0.611 & 0.628 & 0.611 & 0.611 & 0.833 & 0.611 & 0.611 & 0.655 & 0.719 & 0.359 \\
Exathlon & 0.940 & 0.940 & 0.940 & 0.940 & 0.940 & 0.940 & 0.940 & 0.940 & 0.940 & 0.940 & 0.964 \\
HAI & 0.429 & 0.429 & 0.421 & 0.429 & 0.429 & 0.429 & 0.429 & 0.429 & 0.429 & 0.429 & 0.494 \\
MIT-BIH & 0.340 & 0.340 & 0.340 & 0.340 & 0.340 & 0.340 & 0.340 & 0.340 & 0.340 & 0.340 & 0.295 \\
Petrobras 3W & 0.860 & 0.860 & 0.860 & 0.860 & 0.860 & 0.860 & 0.860 & 0.860 & 0.860 & 0.860 & 0.756 \\
\bottomrule
\end{tabular}
\end{table}

\begin{table}[H]\centering
\tiny
\setlength{\tabcolsep}{2.2pt}
\renewcommand{\arraystretch}{0.85}
\caption{Full metric grid for \textbf{MantisV2\,$+$\,MajVote}, per dataset and macro-averaged, at each pollution level.}
\label{tab:a8-mantisv2-ed-majvote}
\begin{tabular}{lccccccccccc}
\toprule
 & P@1 & P@5 & P@10 & HR@1 & HR@5 & HR@10 & N@1 & N@5 & N@10 & N@20 & mF1 \\
\midrule
\multicolumn{12}{l}{\emph{$\rho{=}0\%$}} \\
CTSR & 0.550 & 0.550 & 0.550 & 0.550 & 0.550 & 0.550 & 0.550 & 0.550 & 0.550 & 0.599 & 0.482 \\
DAMADICS & 0.667 & 0.667 & 0.667 & 0.667 & 0.667 & 0.667 & 0.667 & 0.667 & 0.667 & 0.667 & 0.389 \\
Exathlon & 0.920 & 0.920 & 0.920 & 0.920 & 0.920 & 0.920 & 0.920 & 0.920 & 0.920 & 0.920 & 0.792 \\
HAI & 0.393 & 0.393 & 0.386 & 0.393 & 0.393 & 0.393 & 0.393 & 0.393 & 0.393 & 0.393 & 0.465 \\
MIT-BIH & 0.340 & 0.340 & 0.340 & 0.340 & 0.340 & 0.340 & 0.340 & 0.340 & 0.340 & 0.340 & 0.227 \\
Petrobras 3W & 0.850 & 0.850 & 0.850 & 0.850 & 0.850 & 0.850 & 0.850 & 0.850 & 0.850 & 0.850 & 0.743 \\
\bottomrule
\end{tabular}
\end{table}

\begin{table}[H]\centering
\tiny
\setlength{\tabcolsep}{2.2pt}
\renewcommand{\arraystretch}{0.85}
\caption{Full metric grid for \textbf{MantisV2\,$+$\,QPurity}, per dataset and macro-averaged, at each pollution level.}
\label{tab:a8-mantisv2-ed-qpurity}
\begin{tabular}{lccccccccccc}
\toprule
 & P@1 & P@5 & P@10 & HR@1 & HR@5 & HR@10 & N@1 & N@5 & N@10 & N@20 & mF1 \\
\midrule
\multicolumn{12}{l}{\emph{$\rho{=}0\%$}} \\
CTSR & 0.580 & 0.520 & 0.438 & 0.580 & 0.730 & 0.850 & 0.580 & 0.531 & 0.470 & 0.515 & 0.538 \\
DAMADICS & 0.667 & 0.700 & 0.656 & 0.667 & 0.944 & 0.944 & 0.667 & 0.688 & 0.680 & 0.704 & 0.750 \\
Exathlon & 0.930 & 0.926 & 0.922 & 0.930 & 0.930 & 0.930 & 0.930 & 0.927 & 0.924 & 0.922 & 0.800 \\
HAI & 0.429 & 0.371 & 0.361 & 0.429 & 0.500 & 0.643 & 0.429 & 0.384 & 0.376 & 0.380 & 0.436 \\
MIT-BIH & 0.360 & 0.340 & 0.339 & 0.360 & 0.370 & 0.390 & 0.360 & 0.343 & 0.342 & 0.348 & 0.227 \\
Petrobras 3W & 0.860 & 0.832 & 0.797 & 0.860 & 0.890 & 0.890 & 0.860 & 0.838 & 0.812 & 0.764 & 0.708 \\
\bottomrule
\end{tabular}
\end{table}

\begin{table}[H]\centering
\tiny
\setlength{\tabcolsep}{2.2pt}
\renewcommand{\arraystretch}{0.85}
\caption{Full metric grid for \textbf{MantisV2\,$+$\,PRF}, per dataset and macro-averaged, at each pollution level.}
\label{tab:a8-mantisv2-ed-prf}
\begin{tabular}{lccccccccccc}
\toprule
 & P@1 & P@5 & P@10 & HR@1 & HR@5 & HR@10 & N@1 & N@5 & N@10 & N@20 & mF1 \\
\midrule
\multicolumn{12}{l}{\emph{$\rho{=}0\%$}} \\
CTSR & 0.540 & 0.480 & 0.405 & 0.540 & 0.750 & 0.830 & 0.540 & 0.495 & 0.439 & 0.504 & 0.528 \\
DAMADICS & 0.556 & 0.678 & 0.600 & 0.556 & 0.944 & 0.944 & 0.556 & 0.648 & 0.624 & 0.624 & 0.776 \\
Exathlon & 0.940 & 0.922 & 0.913 & 0.940 & 0.960 & 0.980 & 0.940 & 0.926 & 0.918 & 0.904 & 0.958 \\
HAI & 0.500 & 0.471 & 0.393 & 0.500 & 0.643 & 0.857 & 0.500 & 0.481 & 0.429 & 0.395 & 0.569 \\
MIT-BIH & 0.330 & 0.328 & 0.333 & 0.330 & 0.350 & 0.370 & 0.330 & 0.329 & 0.332 & 0.336 & 0.232 \\
Petrobras 3W & 0.850 & 0.808 & 0.758 & 0.850 & 0.940 & 0.980 & 0.850 & 0.818 & 0.780 & 0.707 & 0.738 \\
\bottomrule
\end{tabular}
\end{table}

\begin{table}[H]\centering
\tiny
\setlength{\tabcolsep}{2.2pt}
\renewcommand{\arraystretch}{0.85}
\caption{Full metric grid for \textbf{MantisV2\,$+$\,Purity}, per dataset and macro-averaged, at each pollution level.}
\label{tab:a8-mantisv2-ed-purity}
\begin{tabular}{lccccccccccc}
\toprule
 & P@1 & P@5 & P@10 & HR@1 & HR@5 & HR@10 & N@1 & N@5 & N@10 & N@20 & mF1 \\
\midrule
\multicolumn{12}{l}{\emph{$\rho{=}0\%$}} \\
CTSR & 0.270 & 0.248 & 0.223 & 0.270 & 0.380 & 0.420 & 0.270 & 0.250 & 0.232 & 0.265 & 0.168 \\
DAMADICS & 0.611 & 0.600 & 0.583 & 0.611 & 0.667 & 0.667 & 0.611 & 0.603 & 0.592 & 0.564 & 0.360 \\
Exathlon & 0.940 & 0.902 & 0.888 & 0.940 & 0.940 & 0.940 & 0.940 & 0.910 & 0.897 & 0.884 & 0.609 \\
HAI & 0.464 & 0.500 & 0.454 & 0.464 & 0.643 & 0.714 & 0.464 & 0.498 & 0.465 & 0.438 & 0.484 \\
MIT-BIH & 0.350 & 0.346 & 0.348 & 0.350 & 0.410 & 0.450 & 0.350 & 0.348 & 0.349 & 0.344 & 0.223 \\
Petrobras 3W & 0.740 & 0.694 & 0.657 & 0.740 & 0.820 & 0.850 & 0.740 & 0.706 & 0.676 & 0.627 & 0.612 \\
\bottomrule
\end{tabular}
\end{table}

\begin{table}[H]\centering
\tiny
\setlength{\tabcolsep}{2.2pt}
\renewcommand{\arraystretch}{0.85}
\caption{Full metric grid for \textbf{MantisV2\,$+$\,NR\,$+$\,GPC}, per dataset and macro-averaged, at each pollution level.}
\label{tab:a8-mantisv2-delta-gpc}
\begin{tabular}{lccccccccccc}
\toprule
 & P@1 & P@5 & P@10 & HR@1 & HR@5 & HR@10 & N@1 & N@5 & N@10 & N@20 & mF1 \\
\midrule
\multicolumn{12}{l}{\emph{$\rho{=}0\%$}} \\
CTSR & 0.600 & 0.600 & 0.602 & 0.600 & 0.600 & 0.610 & 0.600 & 0.600 & 0.601 & 0.660 & 0.546 \\
DAMADICS & 0.556 & 0.556 & 0.506 & 0.556 & 0.556 & 0.556 & 0.556 & 0.556 & 0.556 & 0.556 & 0.342 \\
Exathlon & 0.950 & 0.950 & 0.950 & 0.950 & 0.950 & 0.950 & 0.950 & 0.950 & 0.950 & 0.950 & 0.914 \\
HAI & 0.536 & 0.536 & 0.529 & 0.536 & 0.536 & 0.536 & 0.536 & 0.536 & 0.536 & 0.536 & 0.616 \\
MIT-BIH & 0.550 & 0.550 & 0.550 & 0.550 & 0.550 & 0.550 & 0.550 & 0.550 & 0.550 & 0.550 & 0.428 \\
Petrobras 3W & 0.820 & 0.820 & 0.820 & 0.820 & 0.820 & 0.820 & 0.820 & 0.820 & 0.820 & 0.820 & 0.712 \\
\bottomrule
\end{tabular}
\end{table}

\begin{table}[H]\centering
\tiny
\setlength{\tabcolsep}{2.2pt}
\renewcommand{\arraystretch}{0.85}
\caption{Full metric grid for \textbf{MantisV2\,$+$\,NR\,$+$\,MajVote}, per dataset and macro-averaged, at each pollution level.}
\label{tab:a8-mantisv2-delta-majvote}
\begin{tabular}{lccccccccccc}
\toprule
 & P@1 & P@5 & P@10 & HR@1 & HR@5 & HR@10 & N@1 & N@5 & N@10 & N@20 & mF1 \\
\midrule
\multicolumn{12}{l}{\emph{$\rho{=}0\%$}} \\
CTSR & 0.590 & 0.590 & 0.590 & 0.590 & 0.590 & 0.590 & 0.590 & 0.590 & 0.590 & 0.632 & 0.526 \\
DAMADICS & 0.500 & 0.500 & 0.483 & 0.500 & 0.500 & 0.500 & 0.500 & 0.500 & 0.500 & 0.508 & 0.257 \\
Exathlon & 0.940 & 0.940 & 0.940 & 0.940 & 0.940 & 0.940 & 0.940 & 0.940 & 0.940 & 0.940 & 0.789 \\
HAI & 0.357 & 0.357 & 0.350 & 0.357 & 0.357 & 0.357 & 0.357 & 0.357 & 0.357 & 0.357 & 0.342 \\
MIT-BIH & 0.550 & 0.550 & 0.550 & 0.550 & 0.550 & 0.550 & 0.550 & 0.550 & 0.550 & 0.550 & 0.361 \\
Petrobras 3W & 0.750 & 0.750 & 0.750 & 0.750 & 0.750 & 0.750 & 0.750 & 0.750 & 0.750 & 0.750 & 0.568 \\
\bottomrule
\end{tabular}
\end{table}

\begin{table}[H]\centering
\tiny
\setlength{\tabcolsep}{2.2pt}
\renewcommand{\arraystretch}{0.85}
\caption{Full metric grid for \textbf{MantisV2\,$+$\,NR\,$+$\,QPurity}, per dataset and macro-averaged, at each pollution level.}
\label{tab:a8-mantisv2-delta-qpurity}
\begin{tabular}{lccccccccccc}
\toprule
 & P@1 & P@5 & P@10 & HR@1 & HR@5 & HR@10 & N@1 & N@5 & N@10 & N@20 & mF1 \\
\midrule
\multicolumn{12}{l}{\emph{$\rho{=}0\%$}} \\
CTSR & 0.570 & 0.508 & 0.440 & 0.570 & 0.680 & 0.820 & 0.570 & 0.521 & 0.470 & 0.512 & 0.477 \\
DAMADICS & 0.722 & 0.600 & 0.522 & 0.722 & 0.778 & 0.889 & 0.722 & 0.633 & 0.591 & 0.577 & 0.622 \\
Exathlon & 0.920 & 0.922 & 0.915 & 0.920 & 0.970 & 0.970 & 0.920 & 0.923 & 0.918 & 0.906 & 0.840 \\
HAI & 0.357 & 0.379 & 0.368 & 0.357 & 0.429 & 0.464 & 0.357 & 0.377 & 0.372 & 0.371 & 0.405 \\
MIT-BIH & 0.560 & 0.548 & 0.552 & 0.560 & 0.650 & 0.730 & 0.560 & 0.551 & 0.552 & 0.559 & 0.352 \\
Petrobras 3W & 0.770 & 0.734 & 0.696 & 0.770 & 0.840 & 0.890 & 0.770 & 0.743 & 0.714 & 0.667 & 0.559 \\
\bottomrule
\end{tabular}
\end{table}

\begin{table}[H]\centering
\tiny
\setlength{\tabcolsep}{2.2pt}
\renewcommand{\arraystretch}{0.85}
\caption{Full metric grid for \textbf{MantisV2\,$+$\,NR\,$+$\,PRF}, per dataset and macro-averaged, at each pollution level.}
\label{tab:a8-mantisv2-delta-prf}
\begin{tabular}{lccccccccccc}
\toprule
 & P@1 & P@5 & P@10 & HR@1 & HR@5 & HR@10 & N@1 & N@5 & N@10 & N@20 & mF1 \\
\midrule
\multicolumn{12}{l}{\emph{$\rho{=}0\%$}} \\
CTSR & 0.460 & 0.434 & 0.379 & 0.460 & 0.720 & 0.830 & 0.460 & 0.441 & 0.401 & 0.474 & 0.374 \\
DAMADICS & 0.500 & 0.467 & 0.428 & 0.500 & 0.667 & 0.778 & 0.500 & 0.479 & 0.464 & 0.444 & 0.426 \\
Exathlon & 0.870 & 0.886 & 0.877 & 0.870 & 0.930 & 0.940 & 0.870 & 0.883 & 0.878 & 0.876 & 0.664 \\
HAI & 0.321 & 0.343 & 0.354 & 0.321 & 0.464 & 0.643 & 0.321 & 0.334 & 0.349 & 0.326 & 0.477 \\
MIT-BIH & 0.550 & 0.534 & 0.544 & 0.550 & 0.560 & 0.600 & 0.550 & 0.537 & 0.543 & 0.548 & 0.358 \\
Petrobras 3W & 0.730 & 0.692 & 0.661 & 0.730 & 0.820 & 0.870 & 0.730 & 0.701 & 0.677 & 0.628 & 0.541 \\
\bottomrule
\end{tabular}
\end{table}

\begin{table}[H]\centering
\tiny
\setlength{\tabcolsep}{2.2pt}
\renewcommand{\arraystretch}{0.85}
\caption{Full metric grid for \textbf{MantisV2\,$+$\,NR\,$+$\,Purity}, per dataset and macro-averaged, at each pollution level.}
\label{tab:a8-mantisv2-delta-purity}
\begin{tabular}{lccccccccccc}
\toprule
 & P@1 & P@5 & P@10 & HR@1 & HR@5 & HR@10 & N@1 & N@5 & N@10 & N@20 & mF1 \\
\midrule
\multicolumn{12}{l}{\emph{$\rho{=}0\%$}} \\
CTSR & 0.260 & 0.240 & 0.204 & 0.260 & 0.360 & 0.390 & 0.260 & 0.242 & 0.216 & 0.248 & 0.157 \\
DAMADICS & 0.556 & 0.522 & 0.494 & 0.556 & 0.667 & 0.722 & 0.556 & 0.532 & 0.510 & 0.450 & 0.330 \\
Exathlon & 0.950 & 0.936 & 0.927 & 0.950 & 0.960 & 0.970 & 0.950 & 0.938 & 0.932 & 0.920 & 0.730 \\
HAI & 0.286 & 0.307 & 0.304 & 0.286 & 0.536 & 0.750 & 0.286 & 0.303 & 0.302 & 0.327 & 0.233 \\
MIT-BIH & 0.540 & 0.554 & 0.567 & 0.540 & 0.710 & 0.820 & 0.540 & 0.553 & 0.562 & 0.569 & 0.372 \\
Petrobras 3W & 0.740 & 0.696 & 0.636 & 0.740 & 0.810 & 0.860 & 0.740 & 0.707 & 0.661 & 0.609 & 0.580 \\
\bottomrule
\end{tabular}
\end{table}

\begin{table}[H]\centering
\tiny
\setlength{\tabcolsep}{2.2pt}
\renewcommand{\arraystretch}{0.85}
\caption{Full metric grid for \textbf{TiRex\,$+$\,GPC}, per dataset and macro-averaged, at each pollution level.}
\label{tab:a8-tirex-ed-gpc}
\begin{tabular}{lccccccccccc}
\toprule
 & P@1 & P@5 & P@10 & HR@1 & HR@5 & HR@10 & N@1 & N@5 & N@10 & N@20 & mF1 \\
\midrule
\multicolumn{12}{l}{\emph{$\rho{=}0\%$}} \\
CTSR & 0.660 & 0.662 & 0.661 & 0.660 & 0.670 & 0.670 & 0.660 & 0.662 & 0.661 & 0.706 & 0.592 \\
DAMADICS & 0.389 & 0.389 & 0.372 & 0.389 & 0.389 & 0.389 & 0.389 & 0.389 & 0.389 & 0.389 & 0.274 \\
Exathlon & 0.890 & 0.890 & 0.890 & 0.890 & 0.890 & 0.890 & 0.890 & 0.890 & 0.890 & 0.890 & 0.886 \\
HAI & 0.250 & 0.250 & 0.250 & 0.250 & 0.250 & 0.250 & 0.250 & 0.250 & 0.250 & 0.250 & 0.168 \\
MIT-BIH & 0.570 & 0.570 & 0.570 & 0.570 & 0.570 & 0.570 & 0.570 & 0.570 & 0.570 & 0.570 & 0.321 \\
Petrobras 3W & 0.610 & 0.614 & 0.618 & 0.610 & 0.640 & 0.650 & 0.610 & 0.614 & 0.617 & 0.617 & 0.516 \\
\bottomrule
\end{tabular}
\end{table}

\begin{table}[H]\centering
\tiny
\setlength{\tabcolsep}{2.2pt}
\renewcommand{\arraystretch}{0.85}
\caption{Full metric grid for \textbf{TiRex\,$+$\,MajVote}, per dataset and macro-averaged, at each pollution level.}
\label{tab:a8-tirex-ed-majvote}
\begin{tabular}{lccccccccccc}
\toprule
 & P@1 & P@5 & P@10 & HR@1 & HR@5 & HR@10 & N@1 & N@5 & N@10 & N@20 & mF1 \\
\midrule
\multicolumn{12}{l}{\emph{$\rho{=}0\%$}} \\
CTSR & 0.600 & 0.600 & 0.600 & 0.600 & 0.600 & 0.600 & 0.600 & 0.600 & 0.600 & 0.655 & 0.534 \\
DAMADICS & 0.278 & 0.278 & 0.278 & 0.278 & 0.278 & 0.278 & 0.278 & 0.278 & 0.278 & 0.278 & 0.109 \\
Exathlon & 0.870 & 0.870 & 0.870 & 0.870 & 0.870 & 0.870 & 0.870 & 0.870 & 0.870 & 0.870 & 0.818 \\
HAI & 0.250 & 0.250 & 0.250 & 0.250 & 0.250 & 0.250 & 0.250 & 0.250 & 0.250 & 0.250 & 0.161 \\
MIT-BIH & 0.630 & 0.630 & 0.630 & 0.630 & 0.630 & 0.630 & 0.630 & 0.630 & 0.630 & 0.630 & 0.334 \\
Petrobras 3W & 0.630 & 0.630 & 0.630 & 0.630 & 0.630 & 0.630 & 0.630 & 0.630 & 0.630 & 0.631 & 0.499 \\
\bottomrule
\end{tabular}
\end{table}

\begin{table}[H]\centering
\tiny
\setlength{\tabcolsep}{2.2pt}
\renewcommand{\arraystretch}{0.85}
\caption{Full metric grid for \textbf{TiRex\,$+$\,QPurity}, per dataset and macro-averaged, at each pollution level.}
\label{tab:a8-tirex-ed-qpurity}
\begin{tabular}{lccccccccccc}
\toprule
 & P@1 & P@5 & P@10 & HR@1 & HR@5 & HR@10 & N@1 & N@5 & N@10 & N@20 & mF1 \\
\midrule
\multicolumn{12}{l}{\emph{$\rho{=}0\%$}} \\
CTSR & 0.590 & 0.514 & 0.444 & 0.590 & 0.680 & 0.820 & 0.590 & 0.531 & 0.477 & 0.526 & 0.483 \\
DAMADICS & 0.333 & 0.300 & 0.256 & 0.333 & 0.444 & 0.556 & 0.333 & 0.308 & 0.283 & 0.299 & 0.114 \\
Exathlon & 0.820 & 0.828 & 0.830 & 0.820 & 0.860 & 0.890 & 0.820 & 0.827 & 0.828 & 0.827 & 0.580 \\
HAI & 0.179 & 0.257 & 0.243 & 0.179 & 0.464 & 0.571 & 0.179 & 0.246 & 0.242 & 0.263 & 0.097 \\
MIT-BIH & 0.620 & 0.606 & 0.605 & 0.620 & 0.650 & 0.680 & 0.620 & 0.608 & 0.607 & 0.608 & 0.331 \\
Petrobras 3W & 0.670 & 0.614 & 0.587 & 0.670 & 0.700 & 0.740 & 0.670 & 0.627 & 0.604 & 0.566 & 0.529 \\
\bottomrule
\end{tabular}
\end{table}

\begin{table}[H]\centering
\tiny
\setlength{\tabcolsep}{2.2pt}
\renewcommand{\arraystretch}{0.85}
\caption{Full metric grid for \textbf{TiRex\,$+$\,PRF}, per dataset and macro-averaged, at each pollution level.}
\label{tab:a8-tirex-ed-prf}
\begin{tabular}{lccccccccccc}
\toprule
 & P@1 & P@5 & P@10 & HR@1 & HR@5 & HR@10 & N@1 & N@5 & N@10 & N@20 & mF1 \\
\midrule
\multicolumn{12}{l}{\emph{$\rho{=}0\%$}} \\
CTSR & 0.600 & 0.508 & 0.445 & 0.600 & 0.820 & 0.870 & 0.600 & 0.526 & 0.477 & 0.557 & 0.490 \\
DAMADICS & 0.278 & 0.289 & 0.328 & 0.278 & 0.500 & 0.833 & 0.278 & 0.285 & 0.322 & 0.329 & 0.109 \\
Exathlon & 0.860 & 0.818 & 0.814 & 0.860 & 0.920 & 0.990 & 0.860 & 0.826 & 0.821 & 0.783 & 0.780 \\
HAI & 0.286 & 0.314 & 0.286 & 0.286 & 0.750 & 0.857 & 0.286 & 0.325 & 0.302 & 0.302 & 0.166 \\
MIT-BIH & 0.620 & 0.602 & 0.591 & 0.620 & 0.670 & 0.670 & 0.620 & 0.607 & 0.597 & 0.598 & 0.333 \\
Petrobras 3W & 0.720 & 0.568 & 0.512 & 0.720 & 0.850 & 0.890 & 0.720 & 0.600 & 0.550 & 0.494 & 0.543 \\
\bottomrule
\end{tabular}
\end{table}

\begin{table}[H]\centering
\tiny
\setlength{\tabcolsep}{2.2pt}
\renewcommand{\arraystretch}{0.85}
\caption{Full metric grid for \textbf{TiRex\,$+$\,Purity}, per dataset and macro-averaged, at each pollution level.}
\label{tab:a8-tirex-ed-purity}
\begin{tabular}{lccccccccccc}
\toprule
 & P@1 & P@5 & P@10 & HR@1 & HR@5 & HR@10 & N@1 & N@5 & N@10 & N@20 & mF1 \\
\midrule
\multicolumn{12}{l}{\emph{$\rho{=}0\%$}} \\
CTSR & 0.190 & 0.188 & 0.171 & 0.190 & 0.310 & 0.370 & 0.190 & 0.190 & 0.178 & 0.209 & 0.103 \\
DAMADICS & 0.222 & 0.267 & 0.250 & 0.222 & 0.333 & 0.333 & 0.222 & 0.265 & 0.254 & 0.269 & 0.151 \\
Exathlon & 0.830 & 0.788 & 0.764 & 0.830 & 0.900 & 0.940 & 0.830 & 0.795 & 0.776 & 0.744 & 0.548 \\
HAI & 0.214 & 0.214 & 0.236 & 0.214 & 0.321 & 0.500 & 0.214 & 0.211 & 0.226 & 0.228 & 0.051 \\
MIT-BIH & 0.590 & 0.602 & 0.606 & 0.590 & 0.710 & 0.810 & 0.590 & 0.600 & 0.603 & 0.624 & 0.334 \\
Petrobras 3W & 0.480 & 0.446 & 0.432 & 0.480 & 0.560 & 0.630 & 0.480 & 0.454 & 0.441 & 0.430 & 0.255 \\
\bottomrule
\end{tabular}
\end{table}

\begin{table}[H]\centering
\tiny
\setlength{\tabcolsep}{2.2pt}
\renewcommand{\arraystretch}{0.85}
\caption{Full metric grid for \textbf{TiRex\,$+$\,NR\,$+$\,GPC}, per dataset and macro-averaged, at each pollution level.}
\label{tab:a8-tirex-delta-gpc}
\begin{tabular}{lccccccccccc}
\toprule
 & P@1 & P@5 & P@10 & HR@1 & HR@5 & HR@10 & N@1 & N@5 & N@10 & N@20 & mF1 \\
\midrule
\multicolumn{12}{l}{\emph{$\rho{=}0\%$}} \\
CTSR & 0.610 & 0.610 & 0.610 & 0.610 & 0.640 & 0.650 & 0.610 & 0.610 & 0.610 & 0.668 & 0.536 \\
DAMADICS & 0.389 & 0.389 & 0.372 & 0.389 & 0.389 & 0.389 & 0.389 & 0.389 & 0.389 & 0.389 & 0.274 \\
Exathlon & 0.940 & 0.940 & 0.940 & 0.940 & 0.940 & 0.940 & 0.940 & 0.940 & 0.940 & 0.940 & 0.874 \\
HAI & 0.214 & 0.214 & 0.214 & 0.214 & 0.214 & 0.214 & 0.214 & 0.214 & 0.214 & 0.214 & 0.159 \\
MIT-BIH & 0.720 & 0.710 & 0.709 & 0.720 & 0.760 & 0.780 & 0.720 & 0.712 & 0.710 & 0.708 & 0.461 \\
Petrobras 3W & 0.570 & 0.588 & 0.598 & 0.570 & 0.700 & 0.700 & 0.570 & 0.587 & 0.594 & 0.598 & 0.487 \\
\bottomrule
\end{tabular}
\end{table}

\begin{table}[H]\centering
\tiny
\setlength{\tabcolsep}{2.2pt}
\renewcommand{\arraystretch}{0.85}
\caption{Full metric grid for \textbf{TiRex\,$+$\,NR\,$+$\,MajVote}, per dataset and macro-averaged, at each pollution level.}
\label{tab:a8-tirex-delta-majvote}
\begin{tabular}{lccccccccccc}
\toprule
 & P@1 & P@5 & P@10 & HR@1 & HR@5 & HR@10 & N@1 & N@5 & N@10 & N@20 & mF1 \\
\midrule
\multicolumn{12}{l}{\emph{$\rho{=}0\%$}} \\
CTSR & 0.600 & 0.600 & 0.600 & 0.600 & 0.600 & 0.600 & 0.600 & 0.600 & 0.600 & 0.643 & 0.539 \\
DAMADICS & 0.389 & 0.389 & 0.389 & 0.389 & 0.389 & 0.500 & 0.389 & 0.389 & 0.400 & 0.428 & 0.232 \\
Exathlon & 0.930 & 0.930 & 0.930 & 0.930 & 0.930 & 0.930 & 0.930 & 0.930 & 0.930 & 0.930 & 0.836 \\
HAI & 0.393 & 0.393 & 0.393 & 0.393 & 0.393 & 0.393 & 0.393 & 0.393 & 0.393 & 0.393 & 0.392 \\
MIT-BIH & 0.680 & 0.680 & 0.680 & 0.680 & 0.680 & 0.680 & 0.680 & 0.680 & 0.680 & 0.680 & 0.357 \\
Petrobras 3W & 0.570 & 0.570 & 0.570 & 0.570 & 0.570 & 0.570 & 0.570 & 0.570 & 0.570 & 0.570 & 0.478 \\
\bottomrule
\end{tabular}
\end{table}

\begin{table}[H]\centering
\tiny
\setlength{\tabcolsep}{2.2pt}
\renewcommand{\arraystretch}{0.85}
\caption{Full metric grid for \textbf{TiRex\,$+$\,NR\,$+$\,QPurity}, per dataset and macro-averaged, at each pollution level.}
\label{tab:a8-tirex-delta-qpurity}
\begin{tabular}{lccccccccccc}
\toprule
 & P@1 & P@5 & P@10 & HR@1 & HR@5 & HR@10 & N@1 & N@5 & N@10 & N@20 & mF1 \\
\midrule
\multicolumn{12}{l}{\emph{$\rho{=}0\%$}} \\
CTSR & 0.620 & 0.536 & 0.451 & 0.620 & 0.750 & 0.820 & 0.620 & 0.556 & 0.491 & 0.532 & 0.554 \\
DAMADICS & 0.444 & 0.367 & 0.339 & 0.444 & 0.667 & 0.833 & 0.444 & 0.388 & 0.373 & 0.384 & 0.296 \\
Exathlon & 0.850 & 0.856 & 0.848 & 0.850 & 0.920 & 0.940 & 0.850 & 0.856 & 0.850 & 0.828 & 0.584 \\
HAI & 0.393 & 0.371 & 0.368 & 0.393 & 0.536 & 0.643 & 0.393 & 0.371 & 0.370 & 0.345 & 0.390 \\
MIT-BIH & 0.680 & 0.674 & 0.672 & 0.680 & 0.700 & 0.720 & 0.680 & 0.676 & 0.674 & 0.666 & 0.357 \\
Petrobras 3W & 0.540 & 0.510 & 0.486 & 0.540 & 0.620 & 0.690 & 0.540 & 0.516 & 0.498 & 0.470 & 0.431 \\
\bottomrule
\end{tabular}
\end{table}

\begin{table}[H]\centering
\tiny
\setlength{\tabcolsep}{2.2pt}
\renewcommand{\arraystretch}{0.85}
\caption{Full metric grid for \textbf{TiRex\,$+$\,NR\,$+$\,PRF}, per dataset and macro-averaged, at each pollution level.}
\label{tab:a8-tirex-delta-prf}
\begin{tabular}{lccccccccccc}
\toprule
 & P@1 & P@5 & P@10 & HR@1 & HR@5 & HR@10 & N@1 & N@5 & N@10 & N@20 & mF1 \\
\midrule
\multicolumn{12}{l}{\emph{$\rho{=}0\%$}} \\
CTSR & 0.470 & 0.438 & 0.415 & 0.470 & 0.750 & 0.860 & 0.470 & 0.442 & 0.426 & 0.500 & 0.344 \\
DAMADICS & 0.444 & 0.367 & 0.278 & 0.444 & 0.667 & 0.667 & 0.444 & 0.382 & 0.324 & 0.369 & 0.214 \\
Exathlon & 0.850 & 0.840 & 0.838 & 0.850 & 0.950 & 0.960 & 0.850 & 0.842 & 0.840 & 0.812 & 0.682 \\
HAI & 0.321 & 0.329 & 0.311 & 0.321 & 0.643 & 0.857 & 0.321 & 0.335 & 0.322 & 0.299 & 0.259 \\
MIT-BIH & 0.690 & 0.708 & 0.702 & 0.690 & 0.750 & 0.760 & 0.690 & 0.706 & 0.703 & 0.702 & 0.379 \\
Petrobras 3W & 0.480 & 0.480 & 0.431 & 0.480 & 0.680 & 0.740 & 0.480 & 0.482 & 0.447 & 0.411 & 0.404 \\
\bottomrule
\end{tabular}
\end{table}

\begin{table}[H]\centering
\tiny
\setlength{\tabcolsep}{2.2pt}
\renewcommand{\arraystretch}{0.85}
\caption{Full metric grid for \textbf{TiRex\,$+$\,NR\,$+$\,Purity}, per dataset and macro-averaged, at each pollution level.}
\label{tab:a8-tirex-delta-purity}
\begin{tabular}{lccccccccccc}
\toprule
 & P@1 & P@5 & P@10 & HR@1 & HR@5 & HR@10 & N@1 & N@5 & N@10 & N@20 & mF1 \\
\midrule
\multicolumn{12}{l}{\emph{$\rho{=}0\%$}} \\
CTSR & 0.290 & 0.254 & 0.223 & 0.290 & 0.440 & 0.500 & 0.290 & 0.261 & 0.237 & 0.276 & 0.170 \\
DAMADICS & 0.333 & 0.333 & 0.267 & 0.333 & 0.500 & 0.556 & 0.333 & 0.333 & 0.285 & 0.273 & 0.242 \\
Exathlon & 0.850 & 0.844 & 0.791 & 0.850 & 0.970 & 0.990 & 0.850 & 0.845 & 0.809 & 0.749 & 0.659 \\
HAI & 0.250 & 0.221 & 0.243 & 0.250 & 0.500 & 0.679 & 0.250 & 0.221 & 0.235 & 0.234 & 0.107 \\
MIT-BIH & 0.660 & 0.672 & 0.662 & 0.660 & 0.750 & 0.820 & 0.660 & 0.670 & 0.663 & 0.664 & 0.367 \\
Petrobras 3W & 0.500 & 0.460 & 0.427 & 0.500 & 0.610 & 0.650 & 0.500 & 0.466 & 0.441 & 0.411 & 0.357 \\
\bottomrule
\end{tabular}
\end{table}

\begin{table}[H]\centering
\tiny
\setlength{\tabcolsep}{2.2pt}
\renewcommand{\arraystretch}{0.85}
\caption{Full metric grid for \textbf{Chronos-2\,$+$\,NR\,$+$\,GPC}, per dataset and macro-averaged, at each pollution level.}
\label{tab:a8-chronos2-delta-gpc}
\begin{tabular}{lccccccccccc}
\toprule
 & P@1 & P@5 & P@10 & HR@1 & HR@5 & HR@10 & N@1 & N@5 & N@10 & N@20 & mF1 \\
\midrule
\multicolumn{12}{l}{\emph{$\rho{=}0\%$}} \\
CTSR & 0.550 & 0.550 & 0.550 & 0.550 & 0.550 & 0.550 & 0.550 & 0.550 & 0.550 & 0.622 & 0.488 \\
DAMADICS & 0.222 & 0.222 & 0.222 & 0.222 & 0.222 & 0.222 & 0.222 & 0.222 & 0.222 & 0.240 & 0.095 \\
Exathlon & 0.940 & 0.940 & 0.940 & 0.940 & 0.940 & 0.940 & 0.940 & 0.940 & 0.940 & 0.940 & 0.877 \\
HAI & 0.500 & 0.500 & 0.500 & 0.500 & 0.500 & 0.500 & 0.500 & 0.500 & 0.500 & 0.500 & 0.477 \\
MIT-BIH & 0.770 & 0.770 & 0.770 & 0.770 & 0.770 & 0.770 & 0.770 & 0.770 & 0.770 & 0.770 & 0.406 \\
Petrobras 3W & 0.650 & 0.650 & 0.650 & 0.650 & 0.650 & 0.650 & 0.650 & 0.650 & 0.650 & 0.650 & 0.565 \\
\bottomrule
\end{tabular}
\end{table}

\begin{table}[H]\centering
\tiny
\setlength{\tabcolsep}{2.2pt}
\renewcommand{\arraystretch}{0.85}
\caption{Full metric grid for \textbf{Chronos-2\,$+$\,NR\,$+$\,MajVote}, per dataset and macro-averaged, at each pollution level.}
\label{tab:a8-chronos2-delta-majvote}
\begin{tabular}{lccccccccccc}
\toprule
 & P@1 & P@5 & P@10 & HR@1 & HR@5 & HR@10 & N@1 & N@5 & N@10 & N@20 & mF1 \\
\midrule
\multicolumn{12}{l}{\emph{$\rho{=}0\%$}} \\
CTSR & 0.540 & 0.540 & 0.540 & 0.540 & 0.540 & 0.540 & 0.540 & 0.540 & 0.540 & 0.591 & 0.465 \\
DAMADICS & 0.278 & 0.278 & 0.283 & 0.278 & 0.278 & 0.333 & 0.278 & 0.278 & 0.281 & 0.310 & 0.139 \\
Exathlon & 0.870 & 0.870 & 0.870 & 0.870 & 0.870 & 0.870 & 0.870 & 0.870 & 0.870 & 0.870 & 0.715 \\
HAI & 0.429 & 0.429 & 0.429 & 0.429 & 0.429 & 0.429 & 0.429 & 0.429 & 0.429 & 0.429 & 0.401 \\
MIT-BIH & 0.760 & 0.760 & 0.760 & 0.760 & 0.760 & 0.760 & 0.760 & 0.760 & 0.760 & 0.760 & 0.396 \\
Petrobras 3W & 0.510 & 0.510 & 0.510 & 0.510 & 0.510 & 0.510 & 0.510 & 0.510 & 0.510 & 0.510 & 0.442 \\
\bottomrule
\end{tabular}
\end{table}

\begin{table}[H]\centering
\tiny
\setlength{\tabcolsep}{2.2pt}
\renewcommand{\arraystretch}{0.85}
\caption{Full metric grid for \textbf{Chronos-2\,$+$\,NR\,$+$\,QPurity}, per dataset and macro-averaged, at each pollution level.}
\label{tab:a8-chronos2-delta-qpurity}
\begin{tabular}{lccccccccccc}
\toprule
 & P@1 & P@5 & P@10 & HR@1 & HR@5 & HR@10 & N@1 & N@5 & N@10 & N@20 & mF1 \\
\midrule
\multicolumn{12}{l}{\emph{$\rho{=}0\%$}} \\
CTSR & 0.550 & 0.444 & 0.368 & 0.550 & 0.670 & 0.790 & 0.550 & 0.468 & 0.407 & 0.446 & 0.466 \\
DAMADICS & 0.389 & 0.333 & 0.300 & 0.389 & 0.500 & 0.611 & 0.389 & 0.346 & 0.325 & 0.329 & 0.243 \\
Exathlon & 0.860 & 0.842 & 0.837 & 0.860 & 0.870 & 0.900 & 0.860 & 0.844 & 0.840 & 0.821 & 0.629 \\
HAI & 0.357 & 0.393 & 0.371 & 0.357 & 0.643 & 0.679 & 0.357 & 0.393 & 0.378 & 0.363 & 0.352 \\
MIT-BIH & 0.750 & 0.734 & 0.744 & 0.750 & 0.760 & 0.810 & 0.750 & 0.737 & 0.743 & 0.738 & 0.396 \\
Petrobras 3W & 0.620 & 0.520 & 0.488 & 0.620 & 0.750 & 0.800 & 0.620 & 0.540 & 0.511 & 0.468 & 0.500 \\
\bottomrule
\end{tabular}
\end{table}

\begin{table}[H]\centering
\tiny
\setlength{\tabcolsep}{2.2pt}
\renewcommand{\arraystretch}{0.85}
\caption{Full metric grid for \textbf{Chronos-2\,$+$\,NR\,$+$\,PRF}, per dataset and macro-averaged, at each pollution level.}
\label{tab:a8-chronos2-delta-prf}
\begin{tabular}{lccccccccccc}
\toprule
 & P@1 & P@5 & P@10 & HR@1 & HR@5 & HR@10 & N@1 & N@5 & N@10 & N@20 & mF1 \\
\midrule
\multicolumn{12}{l}{\emph{$\rho{=}0\%$}} \\
CTSR & 0.420 & 0.372 & 0.318 & 0.420 & 0.700 & 0.780 & 0.420 & 0.382 & 0.341 & 0.405 & 0.396 \\
DAMADICS & 0.222 & 0.278 & 0.294 & 0.222 & 0.667 & 0.667 & 0.222 & 0.266 & 0.282 & 0.289 & 0.095 \\
Exathlon & 0.840 & 0.854 & 0.846 & 0.840 & 0.930 & 0.940 & 0.840 & 0.850 & 0.846 & 0.822 & 0.714 \\
HAI & 0.393 & 0.393 & 0.361 & 0.393 & 0.750 & 0.821 & 0.393 & 0.389 & 0.369 & 0.335 & 0.398 \\
MIT-BIH & 0.750 & 0.738 & 0.740 & 0.750 & 0.780 & 0.810 & 0.750 & 0.739 & 0.740 & 0.734 & 0.392 \\
Petrobras 3W & 0.450 & 0.442 & 0.423 & 0.450 & 0.730 & 0.780 & 0.450 & 0.447 & 0.432 & 0.397 & 0.308 \\
\bottomrule
\end{tabular}
\end{table}

\begin{table}[H]\centering
\tiny
\setlength{\tabcolsep}{2.2pt}
\renewcommand{\arraystretch}{0.85}
\caption{Full metric grid for \textbf{Chronos-2\,$+$\,NR\,$+$\,Purity}, per dataset and macro-averaged, at each pollution level.}
\label{tab:a8-chronos2-delta-purity}
\begin{tabular}{lccccccccccc}
\toprule
 & P@1 & P@5 & P@10 & HR@1 & HR@5 & HR@10 & N@1 & N@5 & N@10 & N@20 & mF1 \\
\midrule
\multicolumn{12}{l}{\emph{$\rho{=}0\%$}} \\
CTSR & 0.270 & 0.206 & 0.176 & 0.270 & 0.370 & 0.430 & 0.270 & 0.219 & 0.194 & 0.231 & 0.142 \\
DAMADICS & 0.167 & 0.178 & 0.217 & 0.167 & 0.389 & 0.611 & 0.167 & 0.171 & 0.203 & 0.214 & 0.079 \\
Exathlon & 0.830 & 0.826 & 0.804 & 0.830 & 0.930 & 0.960 & 0.830 & 0.828 & 0.812 & 0.781 & 0.573 \\
HAI & 0.357 & 0.343 & 0.325 & 0.357 & 0.679 & 0.750 & 0.357 & 0.354 & 0.337 & 0.331 & 0.324 \\
MIT-BIH & 0.730 & 0.726 & 0.717 & 0.730 & 0.820 & 0.850 & 0.730 & 0.727 & 0.720 & 0.715 & 0.396 \\
Petrobras 3W & 0.540 & 0.468 & 0.426 & 0.540 & 0.650 & 0.700 & 0.540 & 0.484 & 0.449 & 0.405 & 0.296 \\
\bottomrule
\end{tabular}
\end{table}

\begin{table}[H]\centering
\tiny
\setlength{\tabcolsep}{2.2pt}
\renewcommand{\arraystretch}{0.85}
\caption{Full metric grid for \textbf{MR-Hydra}, per dataset and macro-averaged, at each pollution level.}
\label{tab:a8-mrhydra}
\begin{tabular}{lccccccccccc}
\toprule
 & P@1 & P@5 & P@10 & HR@1 & HR@5 & HR@10 & N@1 & N@5 & N@10 & N@20 & mF1 \\
\midrule
\multicolumn{12}{l}{\emph{$\rho{=}0\%$}} \\
CTSR & 0.750 & 0.770 & 0.768 & 0.750 & 0.800 & 0.800 & 0.750 & 0.767 & 0.767 & 0.783 & 0.743 \\
DAMADICS & 0.667 & 0.667 & 0.667 & 0.667 & 0.667 & 0.778 & 0.667 & 0.667 & 0.689 & 0.712 & 0.478 \\
Exathlon & 0.960 & 0.966 & 0.965 & 0.960 & 0.970 & 0.970 & 0.960 & 0.964 & 0.964 & 0.964 & 0.927 \\
HAI & 0.321 & 0.379 & 0.382 & 0.321 & 0.571 & 0.571 & 0.321 & 0.367 & 0.374 & 0.414 & 0.200 \\
MIT-BIH & 0.760 & 0.838 & 0.823 & 0.760 & 0.910 & 0.910 & 0.760 & 0.824 & 0.819 & 0.826 & 0.640 \\
Petrobras 3W & 0.840 & 0.850 & 0.863 & 0.840 & 0.920 & 0.930 & 0.840 & 0.847 & 0.857 & 0.861 & 0.732 \\
\bottomrule
\end{tabular}
\end{table}

\begin{table}[H]\centering
\tiny
\setlength{\tabcolsep}{2.2pt}
\renewcommand{\arraystretch}{0.85}
\caption{Full metric grid for \textbf{RDST}, per dataset and macro-averaged, at each pollution level.}
\label{tab:a8-rdst}
\begin{tabular}{lccccccccccc}
\toprule
 & P@1 & P@5 & P@10 & HR@1 & HR@5 & HR@10 & N@1 & N@5 & N@10 & N@20 & mF1 \\
\midrule
\multicolumn{12}{l}{\emph{$\rho{=}0\%$}} \\
CTSR & 0.670 & 0.670 & 0.670 & 0.670 & 0.670 & 0.670 & 0.670 & 0.670 & 0.670 & 0.673 & 0.605 \\
DAMADICS & 0.667 & 0.667 & 0.667 & 0.667 & 0.667 & 0.667 & 0.667 & 0.667 & 0.667 & 0.667 & 0.392 \\
Exathlon & 0.980 & 0.980 & 0.980 & 0.980 & 0.980 & 0.980 & 0.980 & 0.980 & 0.980 & 0.980 & 0.989 \\
HAI & 0.357 & 0.357 & 0.357 & 0.357 & 0.357 & 0.357 & 0.357 & 0.357 & 0.357 & 0.357 & 0.251 \\
MIT-BIH & 0.580 & 0.580 & 0.580 & 0.580 & 0.580 & 0.580 & 0.580 & 0.580 & 0.580 & 0.580 & 0.628 \\
Petrobras 3W & 0.740 & 0.740 & 0.740 & 0.740 & 0.740 & 0.740 & 0.740 & 0.740 & 0.740 & 0.740 & 0.627 \\
\bottomrule
\end{tabular}
\end{table}

\end{document}